\documentclass[preprint,12pt]{elsarticle}

\usepackage[dvipsnames]{xcolor}
\usepackage{relsize,exscale}
\usepackage{amsmath}
\usepackage{graphicx}
\usepackage[labelfont=bf]{caption}
\usepackage{subcaption}
\usepackage{float}
\usepackage{multirow}
\usepackage{tabularx, ragged2e}
\usepackage{tabu}
\usepackage{enumitem}
\usepackage{hyperref}
\usepackage{siunitx}
\usepackage{booktabs}
\usepackage{comment}
\usepackage{url}
\usepackage{rotating}
\usepackage{pdflscape}

\usepackage{amssymb}

\journal{Cell}

\begin{document}

\begin{frontmatter}



\title{Interpretable and synergistic deep learning for visual explanations and statistical estimations of segmentation of disease features from medical images}


\author[label1]{Sambuddha Ghosal}
\author[label1,label2]{Pratik Shah}

\address[label1]{Massachusetts Institute of Technology, 20 Ames Street, Cambridge MA 02139, USA}
\address[label2]{Corresponding author, pratiks@mit.edu}



\begin{abstract}
Deep learning (DL) models for disease classification or segmentation from medical images are increasingly trained using transfer learning (TL) from unrelated natural world images. However, shortcomings and utility of TL for specialized tasks in the medical imaging domain remain unknown and are based on assumptions that increasing training data will improve performance. We report detailed comparisons, rigorous statistical analysis and comparisons of widely used DL architecture for binary segmentation after TL with ImageNet initialization ($T_{II}$-models) with supervised learning with only medical images ($L_{MI}$-models) of macroscopic optical skin cancer, microscopic prostate core biopsy and Computed Tomography (CT) DICOM images. Through visual inspection of $T_{II}$ and $L_{MI}$ model outputs and their Grad-CAM counterparts, our results identify several counter intuitive scenarios where automated segmentation of one tumor by both models or the use of individual segmentation output masks in various combinations from individual models leads to 10\% increase in performance. We also report sophisticated ensemble DL strategies for achieving clinical grade medical image segmentation and model explanations under low data regimes. For example; estimating performance, explanations and replicability of $L_{MI}$ and $T_{II}$ models described by us can be used for situations in which sparsity promotes better learning. A free GitHub repository of $T_{II}$ and $L_{MI}$ models, code and more than 10,000 medical images and their Grad-CAM output from this study can be used as starting points for advanced computational medicine and DL research for biomedical discovery and applications.
\end{abstract}


\end{frontmatter}


\section*{Introduction}
\label{Sec:1}
Artificial intelligence (AI) applications for precision medicine using medical images require large amounts of well-annotated data. Over the past 10 years thousands~\cite{suzuki2017overview} of research studies have reported the optimization of small amounts of available data to conduct research on deep neural networks (DNNs) for medical image classification, and segmentation~\cite{yauney2017convolutional,rana2020use,javia2018machine} to generate prototypes for real world applications. Researchers routinely leverage the learned weights of pretrained natural world deep learning (DL) models such as AlexNet~\cite{krizhevsky2012imagenet} to fine-tune the use of transfer learning (TL) on medical image segmentation to improve performance. Visual Geometry Group-16 (VGG16)~\cite{simonyan2014very}, U-Net~\cite{ronneberger2015u} and other DL architectures are routinely used to fine-tune DL models~\cite{raghu2019transfusion}. Anecdotal assumptions of superior performance of TL compared with models trained exclusively using medical images, termed as being trained from scratch (TFS) are prevalent despite being poorly understood~\cite{huynh2016digital,nappi2016deep}. For example, when images are similar and even if the classes are not the same, researchers consider that an augmented/larger data set or TL could train better models with superior accuracy than TFS. In a recent study~\cite{raghu2019transfusion}, authors report that transfer learning from natural world images did not increase the performance of classification tasks for medical image data. Medical images have complex biological textures and several high gradient regions sensitive to small perturbations by previously unseen images~\cite{ghorbani2020dermgan}. Variances such as illumination, optics, clinical labels and out-of-training data examples~\cite{finlayson2019adversarial} have  also been reported to dramatically reduce DL model performance~\cite{paschali2018generalizability}. State-of-the-art DNNs designed for large-scale natural image processing consequently get overparameterized for medical imaging tasks and remain vulnerable to adversarial attacks~\cite{finlayson2019adversarial}. Additionally, 90-95\% of TFS and TL deep learning models demonstrate low specificity, sensitivity and accuracy for medical grade segmentation precluding their clinical utility~\cite{chen2019deep,kelly2019key}. Neural network explanation methods~\cite{ghosal2018explainable,simonyan2013deep,ghosal2019weakly,pokuri2019interpretable} and image based visualizations such as gradient-weighted class activation mapping (Grad-CAM)~\cite{selvaraju2017grad} are useful for visualizing and interpreting performance and mechanisms of DL models. Resources such as trained TL and TFS models that are benchmarked using statistical estimation along with visual Grad-CAM explanations for different medical segmentation tasks can thus be extremely valuable for computationally and clinically meaningful inferences.

Another critical area often ignored or not fully understood is biases and confounders introduced by ``splitting” biomedical data into training and test sets for validating the performance of DL models. For example, most DL studies~\cite{mckinney2020international, ardila2019end} that achieve better-than-clinician performance routinely use $\approx$ 10,000-100,0000 medical images to train models from scratch. However, access to such a large quantity of medical image data is unfeasible for most researchers working with fewer images. Added complexity is introduced under low data regimes, in which the splitting of medical data leads to a significant imbalance in the percentage of images and associated clinical labels available for performance estimation of TFS or TL models~\cite{shie2015transfer}. Absent large numbers of balanced and diverse examples for each classification or segmentation target class are often causal to the limited performance of DL approaches. As a result, DL models memorize rather than learn and perform poorly when tested with out-of-distribution (OoD) but related medical tasks~\cite{zech2018variable}. Thus, the use of random 80:20 splits to generate training and validation images results in over or under fitting on biological or underlying clinical labels, thereby leading to the poor reproducibility and replicability of DL models~\cite{collins2019reporting}. One strategy for overcoming these challenges is to collect large and diverse numbers of all types of medical images to train perfect DL models as in non-medical applications~\cite{ILSVRC15}. For example, AlexNet~\cite{krizhevsky2012imagenet}, a deep convolutional neural network (CNN) trained on the ImageNet database~\cite{ILSVRC15}, is a benchmarked and widely used model for natural world image classification. AlexNet was trained using learning from more than one million images (1,000 images in each of the 1,000 categories split into 1.2 million training, 50,000 validation, and 150,000 test images) to achieve high performance levels (top-5 error of 15.3\%)~\cite{krizhevsky2012imagenet}. This approach is unlikely to scale for medical images that are typically available in smaller numbers, or are heterogeneous in clinical labels, intended outcomes and optical properties. The generation of handcrafted interpretable DL models from small amounts of medical image data is thus considered valuable and widely prevalent in computational medicine and vision communities~\cite{chen2019deep,kelly2019key,shah2019artificial}.

In this study, we train and validate DL models of VGG-UNet architecture by either $T_{II}$ models (trained with supervised transfer learning with ImageNet initialization and fine-tuned on medical image data of choice) or $L_{MI}$: models (trained with supervised learning trained natively on only medical image data) for the binary segmentation of clinical disease features and organs. Statistical estimations and the visual explanation of DL model performance on three unique medical image types that are widely used in computational medicine research were set as the targets. Our goal was to not only create the best-performing DL models, but also provide resources to support researchers working with small numbers of medical images using $L_{MI}$ or $T_{II}$. We report (a) the detailed validation of a process for maximizing available clinical labels and medical images for synergistic use of $T_{II}$ and $L_{MI}$ models; (b) a description of parametric and non-parametric statistical methods to test significance and estimate performance using AUROC, F1 score, sensitivity and the specificity of segmentation of disease features; (c) an estimation of the least numbers and types of individual images and target clinical labels for model performance improvement; and (d) Grad-CAM-based visualization and interpretation of $L_{MI}$ and $T_{II}$ DL models and medical image features for explainable segmentation of diseases. We release 30 (and a larger set of 270 derived from five replicates) fully trained and validated DL models and more than 10,000 output images benchmarked for performance using $T_{II}$ or $L_{MI}$ under high and low data regimes. This work communicates specific use cases when $L_{MI}$ and $T_{II}$ can be employed synergistically or in ensemble configurations to improve performance or reduce false positive and negative clinical diagnosis. To our knowledge, this is the largest repository of detailed characterizations of medical image segmentation performance explanation by DL models. Novel knowledge, methods and DL models developed as part of this study can serve as valuable resources and starting points for the segmentation of different medical image types used widely in computational medicine and computer science communities.

\renewcommand{\thesubfigure}{\roman{subfigure}}

\section*{Results}\label{sec:Results}
After estimation of the optimal 80:20 data split and clinical label distributions for each of the three data sets used in this study; the median, mean and standard deviations of distributions of AUROC, Dice scores, sensitivity and specificity were evaluated for performance estimations of $T_{II}$ and $L_{MI}$ models \textbf{(Fig.~\ref{fig1} and Table~\ref{tab1})}. For each image in the test data sets, we calculated pairwise differences between their medians ($\Delta_m$) of individual performance metrics achieved by either $T_{II}$ or $L_{MI}$ models (Table~\ref{tab2}). Additionally, we reported the number of images in each data set that achieved an AUROC or Dice score value greater than or equal to 0.9 (a common threshold to indicate superior segmentation) with 1 representing a perfect score \textbf{(Table~\ref{tab2})}. Furthermore, transfer-learned $T_{II}$ models achieved higher (p $<$ 0.05) mean and median Dice, AUROC, sensitivity and specificity scores for the segmentation of both benign and malignant skin cancer lesions from RGB images (Table~\ref{tab1}). Approximately 13,000 images with binary, benign (n = 12,668), malignant (n = 1,118) clinical labels (Table~\ref{stab2}) of skin cancer diagnoses were used in this study. Here, $T_{II}$ outperformed $L_{MI}$ models with higher performance values and statistically significant (p $<$ 0.05) differences in the segmentation of all benign and malignant skin cancer lesions across 2,758 test images (Table~\ref{tab1} and Fig.~\ref{fig2} [i] and [ii]). Moreover, $T_{II}$ models also achieved higher mean AUROC (86\%) and Dice scores (78\%) compared with $L_{MI}$ models across all skin cancers (Table~\ref{tab2}). Greater numbers of skin images were segmented with 0.9 or higher mean AUROC (n = 1,651 images), Dice scores (n = 1,177) and sensitivity (n = 1,301) by $T_{II}$ models (Table~\ref{tab2}). $T_{II}$ models consistently achieved scores higher than $\approx$ 0.9 (AUROC) for 1,651 (59\% of total images) compared with 984 images by $L_{MI}$ models for segmentation of any skin cancer (Table~\ref{tab2}). Additionally, 40\% (1,177) of all skin cancer images were segmented with Dice scores greater than 0.9 by $T_{II}$ models compared with 671 (22\%) by $L_{MI}$ (Table~\ref{tab2}). Meanwhile, binary segmentation using $L_{MI}$ models performed equally well in detecting background pixels, similar to transfer-learned models (Table~\ref{tab1} and Table~\ref{tab2}).

Visual explanations derived from image-based analysis, revealed that $T_{II}$ models had lower false negative and higher true positive regions than corresponding outputs from $L_{MI}$ models for segmentation of benign \textbf{(Fig.~\ref{fig2})} and malignant skin cancer lesions \textbf{(Fig.~\ref{fig3})}. The majority of $L_{MI}$ outputs in panel D in Fig.~\ref{fig2} (i) and Fig.~\ref{fig3} (i) demonstrated higher false negative (yellow) and lower true positive detection (green) compared with $T_{II}$ models (panel F of Fig.~\ref{fig2} [i] and Fig.~\ref{fig3} [i]). Grad-CAM analysis revealed that $L_{MI}$ models were less capable of distinguishing, given their lowered activation, between background skin or non-cancer areas surrounding benign (Fig.~\ref{fig2} [ii] panels B and C) and malignant skin lesions (Fig.~\ref{fig3} [ii] panels B and C). Additionally, Grad-CAM outputs revealed that the higher segmentation accuracy of $T_{II}$ models was based on their higher and more precise activation and ability to distinguish benign (Fig.~\ref{fig2} [ii] panel E) or malignant (Fig.~\ref{fig3} [ii] panel E) cancer skin lesions from background skim compared with $L_{MI}$ models. For larger and diffuse malignant lesions, $T_{II}$ models were not deceived by background skin with red coloration surrounding cancer moles (Fig.~\ref{fig3} [i], panels 1A and 1F) compared with $L_{MI}$ models, which revealed high false negatives for those regions (Fig.~\ref{fig3} [i], panels 1A and 1D). Interestingly, despite less than 100\% coverage for the segmentation of malignant skin lesions (Fig.~\ref{fig3}), $T_{II}$ models achieved higher AUROC and Dice scores (Table~\ref{tab1} and Table~\ref{stab3}). This may be because ground-truth clinical annotations for malignant tumors were coarse generic labels, which may include skin without tumors. Alternatively. the trained $T_{II}$ models, although greatly superior to $L_{MI}$ models, could not comprehensively distinguish malignant moles from surrounding skin that resembled other cancer lesions in the training data. Correspondingly, Grad-CAM activation profiles revealed that $T_{II}$ models were much more likely to focus or preferentially activate regions with brighter colors or explicit shapes in benign and malignant lesions [Fig.~\ref{fig3} [ii], panels A, C and E]. $T_{II}$ models' superior performance could be explained by the higher activation and precise learning of key features of moles/lesions causal for superior binary segmentation compared with $L_{MI}$ models. The presence of artifacts (e.g., stickers) (panel 1A of Fig.~\ref{fig2}) were identified as background by both $T_{II}$ and $L_{MI}$ models across the data set. Additionally, both models performed equally well in terms of specificity to reduce false positives by not segmenting the majority of healthy tissues and other background pixels as tumors; this may be attributed to shared features and RGB pixel intensities common between these classes of images taken under natural lighting conditions and with sufficient availability of examples of most of the target clinical labels in skin image data. Thus, a transfer-learned $T_{II}$ model could leverage previously learned features from natural world images to segment skin cancer from RGB images, and it exhibited higher discriminatory ability for tumors and lower false negatives and robustness against artifacts in the data.

For the segmentation of prostate tumors, $L_{MI}$ models were trained with $\approx$ 250 whole slide Hematoxylin and Eosin (H $\&$ E) images of prostate core biopsy images using 80\% of the available data (224 images with any tumor labels and 20 without tumors) and achieved slightly higher median AUROC and Dice scores compared with $T_{II}$ models (Table~\ref{tab1} and Fig.~\ref{sfig2} [iii] and [iv]). Furthermore, more prostate core biopsy images segmented by $L_{MI}$ models had overall higher mean AUROC, specificity and Dice scores for tumor segmentation (Table~\ref{tab2}). Additionally, approximately 65\% of images were segmented with higher AUROC and Dice scores by $L_{MI}$ models (Table~\ref{tab2}). $L_{MI}$ models also achieved 0.9 or higher AUROC and Dice scores for 34 test images (61\%) compared with 29 for $T_{II}$. (Table~\ref{tab2}). The higher performance of $L_{MI}$ models’ metric distributions, save for sensitivity (Table\ref{tab1}), did not reach statistical significance of p $<$ 0.05 (Table~\ref{tab2}, Fig.~\ref{sfig2} [iii] and [iv]). \textbf{Fig.~\ref{fig4}} (i) illustrates that both $T_{II}$ and $L_{MI}$ outputs had false negative (colored yellow) regions, but also revealed comparable true positive regions for images with prostate tumors (green). Gleason grade tumor regions were highlighted by $L_{MI}$ models with higher intensity (Supplementary Fig.~\ref{sfig7} [ii] panel C) following Grad-CAM activations for making predictions compared with $T_{II}$ models (Fig.~\ref{sfig7} [ii] panel E). $L_{MI}$ models were thus accurate and more precise at making predictions and demarcating tissues from non-tissue (background) pixels. Conversely, $T_{II}$ models could not segment all tumorous tissue regions and exhibited higher false negatives (Supplementary Fig.~\ref{sfig7} [i] and [ii] images 3 and 4). Two images were without tumors in the test data; $L_{MI}$ and $T_{II}$ models were able to correctly predict one image (image 3 in Fig.~\ref{fig4} [i] and [ii]) as true negative for prostate tumor by actively learning to distinguish those tissue regions. Grad-CAM analysis (Fig.~\ref{fig4} [ii], image 3) highlighted the background region when queried with target class 0 and did not highlight any regions when queried with target class 1 (which corresponds to tumor regions). In contrast, both models incorrectly segmented benign tissue region from image 4 in Fig.~\ref{fig4} (i) and (ii) as tumorous, which was maintained by Grad-CAM outputs. Visual explanations from the image analysis of microscopic RGB images of prostate core biopsy revealed that $L_{MI}$ models achieved slightly better metrics than $T_{II}$ (Table 1). However, unlike the unequivocal performance gain by transfer-learned $T_{II}$ models for skin cancer images, supervised learning with medical images using native prostate data improved segmentation by less than 5\% (see Table~\ref{tab1} and Table~\ref{tab2}) and achieved statistically significant differences in sensitivity. We reason this based, in part, on the lower complexity and fewer numbers of individual histology images necessary for achieving an accurate segmentation of prostate tumors versus skin cancer. Thus, for prostate core biopsy images, $L_{MI}$ models trained using randomly initialized weights can be equally as good if not superior at differentiating background and/or healthy tissue from that with tumors. Conversely, $L_{MI}$ models were less capable at learning tumor segmentation represented by lower numbers of examples in training data for the prostate core biopsy image data set (represented by images that did not have any tumors or had only benign tissues), in which transfer-learned $T_{II}$ models performed better. We reason that DL models trained on natural world images, such as the ImageNet database, lack sufficient and specialized information on the complex domain of microscopic pathology but can be used conjunctively for superior segmentation not achieved by models trained using only pathology biopsy images. A corollary is that transfer learning from larger, more heterogeneous data and model weights to a smaller data set causes suboptimal parameterization on the learning background. Additionally, training using only pathology biopsy images ($L_{MI}$ models) might be beneficial for the comprehensive and accurate learning of detailed features for the demarcation of tumors from non-tumor and healthy tissues and other background pixels.

Forty-five thousand abdominal CT DICOM images with 16,336 kidney tissue and 29,088 non-kidney or background clinical labels were used in this study (Table S2). Differences in the distributions of AUROC and sensitivity were statistically significant (p $<$ 0.05) between models, with $L_{MI}$ performing slightly better but with less than a 5\% gain in performance (Table~\ref{tab1} and Supplementary Fig.~\ref{sfig2} [v] and [vi]). $L_{MI}$ models performed better with 1,937 test images achieving higher AUROC scores ($\approx$ 21\% of total test images, or $\approx$ 60\% of images with kidneys), and images segmented by $T_{II}$ models achieving higher Dice scores ($\approx$ 18\% of all test images or $\approx$ 50\% of images with kidneys) (Table~\ref{tab2} and Table~\ref{stab4}). $L_{MI}$ sensitivity was 0.9985, while $T_{II}$ sensitivity was 0.9979, and the differences were statistically significant. Differences between AUROC and sensitivity distributions achieved by $L_{MI}$ and $T_{II}$ models were statistically significant (p $<$ 0.05). This data set contained significant numbers of images (5,840 out of 9,085) without kidneys (i.e., corresponding ground-truth masks [labels] were black pixels). Both models achieved a perfect median score of $\approx$ 1 for specificity, and comparable Dice scores for segmentation of kidneys from background and other organs (Table 1). Inherent embedded features and properties of gray-scale CT images are unlike natural world images in the ImageNet database. Values for AUROC and Dice scores indicated that both models performed comparably, with marginal performance ($<$ 5\%) gains for $L_{MI}$. Representative CT images with (\textbf{Fig.~\ref{fig5}} and Supplementary Fig.~\ref{sfig8}) and without kidneys or with other organs (Fig.~\ref{fig5}) are displayed for performance evaluation. For most test images, both models could segment out kidney tissue regions with reasonable accuracy (Fig.~\ref{fig5} images 1, 2 and 3). In a few test images, $L_{MI}$ models segmented larger areas of kidney regions (green regions in Fig.~\ref{sfig8} [i], panels D and F), while $T_{II}$ models demonstrated higher false negative (yellow regions) rates. Grad-CAM revealed that both $L_{MI}$ and $T_{II}$ models could distinguish background regions and exhibited the least activation for regions without kidneys (Class 0) (Fig.~\ref{fig5} [ii] panels C E). In particular, $T_{II}$ models had lower sensitivity in demarcating kidneys and non-target tissues from other organs (Table~\ref{tab1} and Table~\ref{tab2}). For image 4 in Fig.~\ref{fig5}, both models reveal false negative segmentations, and corresponding Grad-CAM outputs (Fig.~\ref{fig5} [ii] panel C4 and E4 for $L_{MI}$ and $T_{II}$) did not activate in the final model outputs. Other images in the test data set had similar outcomes, in which both models failed to segment small kidney tissue regions. In summary, when target class kidney tissues were available in lower numbers than in the background, when there were other organs, or when the presence of other organs and tissues outnumbered the presence of kidney tissues, training $L_{MI}$ models from scratch for targeted segmentation was more optimal for higher sensitivity. Meanwhile, in all other scenarios $L_{MI}$ and $T_{II}$ models performed equally well and Grad-CAM explanations suggest they can be used synergistically and interchangeably.

Randomization of available data into 80:20 proportions is standard practice in DL research. Moreover, researchers also use repeats of individual model runs to fine tune and estimate the reproducibility of the results. Randomization may result in bias due to a skewed or non-Gaussian distribution of labels and clinical data available in smaller numbers. In this study after medical images were randomly split into five 80:20 proportions, segmentation performance was observed to be contingent on the DL model, image category, and Gaussian distribution of clinical labels (Fig.~\ref{fig1}). In the prostate core biopsy data (n = 244), fivefold randomization into 80:20 splits resulted in approximately 195 training and 44 test images. Deep learning models needed to learn clinical-grade segmentation of regions with Gleason grade 3, 4 or 5 tumors from approximately 177 training images and were evaluated using 49 test images. This resulted in $\approx$ 8\% (17 images) of training and 4-8\% (4 images) of test data without tumors, and 90\% of training and 91-93\% of test data images with tumors. Under these data and label regimes (Supplementary Table~\ref{stab2}), $L_{MI}$ and $T_{II}$ models demonstrated similar performance for tumor segmentation and accurate detection of background skin across the majority of test images for all five data splits (Table~\ref{tab1}, Supplementary Fig.~\ref{sfig2}). The computed tomography DICOM data set contained 45,424 2D images split into $\approx$ 36,000 training and 9,000 test sets following fivefold randomization.  $T_{II}$ or $L_{MI}$ DL models were trained using 13,000 images with kidneys (36\%) and 23,000 (64\%) without kidneys. Evaluations were performed on approximately 3,200 (36\%) images with and 5,800 (64\%) images without kidneys in the test data (Supplementary Table~\ref{stab2}). Despite lower numbers of training images with kidneys (Supplementary Table~\ref{stab2}), $L_{MI}$ models trained via supervised learning with CT images performed equally as well or marginally better than $T_{II}$ models for a less complex task of detection and kidney segmentation in all five splits (Supplementary Fig.~\ref{sfig2} and Table~\ref{stab4}). Large numbers of data with other organs or only background did not seem to impact model performances across the fivefold repeats of the selected split or for the other four splits (Supplementary Fig.~\ref{sfig2} and Table~\ref{stab4}). In skin data (n = 13,786), fivefold random shuffling and 80:20 splits resulted in approximately 11,000 training and 2,700 test images. Benign or malignant moles and lesions were randomized equally across five training ($\approx$ 91\%) and test ($\approx$ 7\%) splits for individual data (Supplementary Table~\ref{stab2}). However, DL models trained on data split 1 for skin images demonstrated higher segmentation accuracies for clinical class-based distributions achieving divergent performances for moles and lesions available in lower frequencies (Supplementary Table~\ref{stab3}). Our results indicate that blind and random 80:20 splitting did not seem to impact model performance for medical data sets with limited heterogeneity, simple segmentation, and Gaussian distributions of individual classes, but it was not optimal for larger and more heterogeneous clinical tasks such as skin image segmentation in our study (Supplementary Table~\ref{stab1}).

Performance trends for 80:20 split DL experiments were then compared with models trained under low data regime circumstances. \textbf{Table~\ref{tab3}} presents the median values of non-Gaussian distributions of performance metrics for each image modality obtained from depletion experiments. $L_{MI}$ and $T_{II}$ models exhibited a gradual decrease in performance as training data were depleted across all three image types. Differences between $L_{MI}$ and $T_{II}$ model performance were statistically significant for all skin cancer training data regimens, underscoring the impact of heterogeneity and Gaussian distribution of clinical labels on deep learning. Under low data routines (10-20\% training data), transfer-learned $T_{II}$ models outperformed $L_{MI}$ with higher Dice and AUROC scores for skin cancer segmentation. Meanwhile, both models achieved similar Dice and AUROC scores when 40 and 60\% of the training data was available. Interestingly, $T_{II}$ models achieved higher sensitivity (higher true positive pixel segmentations for skin cancer) for all data splits, while both models performed equally well in terms of specificity in reducing false positives (not segmenting healthy tissues or background pixels as tumors). For prostate core biopsy images, $T_{II}$ and $L_{MI}$ models demonstrated an overall gradual decrease in performance for AUROC and sensitivity scores with a decrease in training data availability and with statistically significant differences at 10\% and 60\%. For CT images, all DL models exhibited an increasing trend (less than 5\%) for AUROC, Dice and sensitivity scores with an increase in training data size. For this data set as stated earlier, $L_{MI}$ marginally outperformed $T_{II}$ models except at the 10\% training data mark for the Dice score only (Table~\ref{tab3}).

\section*{Discussion}\label{sec:Discussion}
Previous studies reported performance gain predominantly for offered classification tasks, following transfer learning between medical images of the same modalities but from different sources (magnetic resonance imaging [MRI]) to MRI,~\cite{ghafoorian2017transfer,van2014transfer} cross-modality learning (MRI-CT),~\cite{dou2018unsupervised} or between natural world images and medical images.~\cite{raghu2019transfusion,van2015off,bar2015deep,ciompi2015automatic,shie2015transfer} For example, one study~\cite{banerjee2018transfer} claimed that use of pretrained weights for the segmentation of MRI images by fine-tuning the pretrained AlexNet CNN model to classify RGB fused images achieved better performance. Meanwhile, other studies~\cite{giorgiani2020satellite} reported that for satellite image segmentation, transfer learning from ImageNet provides 30\% savings in computational costs while achieving accuracy levels comparable to training from scratch. Similarly, transfer between similar image modalities (from legacy MRI to MRI) achieved a 0.63 Dice score via fine-tuning on only two images from the target domain.~\cite{ghafoorian2017transfer} This, in turn, reduces the computational complexity by decreasing the training time, and it allows for the circumvention of low data regimes. Other studies~\cite{dou2018unsupervised} have explored the transfer between different medical image modalities (MRI-CT) for image segmentation, in which a domain adaption protocol was utilized to adapt a CNN trained with MRI images to unpaired CT data for cardiac structure segmentation. The main hypothesis or assumptions in these studies were that CNN DL models trained using transfer learning, from large ImageNet or medical data sets to smaller numbers of target images of limited scales, can achieve better performance, reduce the computational training cost, and reduce overfitting on a small training sample.

However, a growing body of emerging literature is skeptical of the true effects and benefits of transfer learning from natural world images such as ImageNet for specialized tasks in the medical imaging domain. For example, He et. al~\cite{he2019rethinking} report that random initialization is surprisingly robust even in the low data regime (10\% available training data), and ImageNet pre-training speeds up convergence early in training but does not necessarily provide regularization or improve the final target task accuracy in their study. Meanwhile, Raghu et. al~\cite{raghu2019transfusion} inspected the effects of transfer from natural world images for two large scale medical imaging classification tasks: chest x-rays and retinal fundus photographs. They found that transfer does not significantly aid performance, and a model’s performance on the ImageNet database does not translate to the medical domain.~\cite{raghu2019transfusion} Kornblith and coworkers~\cite{kornblith2019better} reported that learned features from ImageNet do not transfer well, for fine-grained medical image classification tasks and another study~\cite{raghu2019transfusion} reported that transfer from ImageNet does not significantly aid performance compared with smaller, simpler convolutional models trained natively using only medical images to classify retinal fundus and chest X-ray images. These studies suggest that some of the shortcomings of DL models trained natively using medical images may be attributed to over-parameterization rather than sophisticated feature reuse thought to be provided by transfer learning.

The role of transfer learning and training natively using medical images for medical segmentation tasks is not as well characterized or understood. In addition, detailed characterization of the effects of available learning methods for medical image segmentation across different image modalities has not been reported. In this study, we present several interesting findings on the use of statistical methods, and visual explanations of DNN models for high-accuracy segmentation of clinical information from medical images. Primarily, we focus on two widely used DNN training strategies: transfer learning from natural world images (or, initializing the network with pretrained weights from ImageNet to another model) and training-from-scratch (or, initializing the network with random weights). In this study, several commonly held practices and anecdotal concepts are put to the test, such as randomization and proportioning of training and test data, in addition to model selection and performance under low data regimes for estimating the efficacy of clinical-grade DNN segmentations. We report that although transfer learning from natural world images has benefits in certain cases, it is not a universal solution for improving the performance of medical image segmentations. In fact, rigorous statistical significance testing and Grad-CAM evidence revealed that supervised learning with medical images provides unique gains in sensitivity and specificity, and both approaches can be used synergistically. We recommend splitting the available images into at least five different proportions and repeating the training and validation of each split five times to identify the optimal distribution of clinical labels to thereby prevent skew and memorization. If clinical label distributions are non-Gaussian (i.e., one class heavily outnumbers the other), randomization is not recommended for achieving optimal performance. For example, in the skin cancer data set, malignant examples were fewer (n = 1,118) compared with benign examples (n = 12,668), and the clinical label distribution was uneven between training and testing. Deep learning models trained using these splits demonstrated high false negatives during segmentation (Fig.~\ref{fig3}). Additionally, we recommend checking for the normalcy of performance metrics and Yeo-Johnson transformations of AUROC, Dice score, and sensitivity and specificity values to select the appropriate parametric Gaussian or non-parametric statistical testings to establish significance. For the non-parametric distributions seen in this study, medians served as better measures of central tendencies than did means. Another finding of this study was instances in which both training schemes, $L_{MI}$ and $T_{II}$, could be used synergistically or as ensemble models to improve both higher-level morphological and fine-grained segmentations (discussed in detail below). This is an important distinction from prior studies, which reported that selection and comparisons between $L_{MI}$ or $T_{II}$ perform better and proposed discarding the inferior model.

The selection of model training procedure (i.e., whether to select transfer learning or train using data native for the intended outcome) depends on not only the size of the data set or overall model AUROC that is most commonly used but also the gradation and diversity of clinical labels and desired binary segmentation specificity and sensitivity unique to the image modality. For several target image classes, $L_{MI}$ and $T_{II}$ models can be used synergistically to achieve a mutually beneficial increase in segmentation accuracies. For skin images, $T_{II}$ models segmented background pixels as disease signatures (false positive) in more images than did $L_{MI}$ models (indicated based on specificity numbers in Table~\ref{tab2}). Thus, the synergistic use of $T_{II}$ models to learn cancer lesions from skin images and $L_{MI}$ models to distinguish background pixels can result in desirable segmentations (\textbf{Fig.~\ref{fig6}}). This was reversed for prostate core biopsy and kidney CT images, in which $T_{II}$ models predict target tumor or organ signatures as background (false negative); thus, for prostate core biopsy and kidney CT images, $T_{II}$ models were better at learning background pixels, while $L_{MI}$ models were superior at learning tumors or organs and can therefore be used in tandem (Fig.~\ref{fig6}). Moreover, we report situations in which one model segments the target ground truth at a different location compared with the other model on the same image. We recommend that the union of output masks from both models could be most beneficial for increasing the segmentation of higher proportions of ground-truth pixels for these images (supplementary image file). One limitation for this method arose when one model ($T_{II}$) demonstrated high false positives while the other ($L_{MI}$) revealed high false negatives (images 4 and 8 in Fig.~\ref{fig6}). Here, the union of model output masks contained false positive regions, while the intersection enriched the false negative pixels (Fig.~\ref{fig6}).

Furthermore, we also report several cases in which the automated segmentation of one image by both models and the use of individual segmentation output masks in various combinations was most optimal for achieving the desired performance (Fig.~\ref{fig6} columns E and G). For comparison, Fig.~\ref{fig6} columns C and D display the individual model performance for $L_{MI}$ and $T_{II}$ respectively. Union, the addition of both segmentation masks (Fig.~\ref{fig6} columns E and F for images 1, 3, 5, 7, 9, 10 and 12), improved the performance when one model ($T_{II}$) exhibited high false negatives while the other model did not. Union operations, however, decreased the performance when one model exhibited high false positives (Fig.~\ref{fig6} columns E and F for image 6). Operations that generated common intersections between segmented region outputs from both models (Fig.~\ref{fig6} columns G and H for images 2, 6 and 11) improved the performance when one model revealed high false positives. Meanwhile, intersection operations were not useful when one model ($L_{MI}$ for image 3 in Fig.~\ref{fig6}) exhibited high false negatives. Our overall results indicate that it was most optimal to use intersection of segmentation outputs when both models show high false positives, and using the union of segmentation outputs when both models show high false negatives. Thus, DL strategies can alternate synergistically for priority in the detection and segmentation of disease regions with requirements when false positives are to be kept at a minimum. This was especially important under low data regimes, in which individual models were found to be inadequate for achieving high sensitivity and specificity. Statistically significant differences for estimating the performance, reproducibility and replicability of $L_{MI}$ and $T_{II}$ models can be used for situations in which sparsity promotes better learning. Thus, even with the purported benefits of transfer learning, DL models might achieve sub-optimal outcomes if model outputs and clinical labels are not carefully examined to provide information causal for medical-grade performance. This can be attributed to the tendency of the transfer learning model to gravitate toward the original data set manifold if sufficiently large and diverse new data is not provided for fine-tuning. For example, $L_{MI}$ model weights were easier to update using specific medical images and outperformed transfer-learned models for binary segmentation specificity for two image types in this study. On the other hand, transfer learning models ($T_{II}$) demonstrate significant performance gains in sensitivity for larger and complex segmentation of multicolored skin cancer RGB images, which share features and possibly complexity with natural world images. Training and generating ensemble $T_{II}$ and $L_{MI}$ models with automated computation of segmentation masks with high sensitivity and specificity for the segmentation of target features can therefore serve as a highly effective strategy.

An open-access GitHub repository hosting containing a) 30 fully trained and validated DL models (15 models each for $L_{MI}$ and $T_{II}$, translating to 10 models for each of the three image modalities used in this study), b) Grad-CAM results and the associated software code and performance estimation from more than 10,000 test images used in this study, and c) a separate statistical analysis package for calculating the AUROC, Dice score, and sensitivity, specificity performance of DL models will be shared as a valuable resource for biomedical and computer science researchers. The data, models, and findings of this study make significant and novel contributions to the optimization of available medical images and clinical labels for the selection, training and validation of explainable DL models with high sensitivity, specificity, and reproducibility of segmentation. Additionally, researchers can use our methods, code, and models for further transfer learning or as starting points for custom applications and to benchmark performance of data sets and explainable DL models of choice.

\section*{Limitations}
Despite using many images from benchmarked data sets and widely used DNN architectures and models, the findings were optimized for the medical images, clinical labels and DNN used in this research. The clinical outcomes of this study were thus limited to the evaluation of prostate core biopsy, skin and 2D slices of volumetric CT images. The numbers of images and labels used in this study were found to be sufficient for accurate binary segmentation but could be increased for additional insights into larger data sets and clinical tasks. Skin cancer and prostate tumor regions annotated by pathologists are often coarse and contain non relevant tissue and skin that can increase disagreements with DL segmentation performance. Additional fine-grained clinical image annotation tools and labeled images may be needed for extremely precise analysis of the results generated by DL models. Performance metrics that include less frequently used DL model architectures suitable for classification and generative tasks were not evaluated in this study. To ensure the uniformity of binary segmentation tasks across models and images, anatomical detection and segmentation of entire kidneys as surrogates for tumors were combined, and Gleason grades and skin cancer types were pooled. Multi-class segmentations (e.g., different prostate tumor grades and skin cancer types) represent future growth areas of this work. Statistical testing for the parametric distribution of results and clinical labels achieved by other data and models may require additional specialized optimization and characterization. Nevertheless, due to the shared representations between medical pictures and unified learning mechanisms of deep CNN architectures, the findings of this study should be generalized to other macro and microscopic images and clinical segmentation tasks.

\section*{Methods}\label{Methods}
\subsection*{Data and preprocessing} Three data sets consisting of macroscopic optical skin (13,786), microscopic RGB prostate core biopsy (244), and CT Digital Imaging and Communications in Medicine (DICOM) images (45,937) were used in this study. RGB images were represented by an n-by-3 data array that defines the red, green, and blue color components for each pixel; the pixel colors are determined based on combinations of the red, green, and blue intensities stored in individual color planes. Skin images from the ISIC (The International Skin Imaging Collaboration) archive~\cite{ref_url1} consist of 13,786 3-channel RGB images of resolution ranging from (7000 $\times$ 4500) to (1024 $\times$ 720) (W $\times$ H). Each image is associated with a ground-truth (GT) binary mask as its label, and white pixels denote regions of the image where skin moles and/or cancer signatures are present (class 1). Meanwhile, black pixels in the binary masks represent background and/or non-disease regions (class 0). Fig.~\ref{fig2} depicts examples of input images and the corresponding ground-truth mask for the ISIC data. Prostate core biopsy RGB images from Gleason2019 (Grand Challenge for Pathology at MICCAI 2019)~\cite{ref_url2} consist of 244 tissue micro-array (TMA) 3-channel RGB images of resolution 5120 $\times$ 5120 (W $\times$ H). Each TMA image was annotated in detail by several expert pathologists for assigning a Gleason tumor grade of 1 to 5 and a GT segmentation mask. Gleason grades 1 and 2 are rare and not associated with tumors and were designated as benign (class 0) black pixels, whereas regions of Gleason Grades 3, 4 and 5 were considered non-benign or tumors~\cite{rana2020use}, represented by white pixels (class 1). Fig.~\ref{fig4} presents examples of input images and the corresponding ground truth masks. The Kidney Tumor Segmentation Grand Challenge 2019 (KiTS19) data set~\cite{ref_data1} consists of 300 randomly selected axial CT volumes of patients who underwent partial or radical nephrectomy for one or more kidney tumors at the University of Minnesota medical center between 2010 and 2018. Each pixel represents a tissue or background and has an assigned gray-scale value between 0 and 255~\cite{athanasiou2017atherosclerotic}, which represents the X-ray beam attenuation to the tissue. Pixels with a value close to 0 (darker pixels) represent structures that have less attenuation to the beam (i.e., soft tissue), while pixels close to 255 (light pixels) represent structures that have high attenuation (i.e., calcification). Each 2D CT slice image represents 1 mm or less of the structure cross section. In this study, CT volumes were processed into 45,424 512 $\times$ 512 (W $\times$ H) 2D slices with each axial slice either with or without kidney (along with kidney tumor) slices, where kidneys (along with the tumor) were associated with GT color masks. Black pixels in the mask represent the background and regions within the slice that are not associated with kidney or tumor tissue. Meanwhile, red pixels represent healthy kidney tissues, and blue pixels represent tumorous kidney tissues. Slices in which kidneys or kidney tumor tissues are not present are associated with black pixels in the mask and were considered to be the background. In this study, we performed binary segmentation of all kidney tissues with or without tumors, the multi-label (color) masks were converted to binary, in which black pixels in the mask correspond to non-kidney regions (class 0) and white-pixels correspond to regions with tumor tissues (class 1). Fig.~\ref{fig5} and Supplementary Fig.~\ref{sfig8} illustrate examples of input images and corresponding ground-truth masks for the kidney CT image data.

Additionally, we examined the clinical class discriminatory performance of the segmentation outputs from $T_{II}$ and $L_{MI}$ models trained using 80:20 data splits for different medical image sub-types (Fig.~\ref{fig1}). These classes are region-based (i.e., a class is determined by the presence or absence of certain clinical features or signatures on an image and the severity of those features in the context of the data set). For example, for skin images, lesions were clinically labeled as benign (12,668 images) or malignant (1,084 images) moles (lesions or tumors), and 19 images were clinically labeled as intermediate. For H $\&$ E stained prostate core biopsy images, background or benign regions of the prostate core were considered healthy or without tumors. Gleason grade 3, 4 or 5 regions were combined to represent a separate clinical label of tumors; this process resulted in 22 images with prostate tumors and 20 healthy images. For the kidney CT DICOM images, the presence or absence of kidneys was considered to be two explicit clinical classes. Before feeding into neural networks, input images were resized to 608 $\times$ 416 (W $\times$ H) for skin image (ISIC) and 608 $\times$ 416 (W $\times$ H) for prostate core biopsy image (Gleason2019) data sets. This was performed to reduce training time and prevent ``out of memory" errors during training. Kidney CT images were used in their original resolution of 512 $\times$ 512 (W $\times$ H).

\subsection*{Deep learning} 
\begin{figure*}
\centering\includegraphics[width=1.0\textwidth]{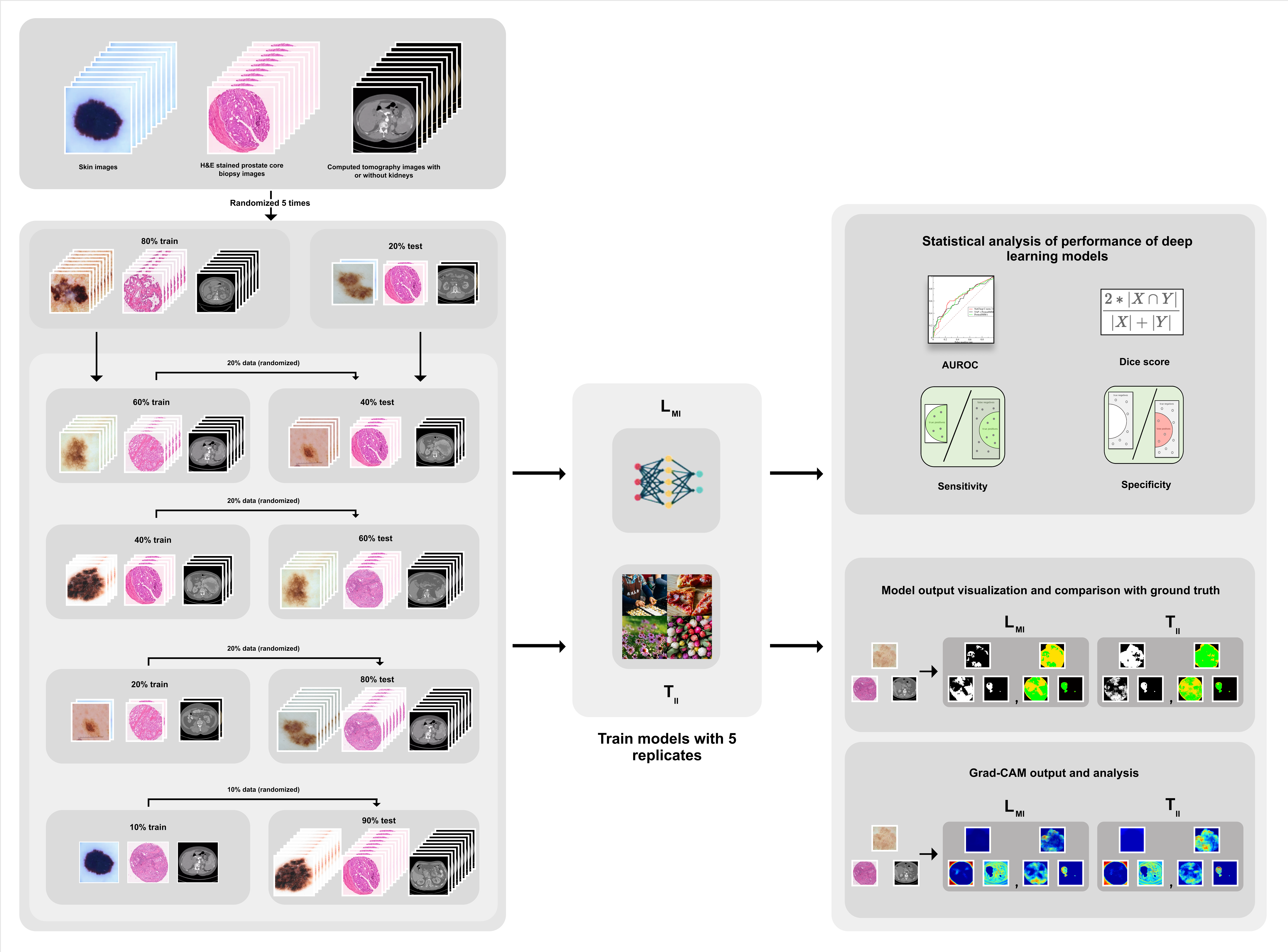}
\caption{}\label{fig1}
\end{figure*}

\begin{figure*}[ht!]
\begin{subfigure}{0.99\textwidth}
\centering\includegraphics[width=1.0\textwidth]{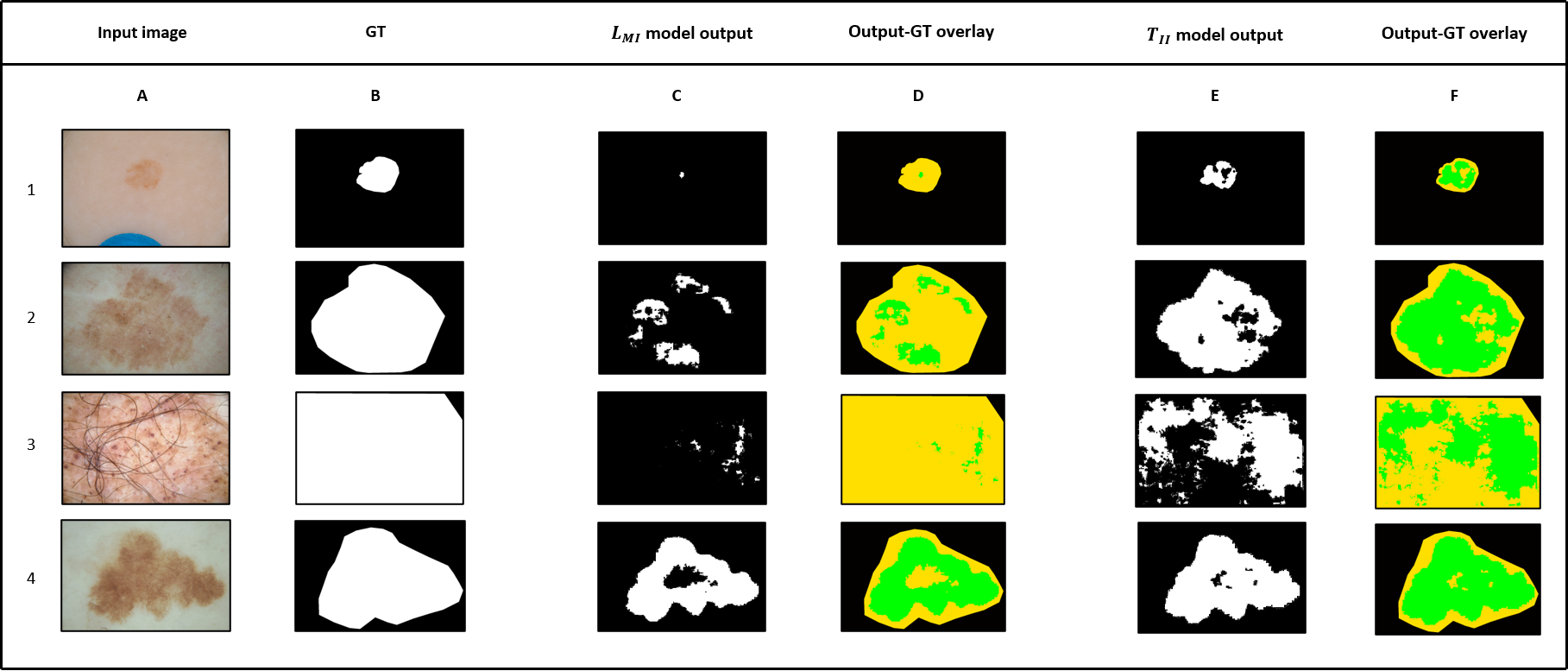}
\caption{}
\end{subfigure}\\
\\
\begin{subfigure}{0.99\textwidth}
\centering\includegraphics[width=0.6\textwidth]{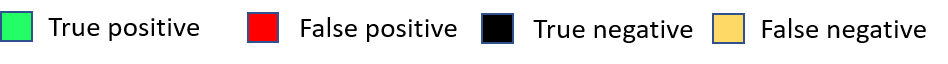}
\end{subfigure}\\
\\
\\
\begin{subfigure}{0.99\textwidth}
\centering\includegraphics[width=1.0\textwidth]{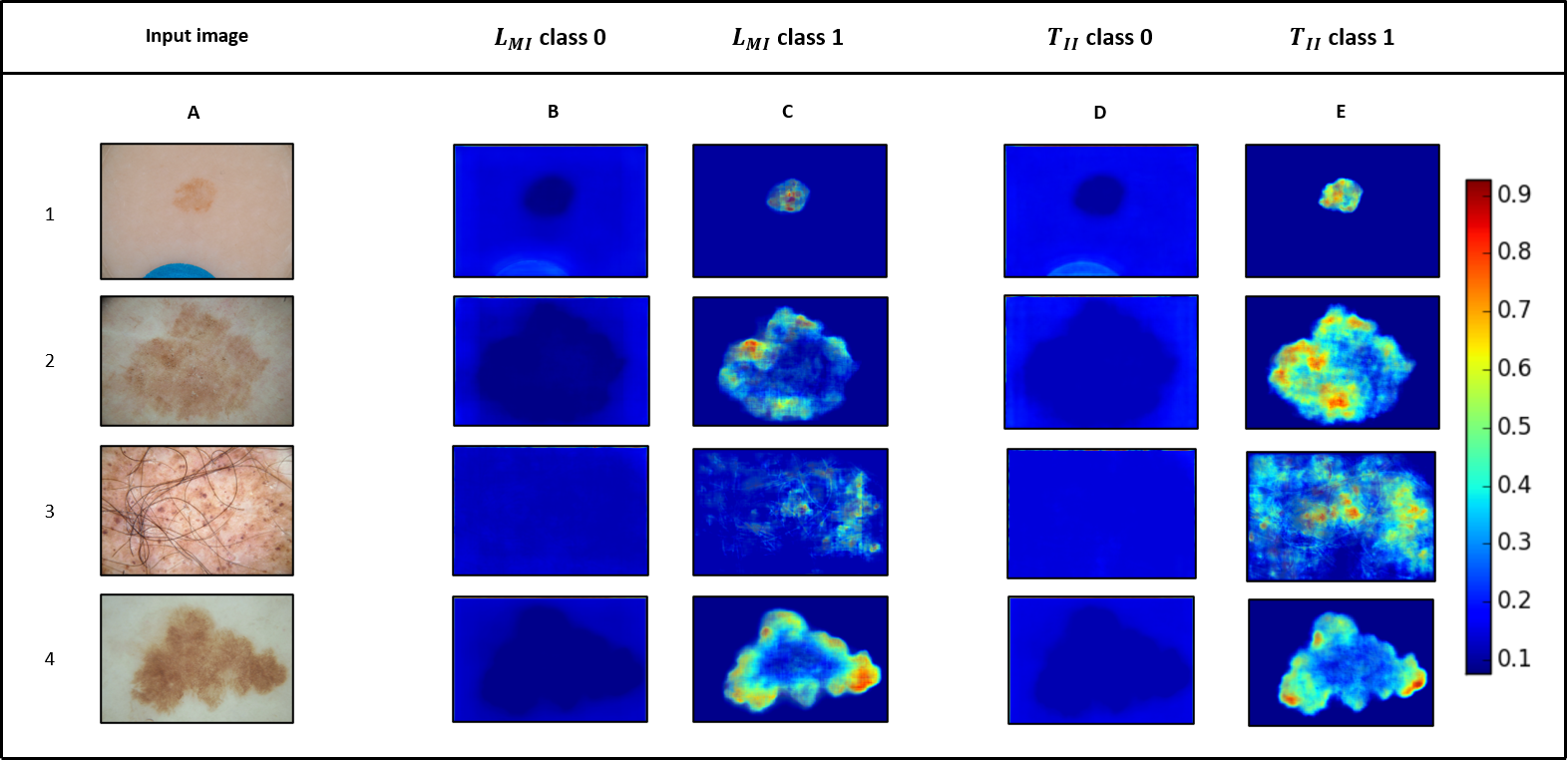}
\caption{}
\end{subfigure}
\caption{}\label{fig2}
\end{figure*}

\begin{figure*}[h!]
\begin{subfigure}{0.99\textwidth}
\centering\includegraphics[width=1.0\textwidth]{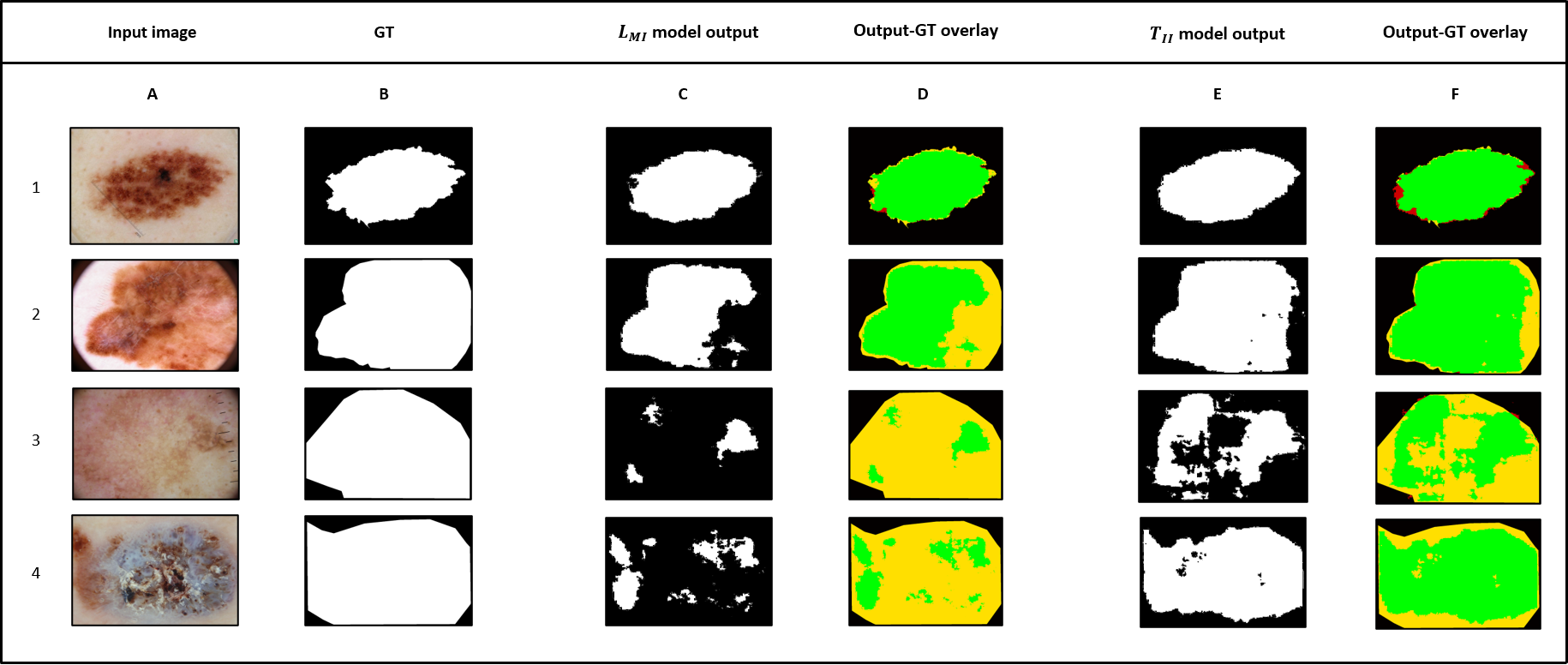}
\caption{}
\end{subfigure}\\
\\
\begin{subfigure}{0.99\textwidth}
\centering\includegraphics[width=0.6\textwidth]{figs/samp_legend.png}
\end{subfigure}\\
\\
\\
\begin{subfigure}{0.99\textwidth}
\centering\includegraphics[width=1.0\textwidth]{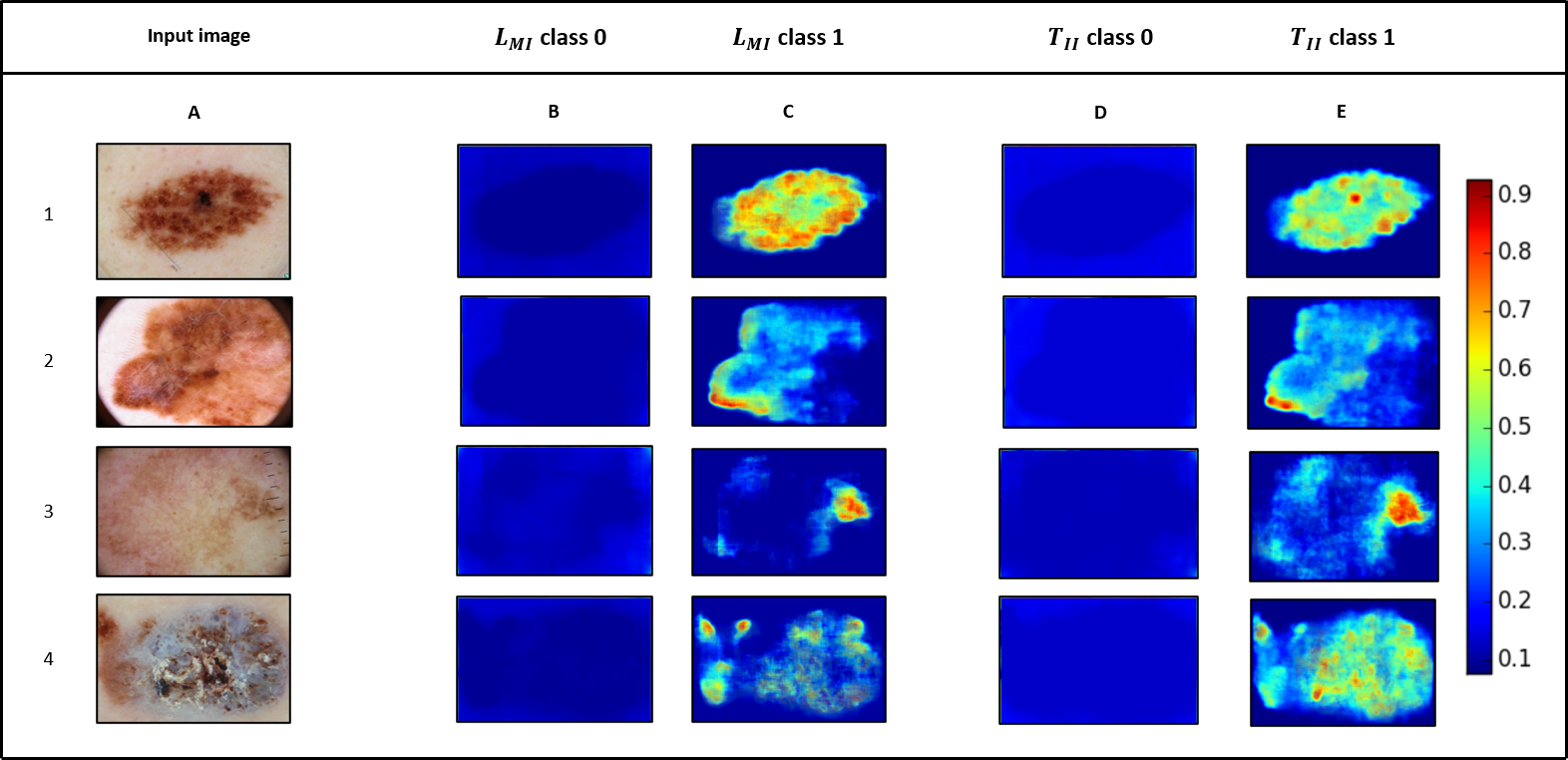}
\caption{}
\end{subfigure}
\caption{}\label{fig3}
\end{figure*}

\begin{figure*}[ht!]
\begin{subfigure}{0.99\textwidth}
\centering\includegraphics[width=1.0\textwidth]{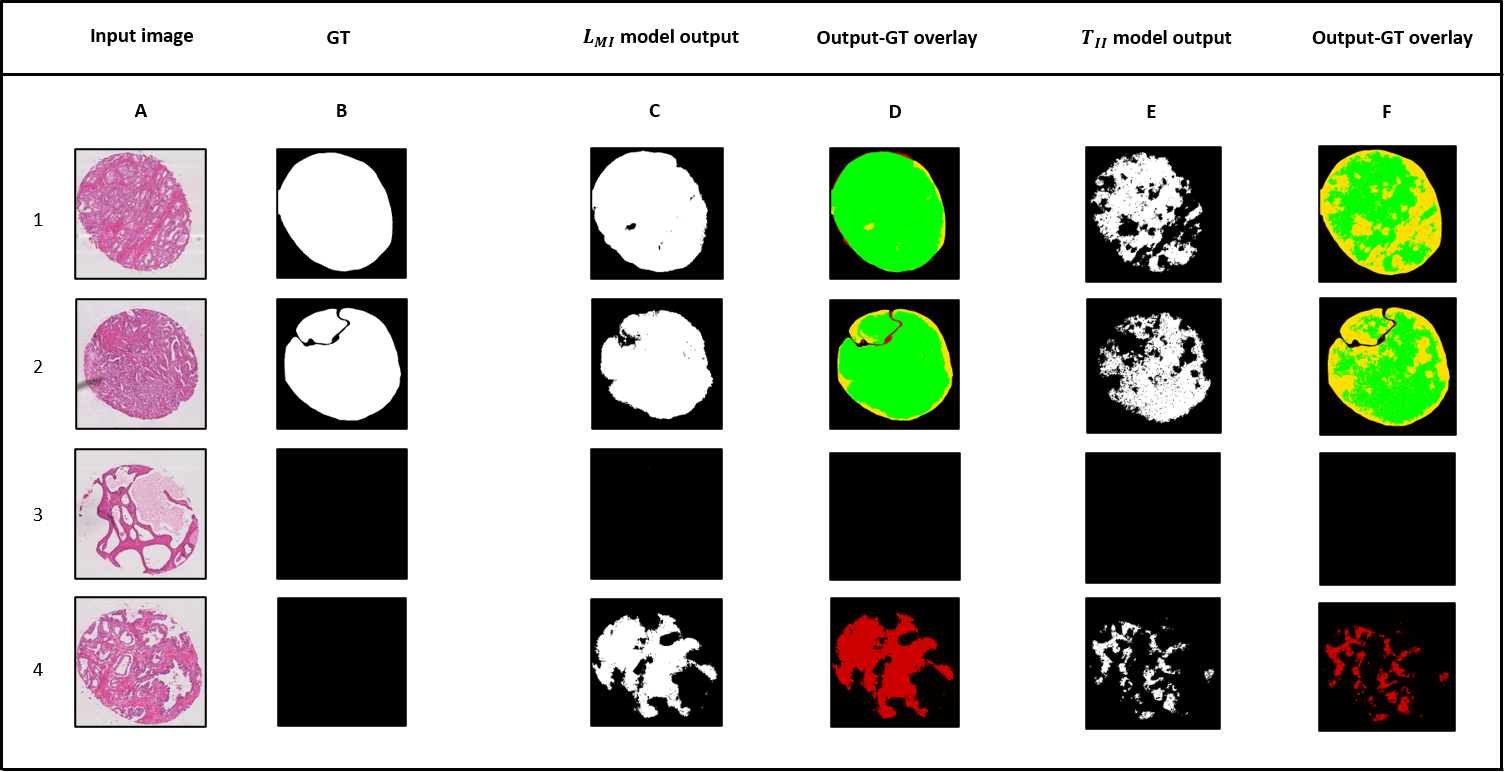}
\caption{}
\end{subfigure}\\
\\
\begin{subfigure}{0.99\textwidth}
\centering\includegraphics[width=0.6\textwidth]{figs/samp_legend.png}
\end{subfigure}\\
\\
\\
\begin{subfigure}{0.99\textwidth}
\centering\includegraphics[width=1.0\textwidth]{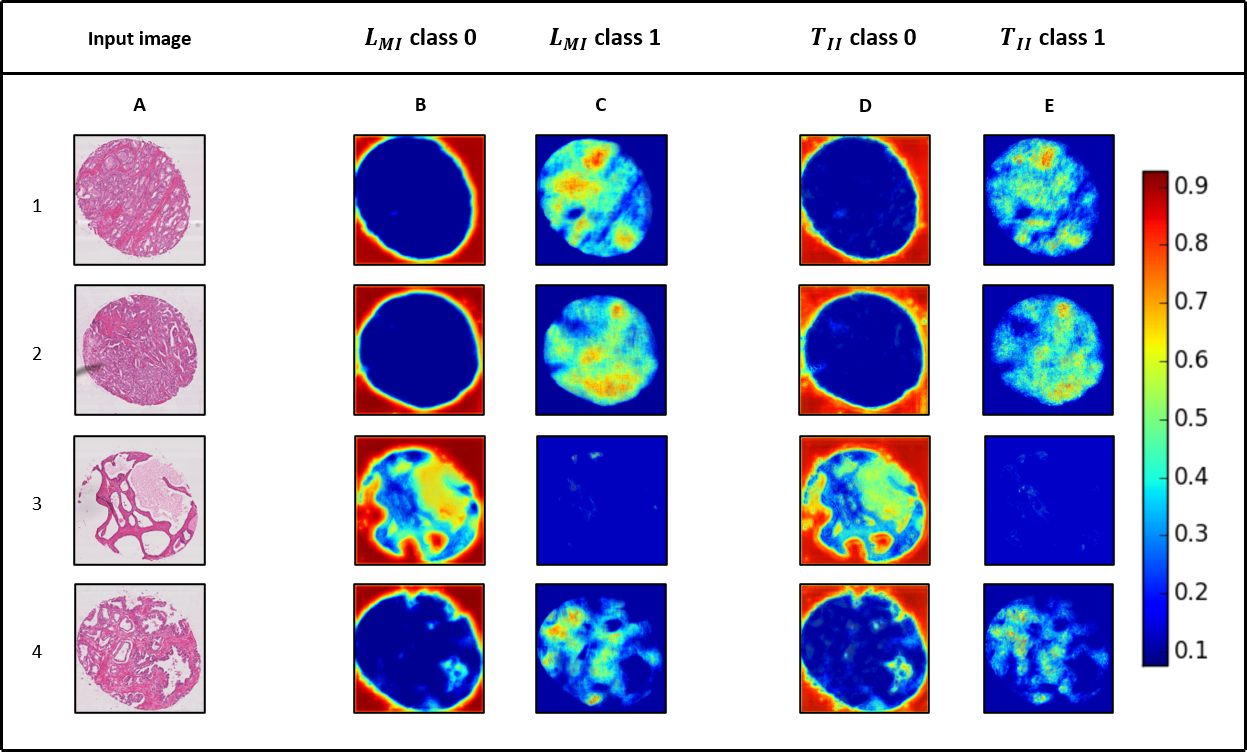}
\caption{}
\end{subfigure}\\
\caption{}\label{fig4}
\end{figure*}

\begin{figure*}[h!]
\begin{subfigure}{0.99\textwidth}
\centering\includegraphics[width=0.75\textwidth]{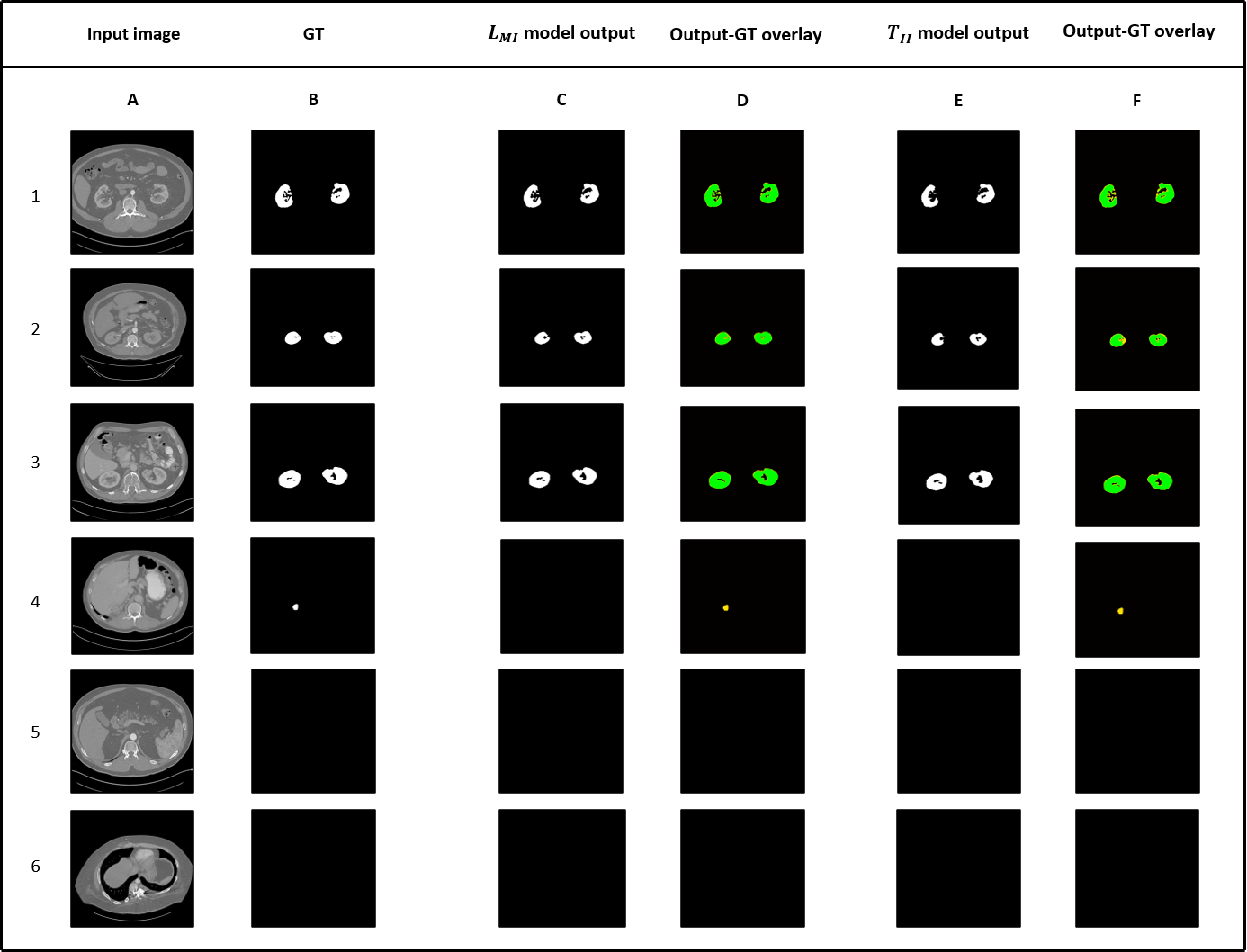}
\caption{}
\end{subfigure}\\
\\
\begin{subfigure}{0.99\textwidth}
\centering\includegraphics[width=0.5\textwidth]{figs/samp_legend.png}
\end{subfigure}
\\
\begin{subfigure}{0.99\textwidth}
\centering\includegraphics[width=0.75\textwidth]{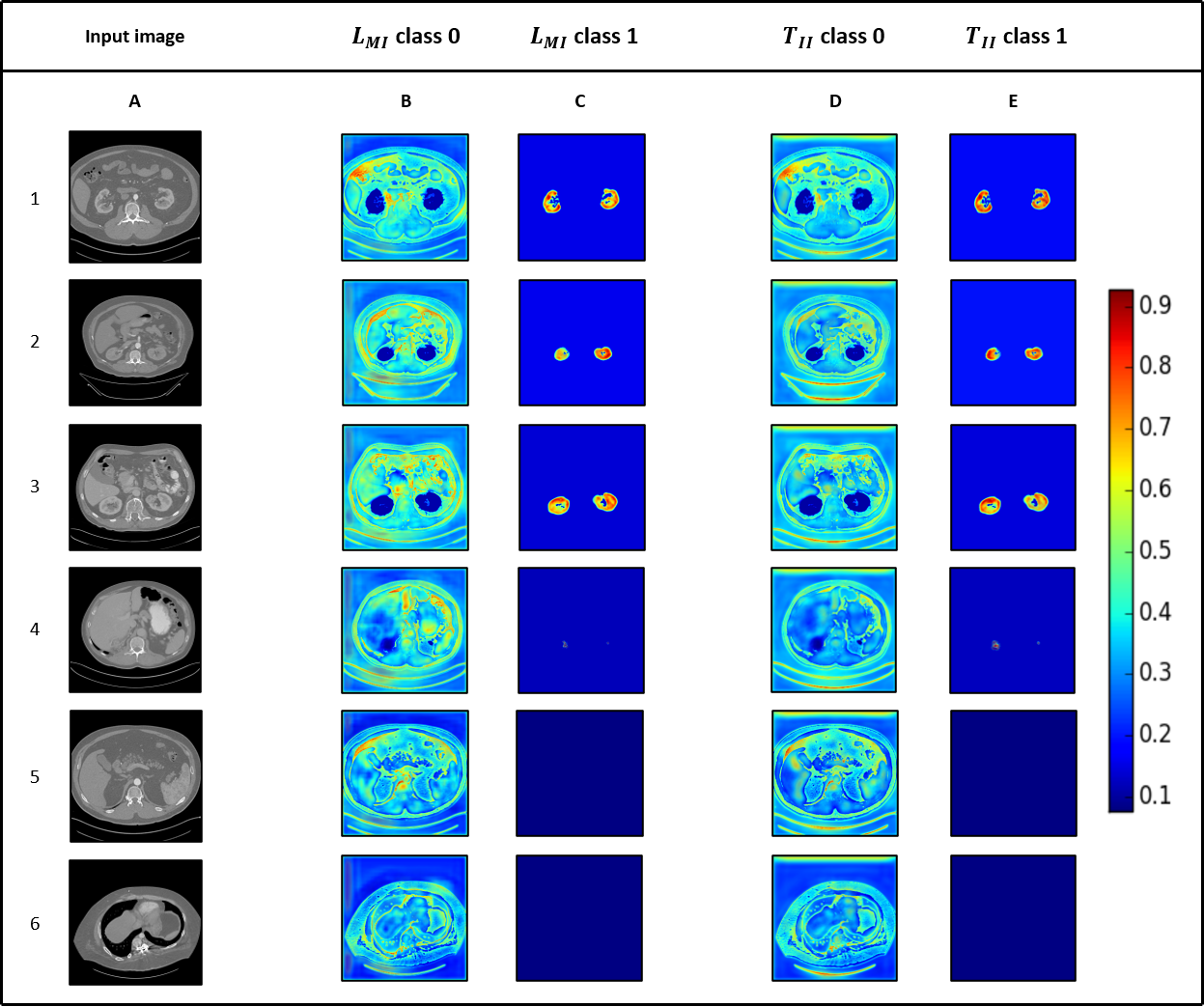}
\caption{}
\end{subfigure}
\caption{} \label{fig5}
\end{figure*}

\begin{figure*}[h!]
\begin{subfigure}{0.99\textwidth}
\centering\includegraphics[width=0.92\textwidth]{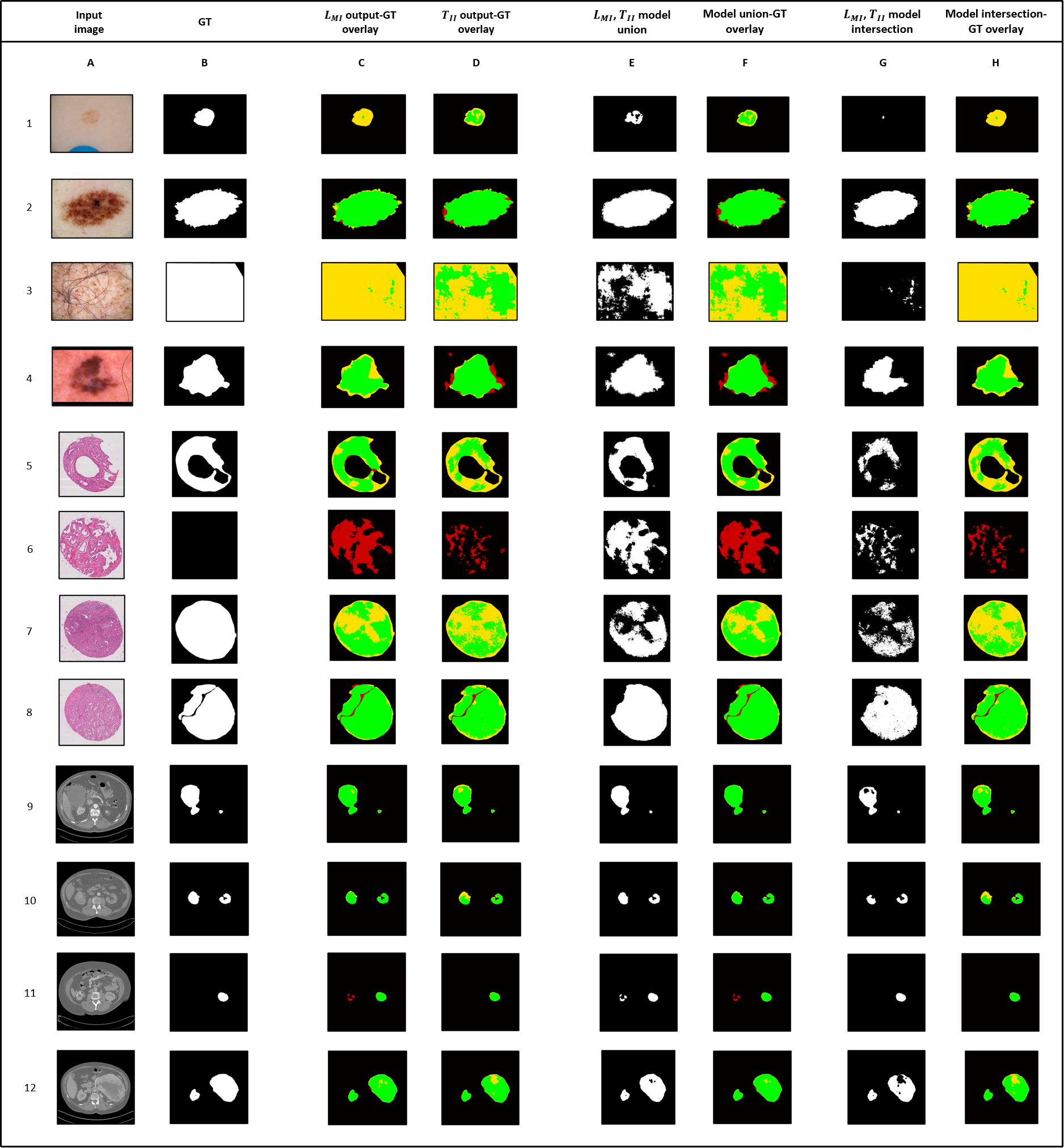}
\end{subfigure}\\
\\
\begin{subfigure}{0.99\textwidth}
\centering\includegraphics[width=0.6\textwidth]{figs/samp_legend.png}
\end{subfigure}
\caption{}\label{fig6}
\end{figure*}

We evaluated several deep architectures suitable for segmentation tasks and selected the well-established and widely used VGG-UNet architecture~\cite{iglovikov2018ternausnet, ronneberger2015u}. The VGG-Unet architecture is an ``encoder-decoder” type framework, in which the encoder is the VGG-16 architecture without the fully-connected (FC) layers and the decoder has subsequent upsampling of previously encoded layers’ output, leading to the final class-dependent predicted segmentation mask. Using the VGG-UNet architecture as the backbone, we trained DL models through ImageNet Initialization ($T_{II}$) that initiates training with pretrained weights from natural world images from the ImageNet database~\cite{ref_url3}. Model Random Initialization ($L_{MI}$) uses the same architecture as $T_{II}$ but was trained without utilizing any pretrained weights (i.e., it is initialized with random weights). All $T_{II}$ and $L_{MI}$ models trained with optimization of the chosen loss function and convergence of learning were found to be 40 epochs for the skin and prostate core biopsy data and 20 epochs for kidney CT images. The categorical cross-entropy loss function for pixel-wise ground-truth-based binary classification and the AdaDelta optimizer with the suggested default learning rate were used~\cite{zeiler2012adadelta}. An NVIDIA GeForce GTX 1080Ti with 12 GB of video memory was used to perform the training and performance evaluation. Five random 80:20 (train:test) splits for each of the three individual image types were trained and tested with internal fivefold repeats using either transfer learning (total of 25 $T_{II}$-models) or $L_{MI}$ (25 $L_{MI}$-models). This random fivefold splitting of the data sets was used for the reallocation of images and for associated clinical labels to avoid overfitting (Table~\ref{stab2}). The resulting five sets of $T_{II}$-$L_{MI}$ models were then compared using statistical testing of the pixel-by-pixel mean, median and standard deviations of AUROC, Dice scores, sensitivity (true positive rate), and specificity (true negative rate) of the associated segmentation masks. A performance difference threshold was defined, such that a difference of 5\% based on the metric values was considered to claim one split (set), or one model performed better than the other. This process was repeated with 25 $L_{MI}$ and $T_{II}$ models (Fig.~\ref{fig1} and Table~\ref{sfig1} for skin image data). Thus, for each of the three individual image modalities 50 DL models were trained and analyzed for performance evaluation of the segmentation of tumors and organs (Fig.~\ref{fig1} and Fig.~\ref{sfig2}). For data depletion experiments, starting with the selected 80:20 (train:test) split, training data was depleted to 60:40, 40:60, 20:80 and 10:90 ratios (Fig.~\ref{fig1} and Table~\ref{tab3}). These subsequent splits were obtained via randomization. Thus, part of this initial training data set was pooled to represent the new test data set which then represented the 40\% test split in the new depleted 60:40 split. Similarly, the next 40:60 (train:test) split was generated similarly from the 60:40 proportions. The depletion was stopped once the train:test split reached a 10:90 ratio.

\subsection*{Evaluation metrics} We used four metrics that are frequently employed to evaluate segmentation models involving binary classes: AUROC, Dice (F1) score, sensitivity (true positive rate), and specificity (true negative rate)~\cite{taha2015metrics,rana2020use}. A segmentation was considered to be a true positive (TP) when tumor tissue pixels were accurately segmented; meanwhile, when non-tumor tissues or background pixels were detected as tumors, the segmentation was considered to be a false positive (FP) or a Type I error. When non-tumor tissues or background pixels were not segmented, the detection was considered to be a true negative (TN). The segmentation of non-tumor tissues or background pixels was considered to be a false negative (FN) or a Type II error. For the calculation of the sensitivity, true positive rate (TPR) or recall = $\frac{TP}{TP + FN}$. In the context of the current study, this denoted the percentage of ground-truth region pixels marked by an expert where the target class is present were segmented correctly as lesion/mole/tumor (skin images), tumor (prostate core biopsy), or organ (kidney CT) by deep learning models. The specificity or true negative rate (TNR) = $\frac{TN}{TN + FP}$ was calculated using the percentage of the DL models’ detection that matched the background or non-target class pixels on ground-truth clinical masks for each of the three data sets. Meanwhile, AUROC for estimating the overall model performance across all test images was calculated from the area of the curve created by plotting TPR against FPR at various threshold settings. The F1 or Dice score is defined as the harmonic mean of precision and sensitivity, where precision = $\frac{TP}{TP + FP}$. In the context of the current study, precision denotes the percentage of the DL models’ predicted segmentation region that matched the ground-truth clinical label. A Dice score = $\frac{2*Precision*Recall}{Precision + Recall}$ was used to measure the segmentation accuracy of the DL models. As outlined below, we propose a new metric based on AUROC, Dice score, sensitivity and specificity for calculating the differences in performances between the two models:

Given the same evaluation metric, $m$ of choice, we define $\Delta_m$ as follows:
\begin{equation}
    \Delta_m = \Delta_{T_{II}-L_{MI}}|_m = S_{m}^{T_{II}} - S_{m}^{L_{MI}}
\end{equation}
where $S$ represents one previously unseen test image or a distribution of images from a particular test data. $m$ represents either AUROC, Dice score, sensitivity or specificity. Thus:
\begin{description}[font=$\bullet$\scshape\bfseries]
    \item $\Delta_m > 0$ $\implies$ $T_{II}$ performs better than $L_{MI}$ for ``m" $\in$ [AUROC, Dice score, sensitivity and specificity].
    \item $\Delta_m = 0$ $\implies$ $T_{II}$ performs as well as $L_{MI}$ for ``m" $\in$ [AUROC, Dice score, sensitivity and specificity].
    \item $\Delta_m < 0$ $\implies$ $L_{MI}$ performs better than $T_{II}$ for ``m" $\in$ [AUROC, Dice score, sensitivity and specificity].
\end{description}

\subsection*{Statistical methods} Binary segmentation masks predicted by individual DL models were used to calculate the AUROC values and Dice scores for each image within a particular test data set. The mean, median and standard deviations (s.d.) of Dice scores and AUROC values associated with individual images segmented by either $L_{MI}$ and $T_{II}$ models were used for performance evaluations. Meanwhile, the overall mean, median and standard deviation of sensitivity and specificity from a particular test data set were used to assess type I (false positive) and type II (false negative) errors for model comparisons used in this study. The Shapiro-Wilk test~\cite{razali2011power} revealed that all four metric distributions deviated from the normality assumptions (Supplementary Fig. S1 and Fig. S2). Yeo-Johnson transformations~\cite{weisberg2001yeo} of the distributions did not achieve normality assumptions (Supplementary Fig.~\ref{sfig1}). Non-parametric Mood's median test~\cite{mood1963introduction,zar2013biostatistical} was used to test the significance between differences of medians of $L_{MI}$ and $T_{II}$ performance values . Furthermore, the null hypothesis for the Mood’s test was that medians of the populations of test images segmented by $L_{MI}$ and $T_{II}$ models were equal (i.e., both models perform equally well based on the segmentation of a certain image and tumor type). The grand median (median of medians) of all the data was then computed. A contingency table was created by classifying the individual AUROC, Dice, sensitivity and specificity values for each test image from a particular data set as being above or below the grand median. This contingency table was also used to compute the test statistic and $p-value$. A $p-value$ of $<$ 0.05 rejected the null hypothesis, and indicated that observations from $L_{MI}$ and $T_{II}$ do not come from the same distribution and exhibit statistically significant differences. This process was used to calculate and compare the means, medians and s.d. of AUROC, Dice score, sensitivity and specificity values for the segmentation of skin, prostate cancer and kidneys by $T_{II}$ and $L_{MI}$ models across all data regimes and images). Based on these comparisons, a 5\% threshold criterion for establishing higher accuracy in predicting clinical labels was used to identify the best-performing 80:20 data split and models. Corresponding output images, segmentation masks and associated $L_{MI}$ and $T_{II}$ models for this 80:20 data split were selected for further analysis. Additionally, for the selected 80:20 split from each of the three image data sets, we calculated and plotted numbers and a non-Gaussian distribution of AUROC and Dice scores (mean, median, and s.d of all test images) was achieved by $T_{II}$ or $L_{MI}$ models (Supplementary Fig.~\ref{sfig2}). A higher Dice score indicates superior segmentation performance for a particular image type and target regions by a trained model~\cite{dice1945measures,taha2015metrics}. Differences between Dice scores were used to rank order individual test images by their best and worst segmentation accuracies achieved by $L_{MI}$ or $T_{II}$ models for comparison (Table~\ref{tab2}).

Representative images from this process were clinically analyzed and Grad-CAM outputs were generated to explain the models' performance. Suitability and comparisons between $T_{II}$ or $L_{MI}$ models for benign and malignant diagnoses from the skin cancer data set are presented in Fig.~\ref{fig2} and Fig.~\ref{fig3}, the prostate core biopsy data set (Fig.~\ref{fig4} and Supplementary Fig.~\ref{sfig7}) and in Fig.~\ref{fig5} and Supplementary Fig.~\ref{sfig8}, for the kidney CT image data set. Subsequently, test images from the selected 80:20 split were assigned clinical diagnoses to evaluate the distribution and performance of key metrics achieved by $L_{MI}$ and $T_{II}$ models for specific clinical outcomes (Supplementary Fig.~\ref{sfig3}, Fig.~\ref{sfig4}, Table~\ref{stab3} and Table~\ref{stab4}). Supplementary Fig.~\ref{sfig5} for the skin cancer data illustrates examples that could not be assigned to a major clinical diagnosis.

\subsection*{Interpretability and explanations of binary segmentation} Grad-CAM, a recently introduced interpretability method, follows the class-activation mapping (CAM) approach for localization~\cite{zhou2016learning}, and enables the modification of neural network architectures that perform stand-alone classification or classification based on segmentation. During a Grad-CAM operation, fully connected (FC) layers are replaced by convolutional layers. The subsequent global-average pooling~\cite{lin2013network} yields class-specific feature maps. This method uses a gradient corresponding to a certain target class that is fed into the final convolutional layer of a network to produce an approximate localization (heat) map of the important regions in the image for each target class. For both $L_{MI}$ and $T_{II}$ models, an input image is chosen after which the input image is passed through the model architecture with the trained weights. Grad-CAM then operates on this trained model and input image pair along with a target class (``0" for background and ``1" for lesion or mole in skin cancer, tumor for prostate and kidney tissues for the kidney CT image data set) to generate a Grad-CAM heatmap. Since there are five trained models for five runs, each image in the test data set has five corresponding Grad-CAM outputs for $L_{MI}$ and $T_{II}$ and for each target class. These five outputs were then averaged to generate the final Grad-CAM heatmap. The $L_{MI}$ and $T_{II}$ model outputs for the qualitative samples are presented in Fig.~\ref{fig2} to Fig.~\ref{fig6} and Fig.~\ref{sfig5} to Fig.~\ref{sfig8} display the intersection of the outputs of the five individual runs. They are obtained as follows:

\begin{equation}
    MO = \bigcap_{n=1}^5 mo_{Run_n}
\end{equation}

for both models, $L_{MI}$ and $T_{II}$ where $MO$ is the final model output and $mo_{Run_n}$ is the output for the $n^{th}$ run. We calculate the averages of the Grad-CAM outputs of the five runs; thus, the final outputs for both models, $L_{MI}$ and $T_{II}$ where $GC_{Run_n}$ is the Grad-CAM output for $Run_n$, were calculated as follows:

\begin{equation}
    GC_{Out} = \frac{1}{5}\mathlarger{\sum}_{n=1}^{5} GC_{Run_n}
\end{equation}

\section*{Data and code availability}
Upon publication, all data, trained models and associated code will be made available at a relevant GitHub repository upon publication. The prepared link will also host the model segmentation and Grad-CAM outputs for the $L_{MI}$ and $T_{II}$ models for the 2758 test images for skin, 49 test images for prostate core biopsy and 9085 images for the kidney CT data. The three data sets used in this study can be obtained from:~\url{https://www.isic-archive.com/#!/topWithHeader/wideContentTop/main} for the skin images, \texttt{https://gleason2019.grand-challenge.org/} for the prostate core biopsy images and \texttt{https://kits19.grand-challenge.org/} for the kidney CT images. All outputs from this project including code, data, figures, documentation, and manuscripts will be made publicly available under an open license (e.g. MIT open source). Codes will be released under a MIT license (\url{https://opensource.org/licenses/MIT}) and a permissive open source software license. Data will be released under the Creative Commons Public Domain Dedication (CC0, version 1.0 or later).

\section*{Author contribution}
Sambuddha Ghosal and Pratik Shah designed the study, computational framework and analysed data and results. Sambuddha Ghosal carried out implementation of the research. Sambuddha Ghosal and Pratik Shah wrote the manuscript and made figures. Pratik Shah led conception, supervision and planning of the research.

\clearpage
\newpage

\noindent
\textbf{Figure 1: Overview of study design and deep learning models.} Three data sets consisting of macroscopic optical skin (n $=13,786$), microscopic RGB prostate core biopsy (n $=244$), and CT Digital Imaging and Communications in Medicine (DICOM) images (n $=45,937$) were randomized and split five times into different percentages of 80 (training) and 20 (validation). Each of the five 80:20 splits (sets) for three individual image types were then used for training a VGG-UNet deep learning model with internal fivefold repeats by transfer learning with pretrained weights from 14 million natural world images with ImageNet initialization ($T_{II}$) or training with only medical images ($L_{MI}$) in a particular data set. Resulting five sets of $T_{II}$ ($n=25$) and $L_{MI}$ ($n=25$) models were then compared using statistical testing of the pixel-by-pixel mean, median and standard deviations of AUROC, Dice scores, sensitivity (true positive rate), and specificity (true negative rate) of the associated segmentation masks. This process was repeated for estimating DL model performance following depletion of training data to smaller proportions.
\newline
\\
\textbf{Figure 2:} Visualization and explanation of transfer learning ($T_{II}$) and learning from medical images ($L_{MI}$) models on the segmentation of images with benign skin cancer. \textbf{Panel (i)} left to right columns: A, input RGB image; B, binary mask of clinical ground truth label; C, binary mask of output image after binary segmentation via $L_{MI}$ model; D, overlay of clinical and $L_{MI}$ model binary output masks; E, binary mask of output image after binary segmentation via $T_{II}$ model; and F, overlay of clinical and $T_{II}$ model binary output masks. Green, true positive (TP); black, true negative (TN); red, false positive (FP); and yellow, false negative (FN). \textbf{Panel (ii)} Target class-based Grad-CAM output. Class 0 represents the background or non-tumor pixel regions of the skin. Class 1 represents pixels with benign tumors, moles and lesions. Color bar represents the degree of model attention and importance with deeper red indicating the most importance and deeper blue indicating the least. ``GT” indicates clinical ground truth.
\newline
\newpage
\noindent
\textbf{Figure 3:} Visualization and explanation of transfer learning ($T_{II}$) and learning from medical images ($L_{MI}$) models on the segmentation of images with malignant skin cancer. \textbf{Panel (i)} left to right columns; A, input RGB image; B, binary mask of clinical ground truth label; C, binary mask of output image after binary segmentation via $L_{MI}$ model; D, overlay of clinical and $L_{MI}$ model binary output masks; E, binary mask of output image after binary segmentation via $T_{II}$ model; and F, overlay of clinical and $T_{II}$ model binary output masks. Green, true positive (TP); black, true negative (TN); red, false positive (FP); and yellow, false negative (FN). \textbf{Panel (ii)} Target class-based Grad-CAM output. Class 0 represents the background or non-tumor pixel regions of the skin. Class 1 represents pixels with benign tumors, moles and lesions. Color bar represents the degree of model attention and importance with deeper red indicating the most importance and deeper blue indicating the least. ``GT” indicates clinical ground truth.
\newline
\\
\textbf{Figure 4:} Visualization and explanation of transfer learning ($T_{II}$) and learning from medical images ($L_{MI}$) models on the segmentation of prostate core biopsy images. \textbf{Panel (i)} left to right columns; A, input RGB image; B, binary mask of clinical ground truth label; C, binary mask of output image after binary segmentation via $L_{MI}$ model; D, overlay of clinical and $L_{MI}$ model binary output masks; E, binary mask of output image after binary segmentation via $T_{II}$ model; and F, overlay of clinical and $T_{II}$ model binary output masks. Green, true positive (TP); black, true negative (TN); red, false positive (FP); and yellow, false negative (FN). \textbf{Panel (ii)} Target class-based Grad-CAM output. Class 0 represents the background or non-tumor pixel regions of the prostate core biopsy. Class 1 represents Gleason grade 3, 4 or 5 tumors. Color bar in this panel represents the degree of model attention and importance with deeper red indicating the most importance and deeper blue indicating the least. ``GT” indicates clinical ground truth.
\newline
\newpage
\noindent
\textbf{Figure 5:} Visualization and explanation of transfer learning ($T_{II}$) and learning from medical images ($L_{MI}$) models on the binary segmentation of additional computed tomography images with kidneys. \textbf{Panel (i)} left to right columns; A, input RGB image; B, binary mask of clinical ground truth label; C, binary mask of output image after binary segmentation via $L_{MI}$ model; D, overlay of clinical and $L_{MI}$ model binary output masks; E, binary mask of output image after segmentation via $T_{II}$ model; and F, overlay of clinical and $T_{II}$ model binary output masks. Green, true positive (TP); black, true negative (TN); red, false positive (FP); and yellow, false negative (FN). \textbf{Panel (ii)} Target class-based Grad-CAM output. Class 0 represents the background or non-kidney class or region pixels of the computed tomography image. Class 1 represents the kidney tissue pixels. Color bar in this panel represents the degree of model attention and importance with deeper red indicating the most importance and deeper blue indicating the least. ``GT” indicates clinical ground truth.
\newline
\\
\textbf{Figure 6:} Visualization of synergistic outputs (unions and intersections) of transfer learning ($T_{II}$) and learning from medical images ($L_{MI}$). Left to right columns; A, input RGB image; B, binary mask of clinical ground truth label; C, overlay of clinical and $L_{MI}$ model binary output masks; D, overlay of clinical and $T_{II}$ model binary output masks; E, binary mask of output image after union of binary segmentations via $L_{MI}$ and $T_{II}$ models; F, overlay of clinical and binary union output masks; G, binary mask of output image after intersection of binary segmentations via $L_{MI}$ and $T_{II}$ models; and H, overlay of clinical and binary intersection output masks. Green, true positive (TP); black, true negative (TN); red, false positive (FP); and yellow, false negative (FN). ``GT” indicates clinical ground truth.

\newpage
\newcommand{\ra}[1]{\renewcommand{\arraystretch}{#1}}
\begin{table*}
\centering
\caption{Grand median, mean of medians, and standard deviations (sd.) of distributions of five replicates of median values of binary segmentation of RGB images with skin cancer and microscopic Hematoxylin $\&$ Eosin (H $\&$ E) stained prostate core biopsy and computed tomography (CT) of kidneys. Area under the receiver operating curve (AUROC), Dice score, sensitivity, and specificity achieved by transfer learning ($T\_{II}$) and learning from medical images ($L\_{MI}$) deep learning models are reported. Statistically significant ($p < 0.05$) differences in distributions calculated by Mood’s median test are indicated by \textbf{*}. Differences greater than 5\% are indicated by ``$\textbf{\textdagger}$" for the better-performing model. A value of 1 indicates a perfect score.}\label{tab1}
\ra{1.3}
\resizebox{\textwidth}{!}{
\begin{tabular}{@{}rccrccrcc@{}}
\toprule
& \multicolumn{2}{c}{Skin} & \phantom{abc} & \multicolumn{2}{c}{Prostate core biopsy} & \phantom{abc} & \multicolumn{2}{c}{Kidney CT}\\
& $L\_{MI}$ & $T\_{II}$ && $L\_{MI}$ & $T\_{II}$ && $L\_{MI}$ & $T\_{II}$ \\
\midrule
\textbf{AUROC}\\
$median$ & $0.8544^*$ & $0.9282^*\textsuperscript{\textdagger}$ && 0.9359 & 0.9337 && $0.9980^*$ & $0.9978^*$\\
$mean$ & 0.8120 & $0.8826\textsuperscript{\textdagger}$ && 0.9083 & 0.8991 && 0.9876 & 0.9871\\
$sd.$ & 0.1514 & 0.1277 && 0.0831 & 0.0852 && 0.0588 & 0.0587\\
\textbf{Dice Score}\\
$median$ & $0.8273^*$ & $0.8857^*\textsuperscript{\textdagger}$ && 0.9476 & 0.9325 && 0.9597 & 0.9598\\
$mean$ & 0.7483 & $0.8250\textsuperscript{\textdagger}$ && 0.8858 & 0.8733 && 0.9509 & 0.9502\\
$sd.$ & 0.2169 & 0.1786 && 0.1536 & 0.1512 && 0.0657 & 0.0622\\
\textbf{Sensitivity}\\
$median$ & $0.7156^*$ & $0.8891^*\textsuperscript{\textdagger}$ && $0.9520^*$ & $0.9059^*$ && $0.9985^*$ & $0.9979^*$\\
$mean$ & 0.6331 & $0.7944\textsuperscript{\textdagger}$ && 0.8988 & 0.8507 && 0.9772 & 0.9761\\
$sd.$ & 0.3080 & 0.2541 && 0.1438 & 0.1599 && 0.1178 & 0.1177\\
\textbf{Specificity}\\
$median$ & $0.9999^*$ & $0.9986^*$ && 0.9572 & 0.9761 && 1 & 1\\
$mean$ & 0.9922 & 0.9722 && 0.9108 & 0.9490 && 0.9993 & 0.9993\\
$sd.$ & 0.0341 & 0.0945 && 0.1056 & 0.0777 && 0.0011 & 0.0011\\
\\
\bottomrule
\end{tabular}
}
\end{table*}

\newpage
\begin{sidewaystable*}
\centering
\caption{Number of test images that achieved higher median values from five replicates for binary segmentation by transfer learning ($T_{II}$) or learning from medical images ($L_{MI}$) deep learning models. Area under the receiver operating curve (AUROC), Dice score, sensitivity, and specificity of RGB images with skin cancer and microscopic Hematoxylin $\&$ Eosin (H $\&$ E) stained prostate core biopsy and computed tomography (CT) of kidneys are reported following the subtraction of $T_{II}$ and $L_{MI}$ scores indicated by $\Delta_m$ >, < and = 0 for each image. Numbers of the test images that achieved metric values of 0.9 and higher for each model are displayed. $n_{test}$ represents the number of test images in each data set. The better-performing model is indicated by ``$\textbf{\textdagger}$".}\label{tab2}
\ra{3.0}
\resizebox{\textwidth}{!}{
\begin{tabular}{@{}rcccccrcccccrccccc@{}}\toprule
& \multicolumn{5}{c}{Skin ($n_{test} = 2758$)} & \phantom{abc} & \multicolumn{5}{c}{Prostate core biopsy ($n_{test} = 49$)} & \phantom{abc} & \multicolumn{5}{c}{Kidney CT ($n_{test} = 9085$)}\\
\midrule
& \phantom{abc} & \phantom{abc} & \phantom{abc} & \multicolumn{2}{c}{$value > 0.9$} && \phantom{abc} & \phantom{abc} & \phantom{abc} & \multicolumn{2}{c}{$value > 0.9$} && \phantom{abc} & \phantom{abc} & \phantom{abc} & \multicolumn{2}{c}{$value > 0.9$}\\
Metric & $\Delta_m > 0$ (\%) & $\Delta_m = 0$ (\%) & $\Delta_m < 0$ (\%) & $L_{MI}$ & $T_{II}$ && $\Delta_m > 0$ (\%) & $\Delta_m = 0$ (\%) & $\Delta_m < 0$ (\%) & $L_{MI}$ & $T_{II}$ && $\Delta_m > 0$ (\%) & $\Delta_m = 0$ (\%) & $\Delta_m < 0$ (\%) & $L_{MI}$ & $T_{II}$\\\midrule
AUROC & 2384 (86)\textsuperscript{\textdagger} & 103 (4) & 271 (10) & 984 & 1651\textsuperscript{\textdagger} && 15 (31) & 2 (4) & 32 (65) & 34 & 29 && 774 (9) & 6374 (70) & 1937 (21)\textsuperscript{\textdagger} & 3167 & 3163\\
Dice Score & 2171 (79)\textsuperscript{\textdagger} & 85 (3) & 502 (18) & 671 & 1177\textsuperscript{\textdagger} && 15 (31) & 2 (4) & 32 (65) & 32 & 29 && 1614 (18) & 6096 (67) & 1375 (15) & 3086 & 3074\\
Sensitivity & 2559 (86)\textsuperscript{\textdagger} & 140 (5) & 59 (2) & 642 & 1301\textsuperscript{\textdagger} && 2 (4) & 2 (4) & 45 (92)\textsuperscript{\textdagger} & 36 & 27 && 804 (9) & 6357 (70) & 1924 (21)\textsuperscript{\textdagger} & 3116 & 3043\\
Specificity & 57 (2) & 589 (21) & 2112 (23)\textsuperscript{\textdagger} & 2707 & 2576 && 47 (96)\textsuperscript{\textdagger} & 1 (2) & 1 (2) & 36 & 42 && 1397 (15)\textsuperscript{\textdagger} & 7051 (78) & 637 (7) & 9085 & 9085\\
\bottomrule
\end{tabular}
}
\end{sidewaystable*}

\newpage
\begin{table*}
\centering
\caption{Low data regimen experiments for the segmentation of medical images by deep learning. Images used for training transfer learning ($T_{II}$) and learning from medical images ($L_{MI}$) deep learning models were depleted into different proportions (indicated by \%). Median of median values of the binary segmentation of RGB images with skin cancer and microscopic Hematoxylin $\&$ Eosin (H $\&$ E) stained prostate core biopsy and computed tomography (CT) of kidneys are displayed. Area under the receiver operating curve (AUROC), Dice scores, and sensitivity from five replicates of individual models are reported. Statistically significant ($p < 0.05$) differences in distributions calculated by Mood’s median test are indicated by \textbf{*}. Differences greater than 5\% are indicated by ``$\textbf{\textdagger}$" for the better-performing model. A value of 1 indicates a perfect score.}\label{tab3}
\ra{1.25}
\resizebox{\textwidth}{!}{
\begin{tabular}{@{}rccrccrcc@{}}\toprule
& \multicolumn{2}{c}{Skin} & \phantom{abc} & \multicolumn{2}{c}{Prostate core biopsy} & \phantom{abc} & \multicolumn{2}{c}{Kidney CT}\\
& $L_{MI}$ & $T_{II}$ && $L_{MI}$ & $T_{II}$ && $L_{MI}$ & $T_{II}$ \\ \midrule
\textbf{AUROC}\\
$10\%$ & $0.8187^*$ & $0.8769^*\textsuperscript{\textdagger}$ && $0.8730^*$ & $0.9158^*$ && $0.9967^*$ & $0.9963^*$\\
$20\%$ & $0.8160^*$ & $0.9126^*\textsuperscript{\textdagger}$ && 0.9185 & 0.9237 && $0.9972^*$ & $0.9973^*$\\
$40\%$ & $0.8613^*$ & $0.8932^*$ && 0.9215 & 0.9183 && 0.9976 & 0.9976\\
$60\%$ & $0.8542^*$ & $0.9044^*\textsuperscript{\textdagger}$ && 0.9153 & 0.8995 && $0.9980^*$ & $0.9976^*$\\
$80\%$ & $0.8544^*$ & $0.9282^*\textsuperscript{\textdagger}$ && 0.9359 & 0.9337 && 0.9980 & $0.9978^*$\\

\textbf{Dice-Score}\\
$10\%$ & $0.7818^*$ & $0.8437^*\textsuperscript{\textdagger}$ && 0.9246 & 0.9250 && $0.9533^*$ & $0.9541^*$\\
$20\%$ & $0.7855^*$ & $0.8743^*\textsuperscript{\textdagger}$ && 0.9268 & 0.9392 && $0.9551^*$ & $0.9543^*$\\
$40\%$ & $0.8346^*$ & $0.8624^*$ && 0.9390 & 0.9302 && 0.9571 & 0.9567\\
$60\%$ & $0.8274^*$ & $0.8721^*$ && 0.9262 & 0.9030 && $0.9570^*$ & $0.9577^*$\\
$80\%$ & $0.8273^*$ & $0.8857^*$ && 0.9476 & 0.9325 && 0.9597 & 0.9598\\

\textbf{Sensitivity}\\
$10\%$ & $0.6571^*$ & $0.7769^*\textsuperscript{\textdagger}$ && $0.9523^*$ & $0.9394^*$ && $0.9959^*$ & $0.9950^*$\\
$20\%$ & $0.6449^*$ & $0.8593^*\textsuperscript{\textdagger}$ && 0.9567 & 0.9559 && $0.9968^*$ & $0.9971^*$\\
$40\%$ & $0.7341^*$ & $0.8195^*\textsuperscript{\textdagger}$ && $0.9726^*$ & $0.9535^*$ && 0.9976 & 0.9975\\
$60\%$ & $0.7174^*$ & $0.8321^*\textsuperscript{\textdagger}$ && $0.9407^*$ & $0.8919^*$ && $0.9983^*$ & $0.9974^*$\\
$80\%$ & $0.7156^*$ & $0.8891^*\textsuperscript{\textdagger}$ && $0.9520^*$ & $0.9059^*$ && $0.9985^*$ & $0.9979^*$\\

\textbf{Specificity}\\
$10\%$ & $1^*$ & $0.9995^*$ && 0.9223 & 0.8960 && 1 & 1\\
$20\%$ & $1^*$ & $0.9990^*$ && 0.9203 & 0.8768 && 1 & 1\\
$40\%$ & $0.9999^*$ & $0.9995^*$ && 0.8987 & 0.8971 && 1 & 1\\
$60\%$ & $0.9999^*$ & $0.9993^*$ && 0.9547 & 0.9657 && 1 & 1\\
$80\%$ & $0.9999^*$ & $0.9986^*$ && 0.9572 & 0.9761 && 1 & 1\\
\bottomrule
\end{tabular}
}
\end{table*}


\clearpage

\begin{thebibliography}{52}
\expandafter\ifx\csname natexlab\endcsname\relax\def\natexlab#1{#1}\fi
\providecommand{\bibinfo}[2]{#2}
\ifx\xfnm\relax \def\xfnm[#1]{\unskip,\space#1}\fi
\bibitem[{Suzuki(2017)}]{suzuki2017overview}
\bibinfo{author}{K.~Suzuki},
\newblock \bibinfo{title}{Overview of deep learning in medical imaging},
\newblock \bibinfo{journal}{Radiological physics and technology}
  \bibinfo{volume}{10} (\bibinfo{year}{2017}) \bibinfo{pages}{257--273}.
\bibitem[{Yauney et~al.(2017)Yauney, Angelino, Edlund, and
  Shah}]{yauney2017convolutional}
\bibinfo{author}{G.~Yauney}, \bibinfo{author}{K.~Angelino},
  \bibinfo{author}{D.~Edlund}, \bibinfo{author}{P.~Shah},
\newblock \bibinfo{title}{Convolutional neural network for combined
  classification of fluorescent biomarkers and expert annotations using white
  light images},
\newblock in: \bibinfo{booktitle}{2017 IEEE 17th International Conference on
  Bioinformatics and Bioengineering (BIBE)}, \bibinfo{organization}{IEEE}, pp.
  \bibinfo{pages}{303--309}.
\bibitem[{Rana et~al.(2020)Rana, Lowe, Lithgow, Horback, Janovitz, Da~Silva,
  Tsai, Shanmugam, Bayat, and Shah}]{rana2020use}
\bibinfo{author}{A.~Rana}, \bibinfo{author}{A.~Lowe},
  \bibinfo{author}{M.~Lithgow}, \bibinfo{author}{K.~Horback},
  \bibinfo{author}{T.~Janovitz}, \bibinfo{author}{A.~Da~Silva},
  \bibinfo{author}{H.~Tsai}, \bibinfo{author}{V.~Shanmugam},
  \bibinfo{author}{A.~Bayat}, \bibinfo{author}{P.~Shah},
\newblock \bibinfo{title}{Use of deep learning to develop and analyze
  computational hematoxylin and eosin staining of prostate core biopsy images
  for tumor diagnosis},
\newblock \bibinfo{journal}{JAMA Network Open} \bibinfo{volume}{3}
  (\bibinfo{year}{2020}) \bibinfo{pages}{e205111--e205111}.
\bibitem[{Javia et~al.(2018)Javia, Rana, Shapiro, and Shah}]{javia2018machine}
\bibinfo{author}{P.~Javia}, \bibinfo{author}{A.~Rana},
  \bibinfo{author}{N.~Shapiro}, \bibinfo{author}{P.~Shah},
\newblock \bibinfo{title}{Machine learning algorithms for classification of
  microcirculation images from septic and non-septic patients},
\newblock in: \bibinfo{booktitle}{2018 17th IEEE International Conference on
  Machine Learning and Applications (ICMLA)}, \bibinfo{organization}{IEEE}, pp.
  \bibinfo{pages}{607--611}.
\bibitem[{Krizhevsky et~al.(2012)Krizhevsky, Sutskever, and
  Hinton}]{krizhevsky2012imagenet}
\bibinfo{author}{A.~Krizhevsky}, \bibinfo{author}{I.~Sutskever},
  \bibinfo{author}{G.~E. Hinton},
\newblock \bibinfo{title}{Imagenet classification with deep convolutional
  neural networks},
\newblock in: \bibinfo{booktitle}{Advances in neural information processing
  systems}, pp. \bibinfo{pages}{1097--1105}.
\bibitem[{Simonyan and Zisserman(2014)}]{simonyan2014very}
\bibinfo{author}{K.~Simonyan}, \bibinfo{author}{A.~Zisserman},
\newblock \bibinfo{title}{Very deep convolutional networks for large-scale
  image recognition},
\newblock \bibinfo{journal}{arXiv preprint arXiv:1409.1556}
  (\bibinfo{year}{2014}).
\bibitem[{Ronneberger et~al.(2015)Ronneberger, Fischer, and
  Brox}]{ronneberger2015u}
\bibinfo{author}{O.~Ronneberger}, \bibinfo{author}{P.~Fischer},
  \bibinfo{author}{T.~Brox},
\newblock \bibinfo{title}{U-net: Convolutional networks for biomedical image
  segmentation},
\newblock in: \bibinfo{booktitle}{International Conference on Medical image
  computing and computer-assisted intervention},
  \bibinfo{organization}{Springer}, pp. \bibinfo{pages}{234--241}.
\bibitem[{Raghu et~al.(2019)Raghu, Zhang, Kleinberg, and
  Bengio}]{raghu2019transfusion}
\bibinfo{author}{M.~Raghu}, \bibinfo{author}{C.~Zhang},
  \bibinfo{author}{J.~Kleinberg}, \bibinfo{author}{S.~Bengio},
\newblock \bibinfo{title}{Transfusion: Understanding transfer learning for
  medical imaging},
\newblock in: \bibinfo{booktitle}{Advances in Neural Information Processing
  Systems}, pp. \bibinfo{pages}{3342--3352}.
\bibitem[{Huynh et~al.(2016)Huynh, Li, and Giger}]{huynh2016digital}
\bibinfo{author}{B.~Q. Huynh}, \bibinfo{author}{H.~Li}, \bibinfo{author}{M.~L.
  Giger},
\newblock \bibinfo{title}{Digital mammographic tumor classification using
  transfer learning from deep convolutional neural networks},
\newblock \bibinfo{journal}{Journal of Medical Imaging} \bibinfo{volume}{3}
  (\bibinfo{year}{2016}) \bibinfo{pages}{034501}.
\bibitem[{N{\"a}ppi et~al.(2016)N{\"a}ppi, Hironaka, Regge, and
  Yoshida}]{nappi2016deep}
\bibinfo{author}{J.~J. N{\"a}ppi}, \bibinfo{author}{T.~Hironaka},
  \bibinfo{author}{D.~Regge}, \bibinfo{author}{H.~Yoshida},
\newblock \bibinfo{title}{Deep transfer learning of virtual endoluminal views
  for the detection of polyps in ct colonography},
\newblock in: \bibinfo{booktitle}{Medical Imaging 2016: Computer-Aided
  Diagnosis}, volume \bibinfo{volume}{9785},
  \bibinfo{organization}{International Society for Optics and Photonics}, p.
  \bibinfo{pages}{97852B}.
\bibitem[{Ghorbani et~al.(2020)Ghorbani, Natarajan, Coz, and
  Liu}]{ghorbani2020dermgan}
\bibinfo{author}{A.~Ghorbani}, \bibinfo{author}{V.~Natarajan},
  \bibinfo{author}{D.~Coz}, \bibinfo{author}{Y.~Liu},
\newblock \bibinfo{title}{Dermgan: Synthetic generation of clinical skin images
  with pathology},
\newblock in: \bibinfo{booktitle}{Machine Learning for Health Workshop}, pp.
  \bibinfo{pages}{155--170}.
\bibitem[{Finlayson et~al.(2019)Finlayson, Bowers, Ito, Zittrain, Beam, and
  Kohane}]{finlayson2019adversarial}
\bibinfo{author}{S.~G. Finlayson}, \bibinfo{author}{J.~D. Bowers},
  \bibinfo{author}{J.~Ito}, \bibinfo{author}{J.~L. Zittrain},
  \bibinfo{author}{A.~L. Beam}, \bibinfo{author}{I.~S. Kohane},
\newblock \bibinfo{title}{Adversarial attacks on medical machine learning},
\newblock \bibinfo{journal}{Science} \bibinfo{volume}{363}
  (\bibinfo{year}{2019}) \bibinfo{pages}{1287--1289}.
\bibitem[{Paschali et~al.(2018)Paschali, Conjeti, Navarro, and
  Navab}]{paschali2018generalizability}
\bibinfo{author}{M.~Paschali}, \bibinfo{author}{S.~Conjeti},
  \bibinfo{author}{F.~Navarro}, \bibinfo{author}{N.~Navab},
\newblock \bibinfo{title}{Generalizability vs. robustness: investigating
  medical imaging networks using adversarial examples},
\newblock in: \bibinfo{booktitle}{International Conference on Medical Image
  Computing and Computer-Assisted Intervention},
  \bibinfo{organization}{Springer}, pp. \bibinfo{pages}{493--501}.
\bibitem[{Chen et~al.(2019)Chen, Liu, Kingsbury, Sohn, Storlie, Habermann,
  Naessens, Larson, and Liu}]{chen2019deep}
\bibinfo{author}{D.~Chen}, \bibinfo{author}{S.~Liu},
  \bibinfo{author}{P.~Kingsbury}, \bibinfo{author}{S.~Sohn},
  \bibinfo{author}{C.~B. Storlie}, \bibinfo{author}{E.~B. Habermann},
  \bibinfo{author}{J.~M. Naessens}, \bibinfo{author}{D.~W. Larson},
  \bibinfo{author}{H.~Liu},
\newblock \bibinfo{title}{Deep learning and alternative learning strategies for
  retrospective real-world clinical data},
\newblock \bibinfo{journal}{NPJ digital medicine} \bibinfo{volume}{2}
  (\bibinfo{year}{2019}) \bibinfo{pages}{1--5}.
\bibitem[{Kelly et~al.(2019)Kelly, Karthikesalingam, Suleyman, Corrado, and
  King}]{kelly2019key}
\bibinfo{author}{C.~J. Kelly}, \bibinfo{author}{A.~Karthikesalingam},
  \bibinfo{author}{M.~Suleyman}, \bibinfo{author}{G.~Corrado},
  \bibinfo{author}{D.~King},
\newblock \bibinfo{title}{Key challenges for delivering clinical impact with
  artificial intelligence},
\newblock \bibinfo{journal}{BMC medicine} \bibinfo{volume}{17}
  (\bibinfo{year}{2019}) \bibinfo{pages}{195}.
\bibitem[{Ghosal et~al.(2018)Ghosal, Blystone, Singh, Ganapathysubramanian,
  Singh, and Sarkar}]{ghosal2018explainable}
\bibinfo{author}{S.~Ghosal}, \bibinfo{author}{D.~Blystone},
  \bibinfo{author}{A.~K. Singh}, \bibinfo{author}{B.~Ganapathysubramanian},
  \bibinfo{author}{A.~Singh}, \bibinfo{author}{S.~Sarkar},
\newblock \bibinfo{title}{An explainable deep machine vision framework for
  plant stress phenotyping},
\newblock \bibinfo{journal}{Proceedings of the National Academy of Sciences}
  \bibinfo{volume}{115} (\bibinfo{year}{2018}) \bibinfo{pages}{4613--4618}.
\bibitem[{Simonyan et~al.(2013)Simonyan, Vedaldi, and
  Zisserman}]{simonyan2013deep}
\bibinfo{author}{K.~Simonyan}, \bibinfo{author}{A.~Vedaldi},
  \bibinfo{author}{A.~Zisserman},
\newblock \bibinfo{title}{Deep inside convolutional networks: Visualising image
  classification models and saliency maps},
\newblock \bibinfo{journal}{arXiv preprint arXiv:1312.6034}
  (\bibinfo{year}{2013}).
\bibitem[{Ghosal et~al.(2019)Ghosal, Zheng, Chapman, Potgieter, Jordan, Wang,
  Singh, Singh, Hirafuji, Ninomiya et~al.}]{ghosal2019weakly}
\bibinfo{author}{S.~Ghosal}, \bibinfo{author}{B.~Zheng}, \bibinfo{author}{S.~C.
  Chapman}, \bibinfo{author}{A.~B. Potgieter}, \bibinfo{author}{D.~R. Jordan},
  \bibinfo{author}{X.~Wang}, \bibinfo{author}{A.~K. Singh},
  \bibinfo{author}{A.~Singh}, \bibinfo{author}{M.~Hirafuji},
  \bibinfo{author}{S.~Ninomiya}, et~al.,
\newblock \bibinfo{title}{A weakly supervised deep learning framework for
  sorghum head detection and counting},
\newblock \bibinfo{journal}{Plant Phenomics} \bibinfo{volume}{2019}
  (\bibinfo{year}{2019}) \bibinfo{pages}{1525874}.
\bibitem[{Pokuri et~al.(2019)Pokuri, Ghosal, Kokate, Sarkar, and
  Ganapathysubramanian}]{pokuri2019interpretable}
\bibinfo{author}{B.~S.~S. Pokuri}, \bibinfo{author}{S.~Ghosal},
  \bibinfo{author}{A.~Kokate}, \bibinfo{author}{S.~Sarkar},
  \bibinfo{author}{B.~Ganapathysubramanian},
\newblock \bibinfo{title}{Interpretable deep learning for guided
  microstructure-property explorations in photovoltaics},
\newblock \bibinfo{journal}{npj Computational Materials} \bibinfo{volume}{5}
  (\bibinfo{year}{2019}) \bibinfo{pages}{1--11}.
\bibitem[{Selvaraju et~al.(2017)Selvaraju, Cogswell, Das, Vedantam, Parikh, and
  Batra}]{selvaraju2017grad}
\bibinfo{author}{R.~R. Selvaraju}, \bibinfo{author}{M.~Cogswell},
  \bibinfo{author}{A.~Das}, \bibinfo{author}{R.~Vedantam},
  \bibinfo{author}{D.~Parikh}, \bibinfo{author}{D.~Batra},
\newblock \bibinfo{title}{Grad-cam: Visual explanations from deep networks via
  gradient-based localization},
\newblock in: \bibinfo{booktitle}{Proceedings of the IEEE international
  conference on computer vision}, pp. \bibinfo{pages}{618--626}.
\bibitem[{McKinney et~al.(2020)McKinney, Sieniek, Godbole, Godwin, Antropova,
  Ashrafian, Back, Chesus, Corrado, Darzi et~al.}]{mckinney2020international}
\bibinfo{author}{S.~M. McKinney}, \bibinfo{author}{M.~Sieniek},
  \bibinfo{author}{V.~Godbole}, \bibinfo{author}{J.~Godwin},
  \bibinfo{author}{N.~Antropova}, \bibinfo{author}{H.~Ashrafian},
  \bibinfo{author}{T.~Back}, \bibinfo{author}{M.~Chesus},
  \bibinfo{author}{G.~C. Corrado}, \bibinfo{author}{A.~Darzi}, et~al.,
\newblock \bibinfo{title}{International evaluation of an ai system for breast
  cancer screening},
\newblock \bibinfo{journal}{Nature} \bibinfo{volume}{577}
  (\bibinfo{year}{2020}) \bibinfo{pages}{89--94}.
\bibitem[{Ardila et~al.(2019)Ardila, Kiraly, Bharadwaj, Choi, Reicher, Peng,
  Tse, Etemadi, Ye, Corrado et~al.}]{ardila2019end}
\bibinfo{author}{D.~Ardila}, \bibinfo{author}{A.~P. Kiraly},
  \bibinfo{author}{S.~Bharadwaj}, \bibinfo{author}{B.~Choi},
  \bibinfo{author}{J.~J. Reicher}, \bibinfo{author}{L.~Peng},
  \bibinfo{author}{D.~Tse}, \bibinfo{author}{M.~Etemadi},
  \bibinfo{author}{W.~Ye}, \bibinfo{author}{G.~Corrado}, et~al.,
\newblock \bibinfo{title}{End-to-end lung cancer screening with
  three-dimensional deep learning on low-dose chest computed tomography},
\newblock \bibinfo{journal}{Nature medicine} \bibinfo{volume}{25}
  (\bibinfo{year}{2019}) \bibinfo{pages}{954--961}.
\bibitem[{Shie et~al.(2015)Shie, Chuang, Chou, Wu, and
  Chang}]{shie2015transfer}
\bibinfo{author}{C.-K. Shie}, \bibinfo{author}{C.-H. Chuang},
  \bibinfo{author}{C.-N. Chou}, \bibinfo{author}{M.-H. Wu},
  \bibinfo{author}{E.~Y. Chang},
\newblock \bibinfo{title}{Transfer representation learning for medical image
  analysis},
\newblock in: \bibinfo{booktitle}{2015 37th annual international conference of
  the IEEE Engineering in Medicine and Biology Society (EMBC)},
  \bibinfo{organization}{IEEE}, pp. \bibinfo{pages}{711--714}.
\bibitem[{Zech et~al.(2018)Zech, Badgeley, Liu, Costa, Titano, and
  Oermann}]{zech2018variable}
\bibinfo{author}{J.~R. Zech}, \bibinfo{author}{M.~A. Badgeley},
  \bibinfo{author}{M.~Liu}, \bibinfo{author}{A.~B. Costa},
  \bibinfo{author}{J.~J. Titano}, \bibinfo{author}{E.~K. Oermann},
\newblock \bibinfo{title}{Variable generalization performance of a deep
  learning model to detect pneumonia in chest radiographs: a cross-sectional
  study},
\newblock \bibinfo{journal}{PLoS medicine} \bibinfo{volume}{15}
  (\bibinfo{year}{2018}) \bibinfo{pages}{e1002683}.
\bibitem[{Collins and Moons(2019)}]{collins2019reporting}
\bibinfo{author}{G.~S. Collins}, \bibinfo{author}{K.~G. Moons},
\newblock \bibinfo{title}{Reporting of artificial intelligence prediction
  models},
\newblock \bibinfo{journal}{The Lancet} \bibinfo{volume}{393}
  (\bibinfo{year}{2019}) \bibinfo{pages}{1577--1579}.
\bibitem[{Russakovsky et~al.(2015)Russakovsky, Deng, Su, Krause, Satheesh, Ma,
  Huang, Karpathy, Khosla, Bernstein, Berg, and Fei-Fei}]{ILSVRC15}
\bibinfo{author}{O.~Russakovsky}, \bibinfo{author}{J.~Deng},
  \bibinfo{author}{H.~Su}, \bibinfo{author}{J.~Krause},
  \bibinfo{author}{S.~Satheesh}, \bibinfo{author}{S.~Ma},
  \bibinfo{author}{Z.~Huang}, \bibinfo{author}{A.~Karpathy},
  \bibinfo{author}{A.~Khosla}, \bibinfo{author}{M.~Bernstein},
  \bibinfo{author}{A.~C. Berg}, \bibinfo{author}{L.~Fei-Fei},
\newblock \bibinfo{title}{{ImageNet Large Scale Visual Recognition Challenge}},
\newblock \bibinfo{journal}{International Journal of Computer Vision (IJCV)}
  \bibinfo{volume}{115} (\bibinfo{year}{2015}) \bibinfo{pages}{211--252}.
\bibitem[{Shah et~al.(2019)Shah, Kendall, Khozin, Goosen, Hu, Laramie, Ringel,
  and Schork}]{shah2019artificial}
\bibinfo{author}{P.~Shah}, \bibinfo{author}{F.~Kendall},
  \bibinfo{author}{S.~Khozin}, \bibinfo{author}{R.~Goosen},
  \bibinfo{author}{J.~Hu}, \bibinfo{author}{J.~Laramie},
  \bibinfo{author}{M.~Ringel}, \bibinfo{author}{N.~Schork},
\newblock \bibinfo{title}{Artificial intelligence and machine learning in
  clinical development: a translational perspective},
\newblock \bibinfo{journal}{NPJ digital medicine} \bibinfo{volume}{2}
  (\bibinfo{year}{2019}) \bibinfo{pages}{1--5}.
\bibitem[{Ghafoorian et~al.(2017)Ghafoorian, Mehrtash, Kapur, Karssemeijer,
  Marchiori, Pesteie, Guttmann, de~Leeuw, Tempany, Van~Ginneken
  et~al.}]{ghafoorian2017transfer}
\bibinfo{author}{M.~Ghafoorian}, \bibinfo{author}{A.~Mehrtash},
  \bibinfo{author}{T.~Kapur}, \bibinfo{author}{N.~Karssemeijer},
  \bibinfo{author}{E.~Marchiori}, \bibinfo{author}{M.~Pesteie},
  \bibinfo{author}{C.~R. Guttmann}, \bibinfo{author}{F.-E. de~Leeuw},
  \bibinfo{author}{C.~M. Tempany}, \bibinfo{author}{B.~Van~Ginneken}, et~al.,
\newblock \bibinfo{title}{Transfer learning for domain adaptation in mri:
  Application in brain lesion segmentation},
\newblock in: \bibinfo{booktitle}{International conference on medical image
  computing and computer-assisted intervention},
  \bibinfo{organization}{Springer}, pp. \bibinfo{pages}{516--524}.
\bibitem[{Van~Opbroek et~al.(2014)Van~Opbroek, Ikram, Vernooij, and
  De~Bruijne}]{van2014transfer}
\bibinfo{author}{A.~Van~Opbroek}, \bibinfo{author}{M.~A. Ikram},
  \bibinfo{author}{M.~W. Vernooij}, \bibinfo{author}{M.~De~Bruijne},
\newblock \bibinfo{title}{Transfer learning improves supervised image
  segmentation across imaging protocols},
\newblock \bibinfo{journal}{IEEE transactions on medical imaging}
  \bibinfo{volume}{34} (\bibinfo{year}{2014}) \bibinfo{pages}{1018--1030}.
\bibitem[{Dou et~al.(2018)Dou, Ouyang, Chen, Chen, and
  Heng}]{dou2018unsupervised}
\bibinfo{author}{Q.~Dou}, \bibinfo{author}{C.~Ouyang},
  \bibinfo{author}{C.~Chen}, \bibinfo{author}{H.~Chen}, \bibinfo{author}{P.-A.
  Heng},
\newblock \bibinfo{title}{Unsupervised cross-modality domain adaptation of
  convnets for biomedical image segmentations with adversarial loss},
\newblock \bibinfo{journal}{arXiv preprint arXiv:1804.10916}
  (\bibinfo{year}{2018}).
\bibitem[{Van~Ginneken et~al.(2015)Van~Ginneken, Setio, Jacobs, and
  Ciompi}]{van2015off}
\bibinfo{author}{B.~Van~Ginneken}, \bibinfo{author}{A.~A. Setio},
  \bibinfo{author}{C.~Jacobs}, \bibinfo{author}{F.~Ciompi},
\newblock \bibinfo{title}{Off-the-shelf convolutional neural network features
  for pulmonary nodule detection in computed tomography scans},
\newblock in: \bibinfo{booktitle}{2015 IEEE 12th International symposium on
  biomedical imaging (ISBI)}, \bibinfo{organization}{IEEE}, pp.
  \bibinfo{pages}{286--289}.
\bibitem[{Bar et~al.(2015)Bar, Diamant, Wolf, and Greenspan}]{bar2015deep}
\bibinfo{author}{Y.~Bar}, \bibinfo{author}{I.~Diamant},
  \bibinfo{author}{L.~Wolf}, \bibinfo{author}{H.~Greenspan},
\newblock \bibinfo{title}{Deep learning with non-medical training used for
  chest pathology identification},
\newblock in: \bibinfo{booktitle}{Medical Imaging 2015: Computer-Aided
  Diagnosis}, volume \bibinfo{volume}{9414},
  \bibinfo{organization}{International Society for Optics and Photonics}, p.
  \bibinfo{pages}{94140V}.
\bibitem[{Ciompi et~al.(2015)Ciompi, de~Hoop, van Riel, Chung, Scholten,
  Oudkerk, de~Jong, Prokop, and van Ginneken}]{ciompi2015automatic}
\bibinfo{author}{F.~Ciompi}, \bibinfo{author}{B.~de~Hoop},
  \bibinfo{author}{S.~J. van Riel}, \bibinfo{author}{K.~Chung},
  \bibinfo{author}{E.~T. Scholten}, \bibinfo{author}{M.~Oudkerk},
  \bibinfo{author}{P.~A. de~Jong}, \bibinfo{author}{M.~Prokop},
  \bibinfo{author}{B.~van Ginneken},
\newblock \bibinfo{title}{Automatic classification of pulmonary peri-fissural
  nodules in computed tomography using an ensemble of 2d views and a
  convolutional neural network out-of-the-box},
\newblock \bibinfo{journal}{Medical image analysis} \bibinfo{volume}{26}
  (\bibinfo{year}{2015}) \bibinfo{pages}{195--202}.
\bibitem[{Banerjee et~al.(2018)Banerjee, Crawley, Bhethanabotla, Daldrup-Link,
  and Rubin}]{banerjee2018transfer}
\bibinfo{author}{I.~Banerjee}, \bibinfo{author}{A.~Crawley},
  \bibinfo{author}{M.~Bhethanabotla}, \bibinfo{author}{H.~E. Daldrup-Link},
  \bibinfo{author}{D.~L. Rubin},
\newblock \bibinfo{title}{Transfer learning on fused multiparametric mr images
  for classifying histopathological subtypes of rhabdomyosarcoma},
\newblock \bibinfo{journal}{Computerized Medical Imaging and Graphics}
  \bibinfo{volume}{65} (\bibinfo{year}{2018}) \bibinfo{pages}{167--175}.
\bibitem[{Giorgiani~do Nascimento and Viana(2020)}]{giorgiani2020satellite}
\bibinfo{author}{R.~Giorgiani~do Nascimento}, \bibinfo{author}{F.~Viana},
\newblock \bibinfo{title}{Satellite image classification and segmentation with
  transfer learning},
\newblock in: \bibinfo{booktitle}{AIAA Scitech 2020 Forum}, p.
  \bibinfo{pages}{1864}.
\bibitem[{He et~al.(2019)He, Girshick, and Doll{\'a}r}]{he2019rethinking}
\bibinfo{author}{K.~He}, \bibinfo{author}{R.~Girshick},
  \bibinfo{author}{P.~Doll{\'a}r},
\newblock \bibinfo{title}{Rethinking imagenet pre-training},
\newblock in: \bibinfo{booktitle}{Proceedings of the IEEE international
  conference on computer vision}, pp. \bibinfo{pages}{4918--4927}.
\bibitem[{Kornblith et~al.(2019)Kornblith, Shlens, and
  Le}]{kornblith2019better}
\bibinfo{author}{S.~Kornblith}, \bibinfo{author}{J.~Shlens},
  \bibinfo{author}{Q.~V. Le},
\newblock \bibinfo{title}{Do better imagenet models transfer better?},
\newblock in: \bibinfo{booktitle}{Proceedings of the IEEE conference on
  computer vision and pattern recognition}, pp. \bibinfo{pages}{2661--2671}.
\bibitem[{ref(2019{\natexlab{a}})}]{ref_url1}
\bibinfo{title}{{ISIC Archive} homepage}, \bibinfo{year}{2019}{\natexlab{a}}.
  \bibinfo{note}{\url{https://www.isic-archive.com}}.
\bibitem[{ref(2019{\natexlab{b}})}]{ref_url2}
\bibinfo{title}{Gleason 2019 homepage}, \bibinfo{year}{2019}{\natexlab{b}}.
  \bibinfo{note}{\url{https://gleason2019.grand-challenge.org/}}.
\bibitem[{Heller et~al.(2019)Heller, Sathianathen, Kalapara, Walczak, Moore,
  Kaluzniak, Rosenberg, Blake, Rengel, Oestreich et~al.}]{ref_data1}
\bibinfo{author}{N.~Heller}, \bibinfo{author}{N.~Sathianathen},
  \bibinfo{author}{A.~Kalapara}, \bibinfo{author}{E.~Walczak},
  \bibinfo{author}{K.~Moore}, \bibinfo{author}{H.~Kaluzniak},
  \bibinfo{author}{J.~Rosenberg}, \bibinfo{author}{P.~Blake},
  \bibinfo{author}{Z.~Rengel}, \bibinfo{author}{M.~Oestreich}, et~al.,
\newblock \bibinfo{title}{The kits19 challenge data: 300 kidney tumor cases
  with clinical context, ct semantic segmentations, and surgical outcomes},
\newblock \bibinfo{journal}{arXiv preprint arXiv:1904.00445}
  (\bibinfo{year}{2019}).
\bibitem[{Athanasiou et~al.(2017)Athanasiou, Fotiadis, and
  Michalis}]{athanasiou2017atherosclerotic}
\bibinfo{author}{L.~S. Athanasiou}, \bibinfo{author}{D.~I. Fotiadis},
  \bibinfo{author}{L.~K. Michalis}, \bibinfo{title}{Atherosclerotic Plaque
  Characterization Methods Based on Coronary Imaging},
  \bibinfo{publisher}{Academic Press}, \bibinfo{year}{2017}.
\bibitem[{Iglovikov and Shvets(2018)}]{iglovikov2018ternausnet}
\bibinfo{author}{V.~Iglovikov}, \bibinfo{author}{A.~Shvets},
\newblock \bibinfo{title}{Ternausnet: U-net with vgg11 encoder pre-trained on
  imagenet for image segmentation},
\newblock \bibinfo{journal}{arXiv preprint arXiv:1801.05746}
  (\bibinfo{year}{2018}).
\bibitem[{ref(2012)}]{ref_url3}
\bibinfo{title}{{ImageNet} archive}, \bibinfo{year}{2012}.
  \bibinfo{note}{\url{http://www.image-net.org/}}.
\bibitem[{Zeiler(2012)}]{zeiler2012adadelta}
\bibinfo{author}{M.~D. Zeiler},
\newblock \bibinfo{title}{Adadelta: an adaptive learning rate method},
\newblock \bibinfo{journal}{arXiv preprint arXiv:1212.5701}
  (\bibinfo{year}{2012}).
\bibitem[{Taha and Hanbury(2015)}]{taha2015metrics}
\bibinfo{author}{A.~A. Taha}, \bibinfo{author}{A.~Hanbury},
\newblock \bibinfo{title}{Metrics for evaluating 3d medical image segmentation:
  analysis, selection, and tool},
\newblock \bibinfo{journal}{BMC medical imaging} \bibinfo{volume}{15}
  (\bibinfo{year}{2015}) \bibinfo{pages}{29}.
\bibitem[{Razali et~al.(2011)Razali, Wah et~al.}]{razali2011power}
\bibinfo{author}{N.~M. Razali}, \bibinfo{author}{Y.~B. Wah}, et~al.,
\newblock \bibinfo{title}{Power comparisons of shapiro-wilk,
  kolmogorov-smirnov, lilliefors and anderson-darling tests},
\newblock \bibinfo{journal}{Journal of statistical modeling and analytics}
  \bibinfo{volume}{2} (\bibinfo{year}{2011}) \bibinfo{pages}{21--33}.
\bibitem[{Weisberg(2001)}]{weisberg2001yeo}
\bibinfo{author}{S.~Weisberg},
\newblock \bibinfo{title}{Yeo-johnson power transformations},
\newblock \bibinfo{journal}{Department of Applied Statistics, University of
  Minnesota. Retrieved June} \bibinfo{volume}{1} (\bibinfo{year}{2001})
  \bibinfo{pages}{2003}.
\bibitem[{Mood et~al.(1963)Mood, Graybill, and Boes}]{mood1963introduction}
\bibinfo{author}{A.~Mood}, \bibinfo{author}{F.~Graybill},
  \bibinfo{author}{D.~Boes},
\newblock \bibinfo{title}{Introduction to the theory of statistics. mc-graw
  hill book company},
\newblock \bibinfo{journal}{Inc., New York}  (\bibinfo{year}{1963}).
\bibitem[{Zar(2013)}]{zar2013biostatistical}
\bibinfo{author}{J.~H. Zar}, \bibinfo{title}{Biostatistical analysis: Pearson
  new international edition}, \bibinfo{publisher}{Pearson Higher Ed},
  \bibinfo{year}{2013}.
\bibitem[{Dice(1945)}]{dice1945measures}
\bibinfo{author}{L.~R. Dice},
\newblock \bibinfo{title}{Measures of the amount of ecologic association
  between species},
\newblock \bibinfo{journal}{Ecology} \bibinfo{volume}{26}
  (\bibinfo{year}{1945}) \bibinfo{pages}{297--302}.
\bibitem[{Zhou et~al.(2016)Zhou, Khosla, Lapedriza, Oliva, and
  Torralba}]{zhou2016learning}
\bibinfo{author}{B.~Zhou}, \bibinfo{author}{A.~Khosla},
  \bibinfo{author}{A.~Lapedriza}, \bibinfo{author}{A.~Oliva},
  \bibinfo{author}{A.~Torralba},
\newblock \bibinfo{title}{Learning deep features for discriminative
  localization},
\newblock in: \bibinfo{booktitle}{Proceedings of the IEEE conference on
  computer vision and pattern recognition}, pp. \bibinfo{pages}{2921--2929}.
\bibitem[{Lin et~al.(2013)Lin, Chen, and Yan}]{lin2013network}
\bibinfo{author}{M.~Lin}, \bibinfo{author}{Q.~Chen}, \bibinfo{author}{S.~Yan},
\newblock \bibinfo{title}{Network in network},
\newblock \bibinfo{journal}{arXiv preprint arXiv:1312.4400}
  (\bibinfo{year}{2013}).

\end{thebibliography}

\clearpage



\clearpage
\newpage

\section*{SUPPLEMENTARY INFORMATION}
\newpage

\renewcommand\thefigure{S\arabic{figure}}
\setcounter{figure}{0}

\renewcommand\thesection{S\arabic{section}}
\setcounter{section}{0}

\renewcommand\thetable{S\arabic{table}}
\setcounter{table}{0}

\begin{table*}
\centering
\caption{Median, mean and standard deviations (sd.) of area under the receiver operating curve (AUROC), Dice score, sensitivity, and specificity median values of binary segmentation of RGB skin images. Randomization of the skin image data set into five different 80\% (training images): 20\% (test images) sets and bootstrapping (five replicates) of transfer learning ($T_{II}$) and learning from medical images ($L_{MI}$) deep learning models are reported. Differences greater than 5\% are indicated by ``$\textbf{\textdagger}$" for the better-performing model. A value of 1 indicates a perfect score.}\label{stab1}
\ra{1.3}
\resizebox{\textwidth}{!}{
\begin{tabular}{@{}rccrccrccrccrcc@{}}
\toprule
& \multicolumn{2}{c}{Set 1} & \phantom{abc} & \multicolumn{2}{c}{Set 2} & \phantom{abc} & \multicolumn{2}{c}{Set 3} & \phantom{abc} & \multicolumn{2}{c}{Set 4} & \phantom{abc} & \multicolumn{2}{c}{Set 5}\\
& $L_{MI}$ & $T_{II}$ && $L_{MI}$ & $T_{II}$ && $L_{MI}$ & $T_{II}$ && $L_{MI}$ & $T_{II}$ && $L_{MI}$ & $T_{II}$ \\
\toprule
\textbf{AUROC}\\
median & 0.8544 & 0.9282\textsuperscript{\textdagger} && 0.8332 & 0.9182\textsuperscript{\textdagger} && 0.8633 & 0.9247\textsuperscript{\textdagger} && 0.8695 & 0.9162 && 0.8825 & 0.9377\textsuperscript{\textdagger}\\
mean & 0.8120 & 0.8826 && 0.7992 & 0.8696 && 0.8237 & 0.8805 && 0.8260 & 0.8704 && 0.8323 & 0.8858\\
sd. & 0.1514 & 0.1277 && 0.1510 & 0.1382 && 0.1446 & 0.1252 && 0.1491 & 0.1340 && 0.1534 & 0.1336\\
\textbf{Dice Score}\\
median & 0.8273 & 0.8857\textsuperscript{\textdagger} && 0.7974 & 0.8774\textsuperscript{\textdagger} && 0.8370 & 0.8806 && 0.8470 & 0.8815 && 0.8620 & 0.8880\\
mean & 0.7483 & 0.8250 && 0.7217 & 0.8135 && 0.7599 & 0.8221 && 0.7690 & 0.8132 && 0.7773 & 0.8264\\
sd. & 0.2169 & 0.1786 && 0.2324 & 0.1857 && 0.2167 & 0.1761 && 0.2128 & 0.1882 && 0.2180 & 0.1803\\
\textbf{Sensitivity}\\
median & 0.7156 & 0.8891\textsuperscript{\textdagger} && 0.6711 & 0.8572\textsuperscript{\textdagger} && 0.7329 & 0.8834\textsuperscript{\textdagger} && 0.7439 & 0.8704\textsuperscript{\textdagger} && 0.7775 & 0.8976\textsuperscript{\textdagger}\\
mean & 0.6331 & 0.7944 && 0.6047 & 0.7577 && 0.6568 & 0.7922 && 0.6579 & 0.7704 && 0.6759 & 0.7901\\
sd. & 0.3080 & 0.2541 && 0.3052 & 0.2794 && 0.2935 & 0.2497 && 0.3007 & 0.2710 && 0.3109 & 0.2702\\
\textbf{Specificity}\\
median & 0.9999 & 0.9986 && 1 & 0.9989 && 0.9999 & 0.9986 && 0.9999 & 0.9987 && 0.9998 & 0.9981\\
mean & 0.9922 & 0.9722 && 0.9949 & 0.9846 && 0.9912 & 0.9725 && 0.9953 & 0.9745 && 0.9896 & 0.9829\\
sd. & 0.0341 & 0.0945 && 0.0277 & 0.0544 && 0.0433 & 0.0931 && 0.0216 & 0.0830 && 0.0495 & 0.0543\\
\\
\bottomrule
\end{tabular}
}
\end{table*}

\clearpage
\newpage
\begin{table*}
\centering
\caption{Distribution of clinical label class for each of the five 80:20 sets of RGB images with skin cancer and microscopic Hematoxylin $\&$ Eosin (H $\&$ E) stained prostate core biopsy and computed tomography (CT) of kidneys. Randomization of data sets and clinical label classes into five different 80\% (training images): 20\% (test images) sets were used for training transfer learning ($T_{II}$) and learning from medical images ($L_{MI}$ ) deep learning models. Percentages correspond to the ratio of the number of images belonging to a particular clinical label class to the total number of images in the training or testing set.}\label{stab2}
\ra{1.5}
\resizebox{\textwidth}{!}{
\begin{tabular}{@{}rccrccrccrccrcc@{}}
\toprule
& \multicolumn{2}{c}{Set 1} & \phantom{abc} & \multicolumn{2}{c}{Set 2} & \phantom{abc} & \multicolumn{2}{c}{Set 3} & \phantom{abc} & \multicolumn{2}{c}{Set 4} & \phantom{abc} & \multicolumn{2}{c}{Set 5}\\
Input image & Train (\%) & Test (\%) && Train (\%) & Test (\%) && Train (\%) & Test (\%) && Train (\%) & Test (\%) && Train (\%) & Test (\%)\\
\toprule
\textbf{Skin}\\
Benign & 10132 (92) & 2536 (92) && 10126 (92) & 2542 (92) && 10123 (92) & 2545 (92) && 10147 (92) & 2521 (91) && 10122 (92) & 2546 (92)\\
Malignant & 896 (8) & 222 (8) && 902 (8) & 216 (8) && 905 (8) & 213 (8) && 881 (8)) & 237 (9) && 906 (8) & 212 (8)\\
\textbf{Prostate core biopsy}\\
Tumor & 177 (91) & 47 (96) && 178 (91) & 46 (94) && 179 (92) & 45 (92) && 178 (91) & 46 (94) && 179 (92) & 45 (92)\\
No Tumor or Background & 18 (9) & 2 (4) && 17 (9) & 3 (6) && 16 (8) & 4 (8) && 17 (9) & 3 (6) && 16 (8) & 4 (8)\\
\textbf{Kidney CT}\\
Kidney tissue present & 13091 (36) & 3245 (36) && 13055 (36) & 3281 (36) && 13105 (36) & 3231 (36) && 13056 (36) & 3280 (36) && 13065 (36) & 3271 (36)\\
Kidney tissue absent & 23248 (64) & 5840 (64) && 23284 (64) & 5804 (64) && 23234 (64) & 5854 (64) && 23283 (64) & 5805 (64) && 23274 (64) & 5814 (64)\\
\\
\bottomrule
\end{tabular}
}
\end{table*}

\clearpage
\newpage
\begin{table*}
\centering
\caption{Median, mean, and standard deviations (sd.) of Dice scores from five replicates of binary segmentation of RGB images of skin with benign or malignant tumors by transfer learning ($T_{II}$) and learning from medical images ($L_{MI}$) deep learning models. Differences greater than 5\% are indicated by ``$\textbf{\textdagger}$" for the better-performing model. A value of 1 indicates a perfect score.}\label{stab3}
\ra{1}
\resizebox{0.7\textwidth}{!}{
\begin{tabular}{@{}rccrcc@{}}\toprule
& \multicolumn{2}{c}{Benign}  & \phantom{abc} & \multicolumn{2}{c}{Malignant} \\
& $L_{MI}$ & $T_{II}$ && $L_{MI}$ & $T_{II}$ \\
\midrule
median & 0.8260 & 0.8860\textsuperscript{\textdagger} && 0.8402 & 0.8835\\
mean & 0.7467 & 0.8231\textsuperscript{\textdagger} && 0.7650 & 0.8456\textsuperscript{\textdagger}\\
sd. & 0.2184 & 0.2018 && 0.1821 & 0.1335\\
\bottomrule
\end{tabular}
}
\end{table*}
\begin{table*}
\centering
\caption{Median, mean, and standard deviations (sd.) of the median distributions of Dice scores for computed tomography (CT) images with kidneys. Fivefold replicates and binary segmentation by transfer learning ($T_{II}$) and learning using medical images ($L_{MI}$) deep learning models. A value of 1 indicates a perfect score.}\label{stab4}
\ra{1.3}
\resizebox{0.6\textwidth}{!}{
\begin{tabular}{@{}rcc@{}}\toprule
& \multicolumn{2}{c}{Dice score for images with kidneys}\\
\midrule
& $L_{MI}$ & $T_{II}$\\
median & 0.9597 & 0.9598\\
mean & 0.9509 & 0.9502\\
sd. & 0.0657 & 0.0622\\
\bottomrule
\end{tabular}
}
\end{table*}


\begin{figure*}[ht!]
\begin{subfigure}{0.24\textwidth}
\centering\includegraphics[width=0.9\textwidth]{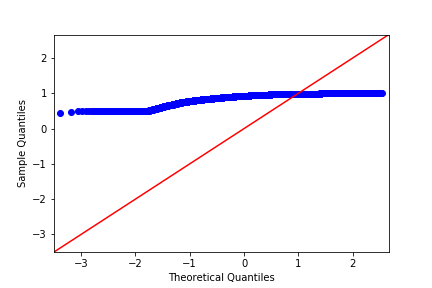}
\caption{}
\end{subfigure}
\begin{subfigure}{0.24\textwidth}
\centering\includegraphics[width=0.9\textwidth]{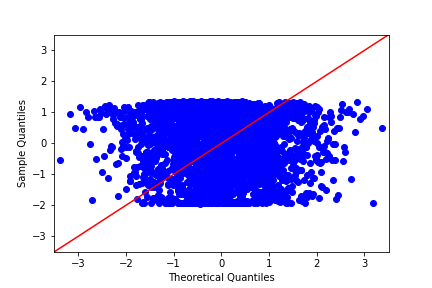}
\caption{}
\end{subfigure}
\begin{subfigure}{0.24\textwidth}
\centering\includegraphics[width=0.9\textwidth]{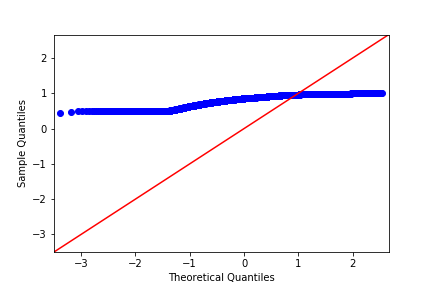}
\caption{}
\end{subfigure}
\begin{subfigure}{0.24\textwidth}
\centering\includegraphics[width=0.9\textwidth]{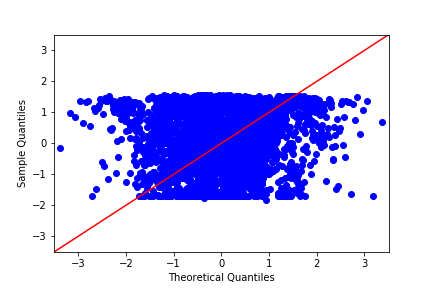}
\caption{}
\end{subfigure}
\\
\newline
\begin{subfigure}{0.24\textwidth}
\centering\includegraphics[width=0.9\textwidth]{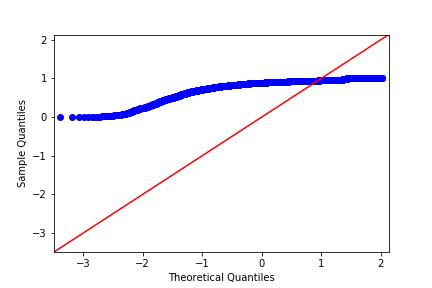}
\caption{}
\end{subfigure}
\begin{subfigure}{0.24\textwidth}
\centering\includegraphics[width=0.9\textwidth]{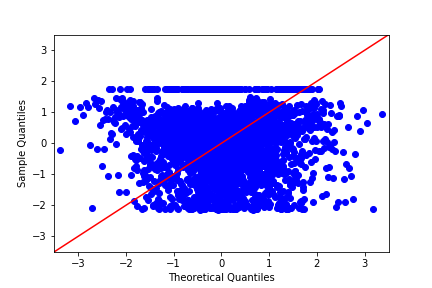}
\caption{}
\end{subfigure}
\begin{subfigure}{0.24\textwidth}
\centering\includegraphics[width=0.9\textwidth]{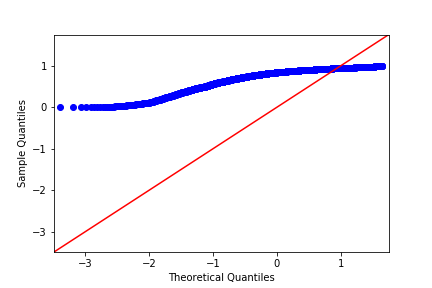}
\caption{}
\end{subfigure}
\begin{subfigure}{0.24\textwidth}
\centering\includegraphics[width=0.9\textwidth]{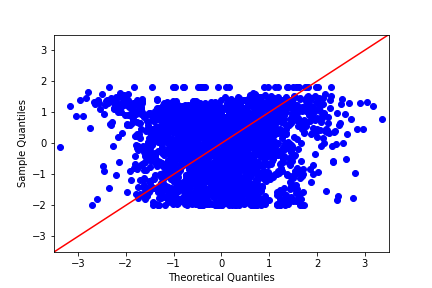}
\caption{}
\end{subfigure}
\\
\begin{subfigure}{0.24\textwidth}
\centering\includegraphics[width=0.9\textwidth]{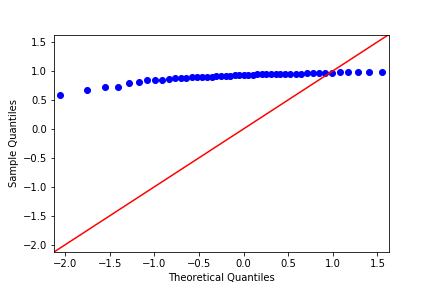}
\caption{}
\end{subfigure}
\begin{subfigure}{0.24\textwidth}
\centering\includegraphics[width=0.9\textwidth]{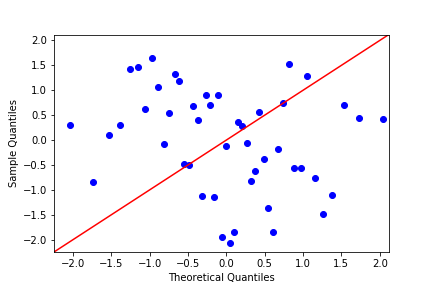}
\caption{}
\end{subfigure}
\begin{subfigure}{0.24\textwidth}
\centering\includegraphics[width=0.9\textwidth]{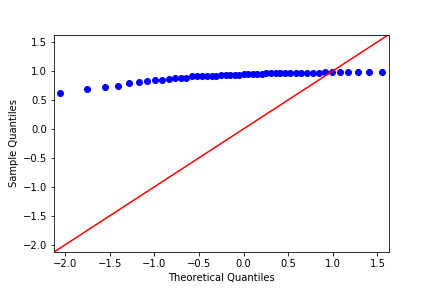}
\caption{}
\end{subfigure}
\begin{subfigure}{0.24\textwidth}
\centering\includegraphics[width=0.9\textwidth]{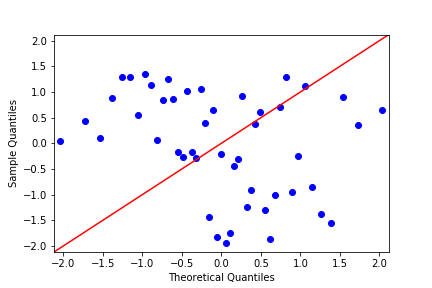}
\caption{}
\end{subfigure}
\\
\newline
\begin{subfigure}{0.24\textwidth}
\centering\includegraphics[width=0.9\textwidth]{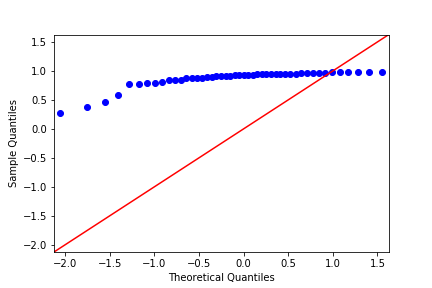}
\caption{}
\end{subfigure}
\begin{subfigure}{0.24\textwidth}
\centering\includegraphics[width=0.9\textwidth]{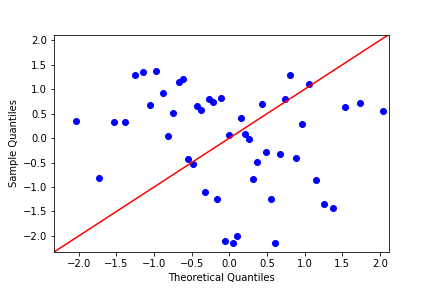}
\caption{}
\end{subfigure}
\begin{subfigure}{0.24\textwidth}
\centering\includegraphics[width=0.9\textwidth]{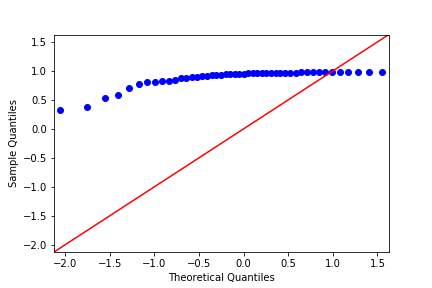}
\caption{}
\end{subfigure}
\begin{subfigure}{0.24\textwidth}
\centering\includegraphics[width=0.9\textwidth]{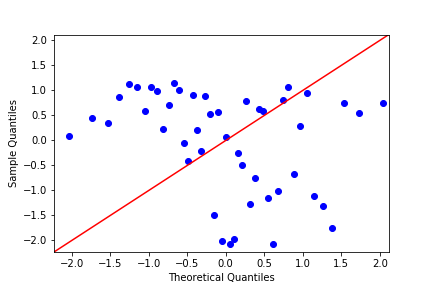}
\caption{}
\end{subfigure}
\\
\begin{subfigure}{0.24\textwidth}
\centering\includegraphics[width=0.9\textwidth]{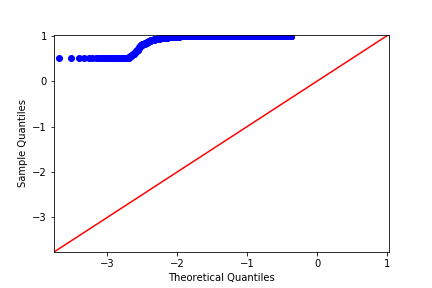}
\caption{}
\end{subfigure}
\begin{subfigure}{0.24\textwidth}
\centering\includegraphics[width=0.9\textwidth]{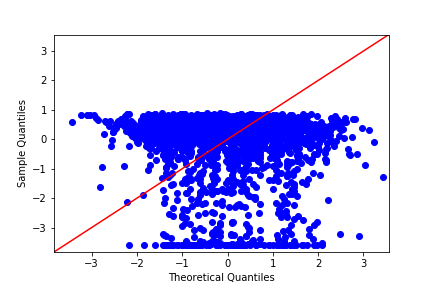}
\caption{}
\end{subfigure}
\begin{subfigure}{0.24\textwidth}
\centering\includegraphics[width=0.9\textwidth]{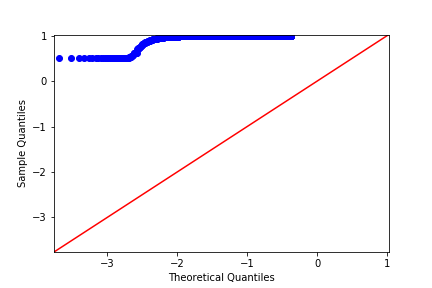}
\caption{}
\end{subfigure}
\begin{subfigure}{0.24\textwidth}
\centering\includegraphics[width=0.9\textwidth]{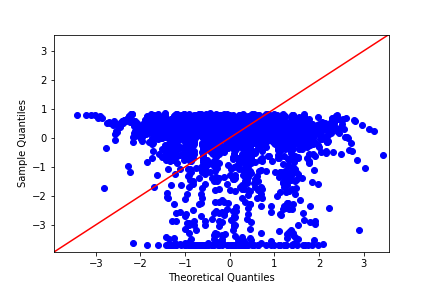}
\caption{}
\end{subfigure}
\\
\newline
\begin{subfigure}{0.24\textwidth}
\centering\includegraphics[width=0.9\textwidth]{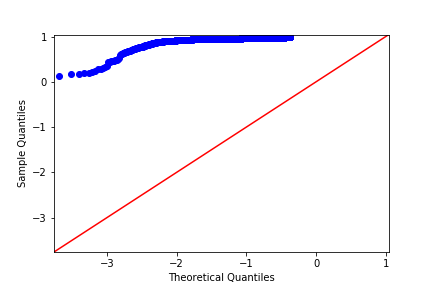}
\caption{}
\end{subfigure}
\begin{subfigure}{0.24\textwidth}
\centering\includegraphics[width=0.9\textwidth]{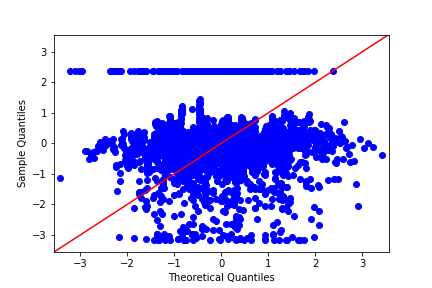}
\caption{}
\end{subfigure}
\begin{subfigure}{0.24\textwidth}
\centering\includegraphics[width=0.9\textwidth]{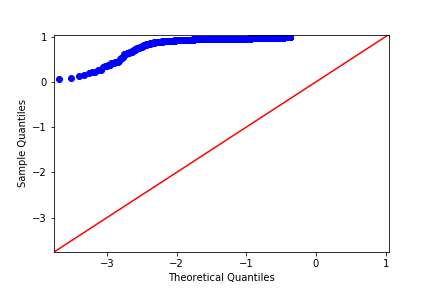}
\caption{}
\end{subfigure}
\begin{subfigure}{0.24\textwidth}
\centering\includegraphics[width=0.9\textwidth]{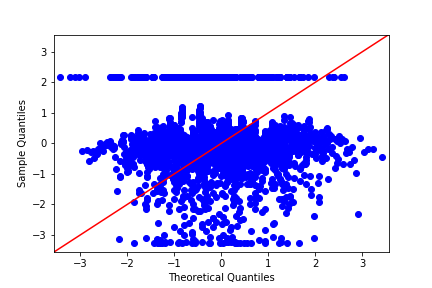}
\caption{}
\end{subfigure}
\caption{\scriptsize{Quantile-quantile plots of AUROC and Dice score distributions for skin ([i]–[viii]), prostate core biopsy ([ix]–[xvi]), and kidney CT ([xvii]–[xxiv]) images data displaying the deviation from normality assumptions. Quantile-quantile plots presented in even-numbered sub-images after Yeo-Johnson (YJ) transformation. AUROC distributions are in panels ([i]–[iv], [ix]–[xii] and [xvii]–[xx]). Dice score distributions are in panels ([v]–[viii], [xiii]–[xvi] and [xxi]–[xxiv]). First two columns display $T_{II}$ distributions, and last two columns display $L_{MI}$ distributions. Theoretical quantiles (y-axis on each subfigure) present the standard normal distribution. Sample quantiles (x-axis) are from the metric distributions of AUROC and Dice scores from this study. If a distribution follows normality assumptions, blue data points indicate a linear trend. The red diagonal line represents the standard normal distribution.}}\label{sfig1}
\end{figure*}

\begin{figure*}[ht!]
\begin{subfigure}{0.5\textwidth}
\centering\includegraphics[width=0.9\textwidth]{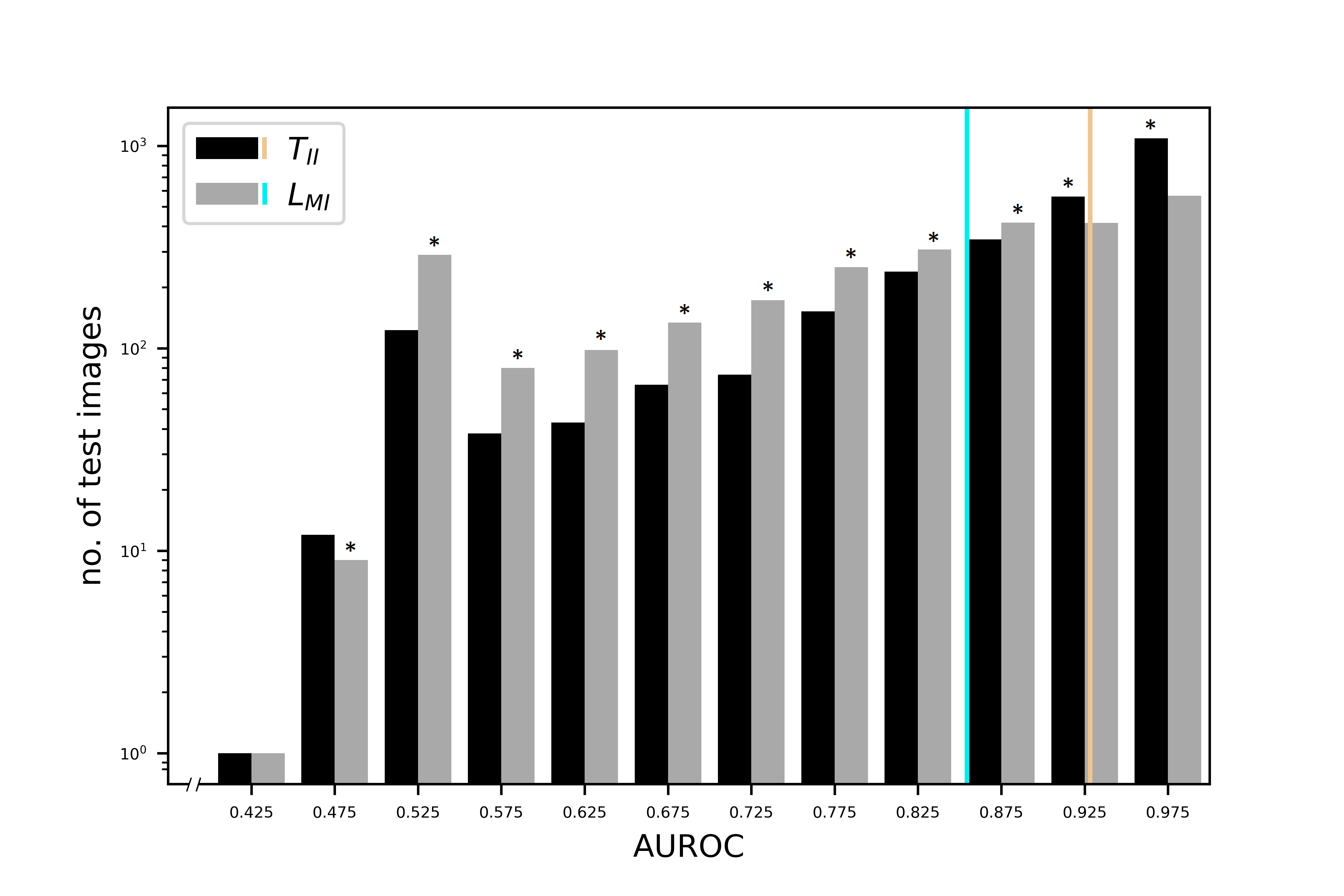}
\caption{}
\end{subfigure}
\begin{subfigure}{0.5\textwidth}
\centering\includegraphics[width=0.9\textwidth]{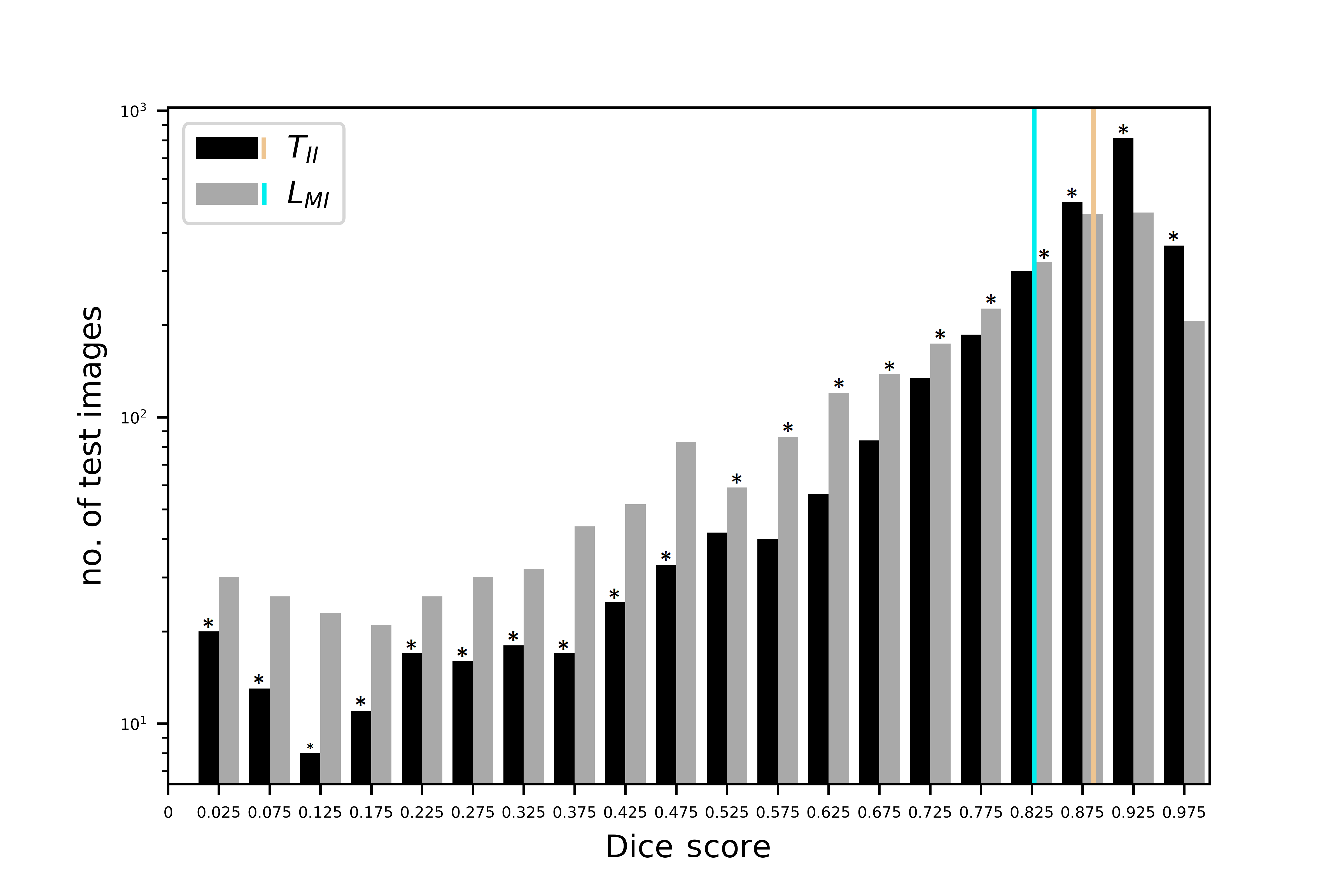}
\caption{}
\end{subfigure}\\
\newline
\begin{subfigure}{0.5\textwidth}
\centering\includegraphics[width=0.9\textwidth]{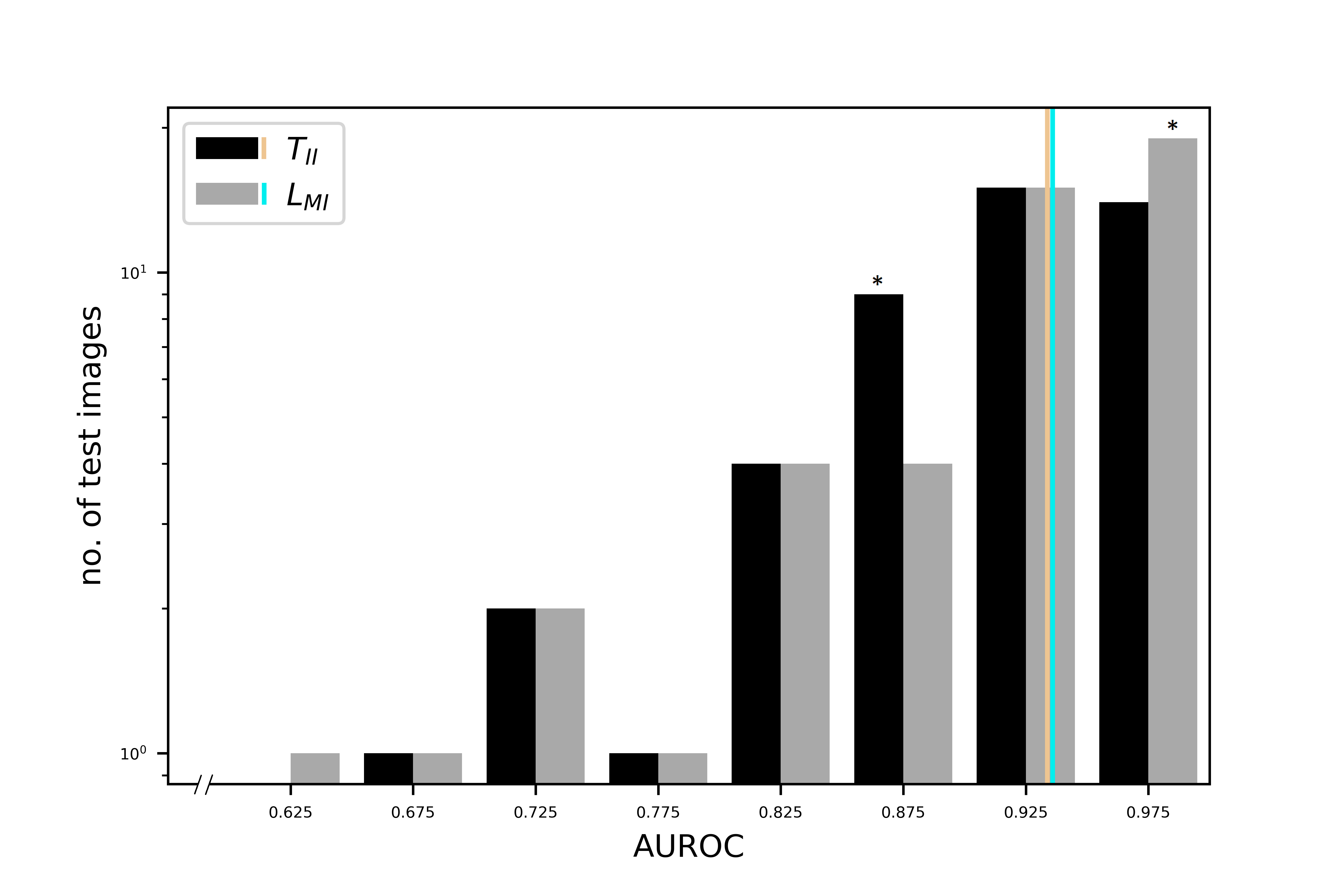}
\caption{}
\end{subfigure}
\begin{subfigure}{0.5\textwidth}
\centering\includegraphics[width=0.9\textwidth]{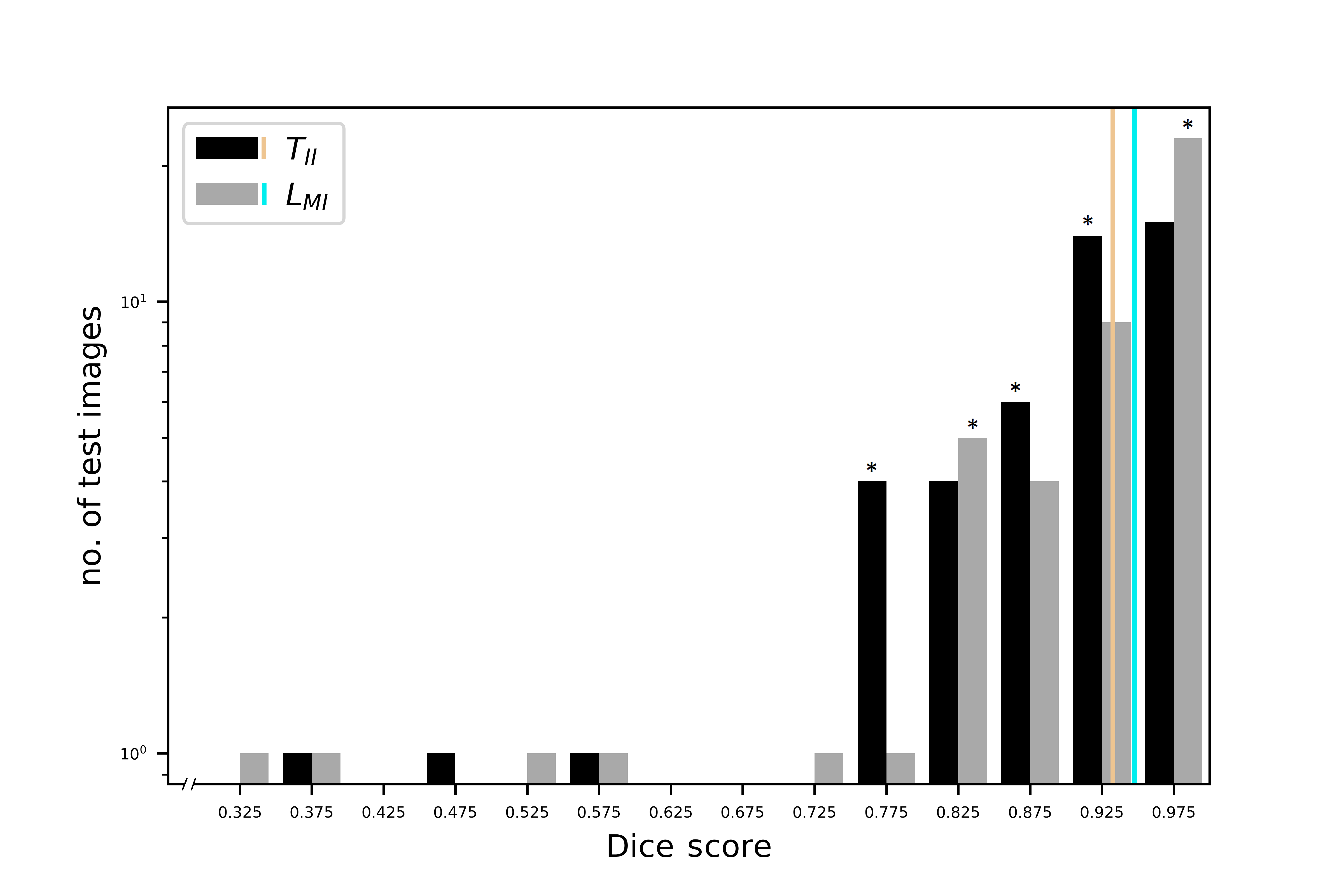}
\caption{}
\end{subfigure}\\
\begin{subfigure}{0.5\textwidth}
\centering\includegraphics[width=0.9\textwidth]{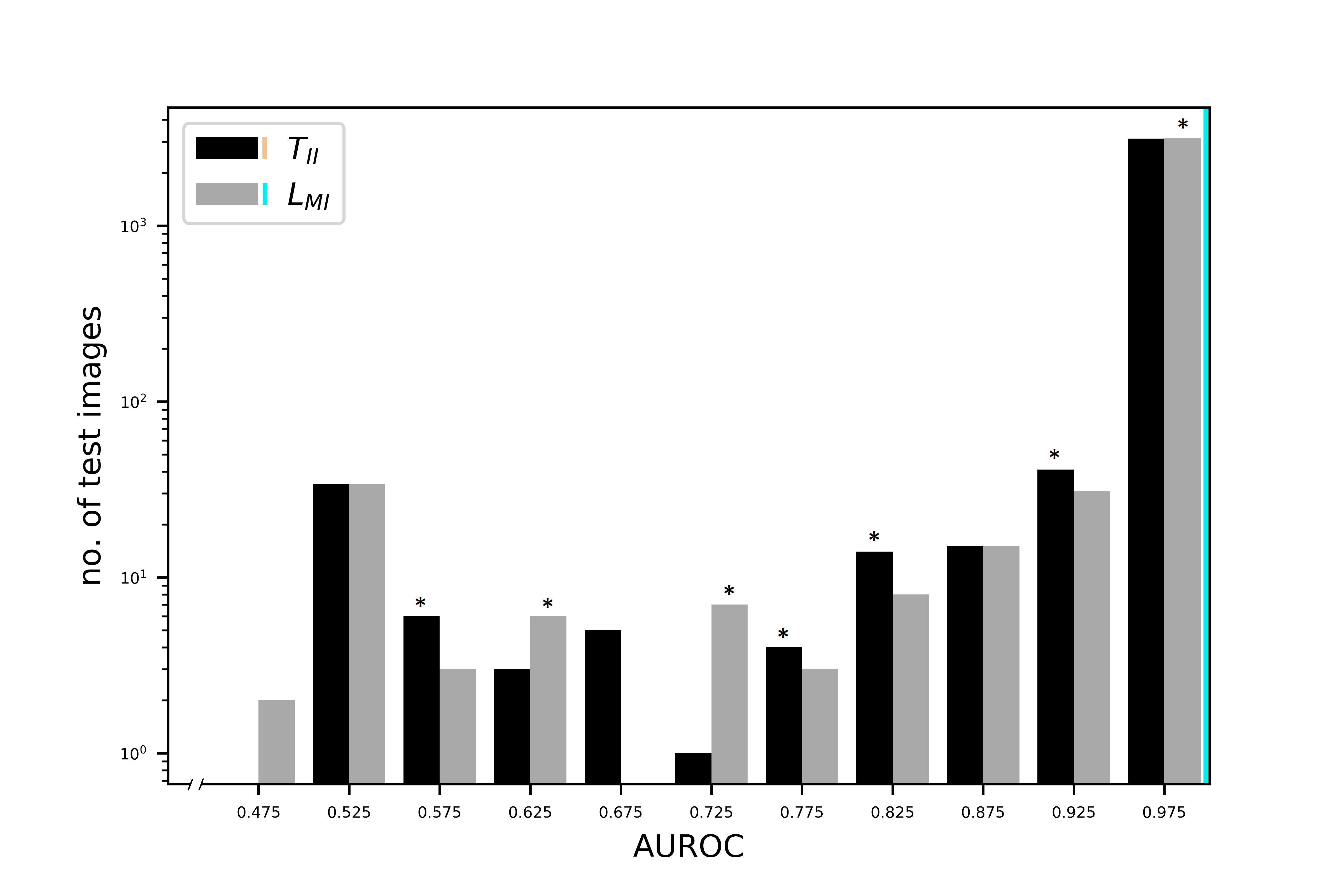}
\caption{}
\end{subfigure}
\begin{subfigure}{0.5\textwidth}
\centering\includegraphics[width=0.9\textwidth]{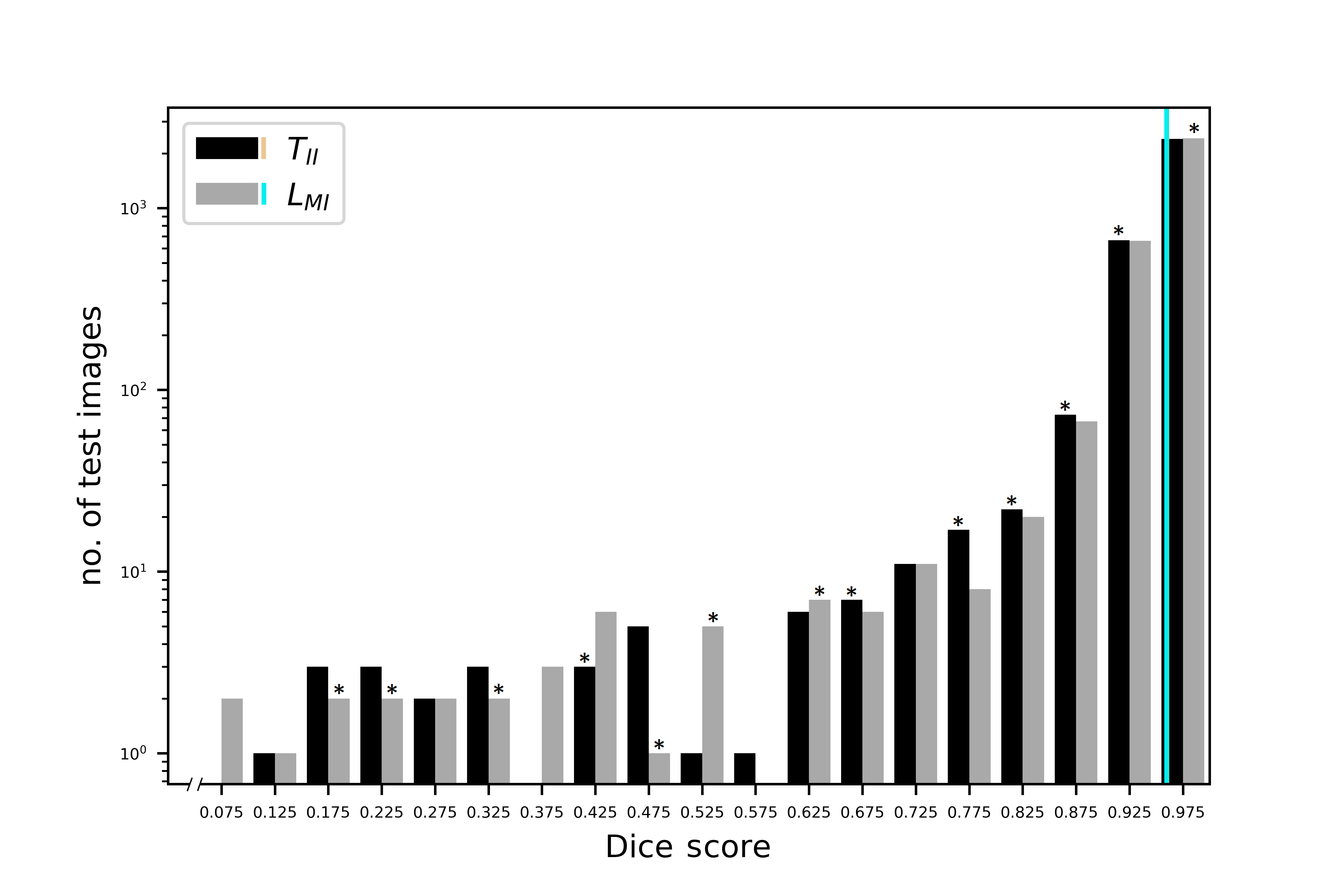}
\caption{}
\end{subfigure}
\caption{\scriptsize{Grouped bar plots presenting the distribution of AUROC and Dice scores achieved by transfer learning ($T_{II}$) and learning from medical images ($L_{MI}$) models for all test images in 80:20 data splits. AUROC and Dice scores for skin images are displayed in panels (i) and (ii); prostate core biopsy in (iii) and (iv); and kidney computed tomography images in (v) and (vi). Brown and blue vertical lines represent the median values of the distributions for $T_{II}$ and $L_{MI}$. Bars above 0.5 and below 0.5 indicate better model performance. Better-performing models at a particular AUROC or Dice score are indicated by \textbf{*}.}} \label{sfig2}
\end{figure*}

\begin{figure*}[h!]
\begin{subfigure}{0.5\textwidth}
\centering\includegraphics[width=0.9\textwidth]{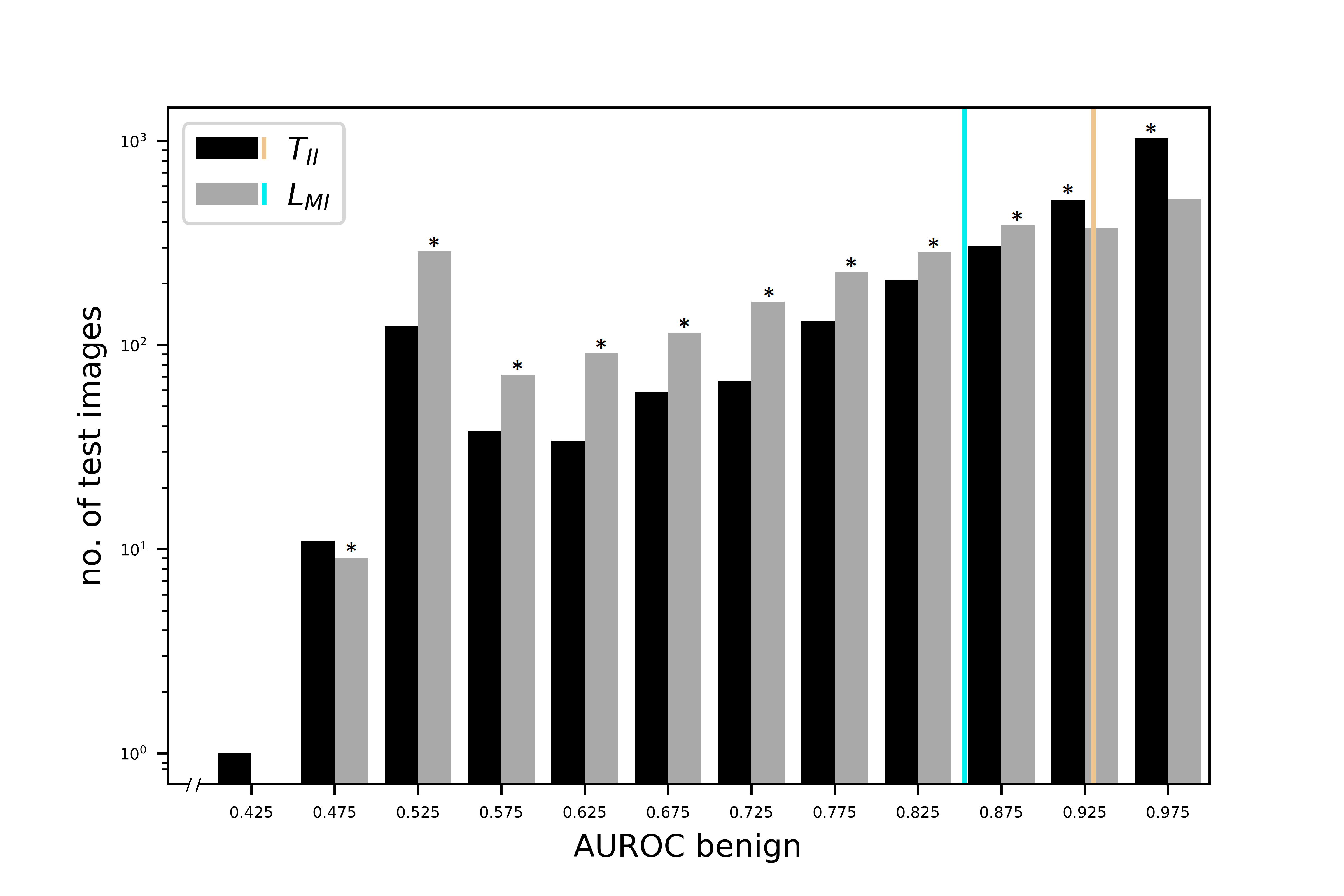}
\caption{}
\end{subfigure}
\begin{subfigure}{0.5\textwidth}
\centering\includegraphics[width=0.9\textwidth]{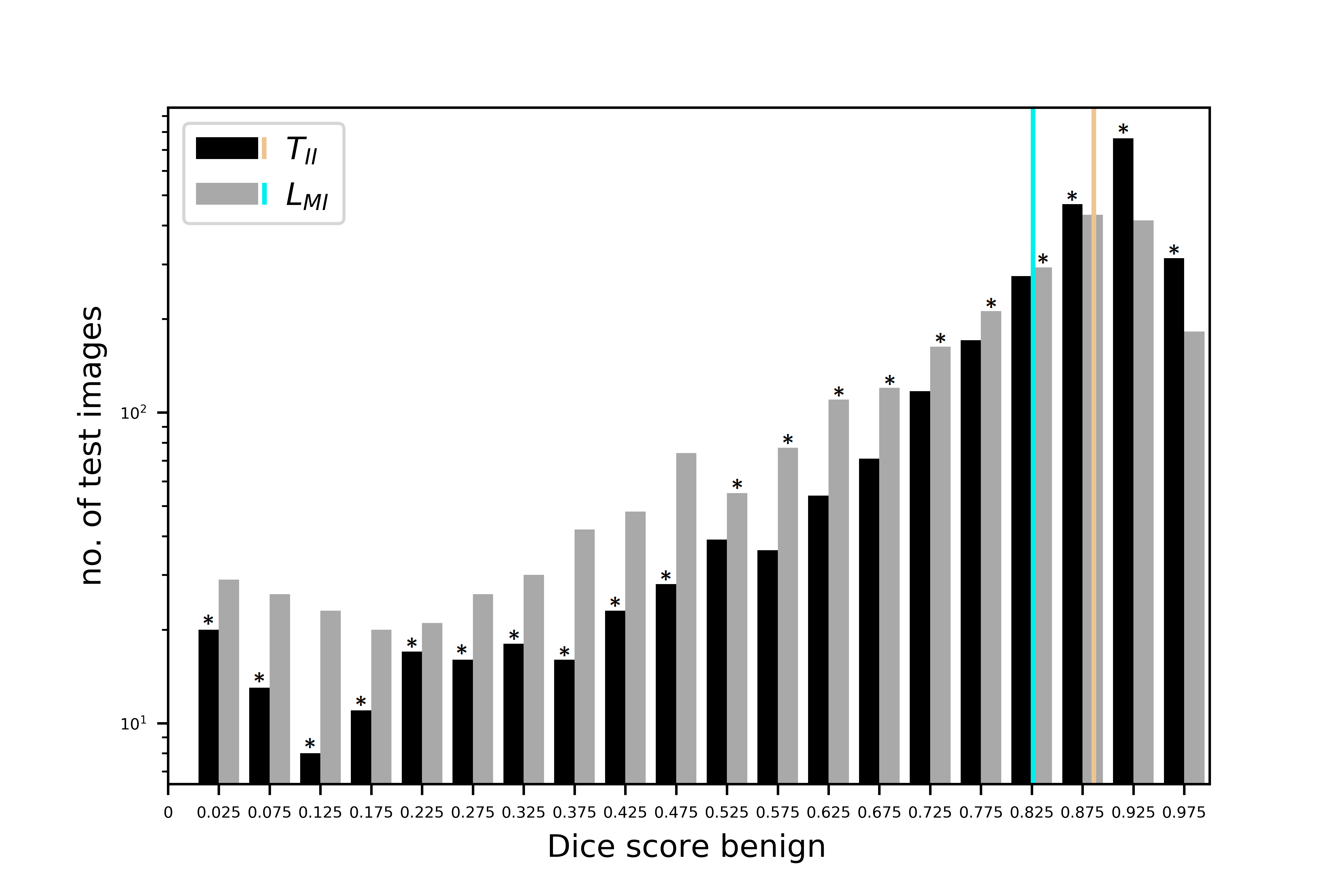}
\caption{}
\end{subfigure}\\
\newline
\begin{subfigure}{0.5\textwidth}
\centering\includegraphics[width=0.9\textwidth]{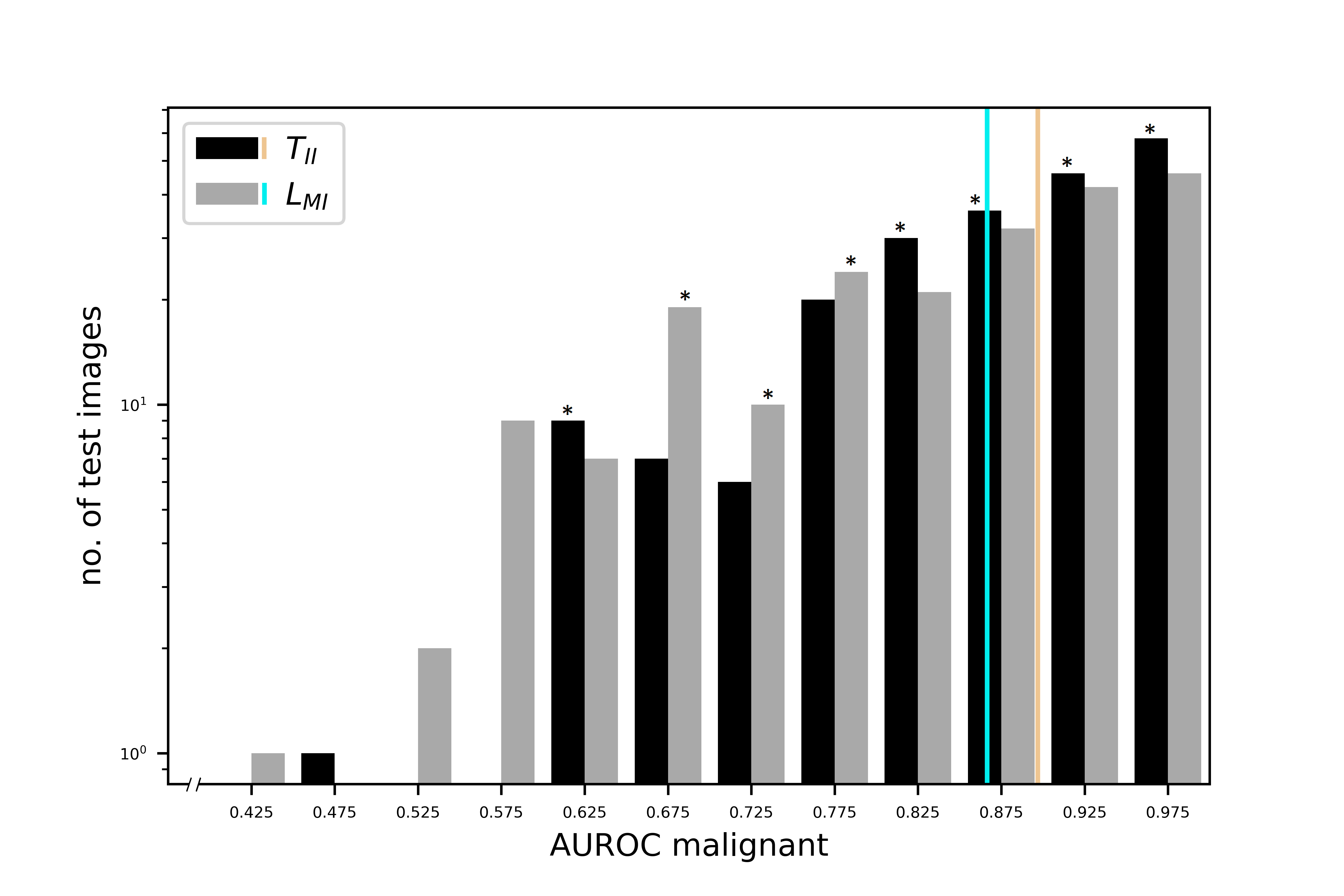}
\caption{}
\end{subfigure}
\begin{subfigure}{0.5\textwidth}
\centering\includegraphics[width=0.9\textwidth]{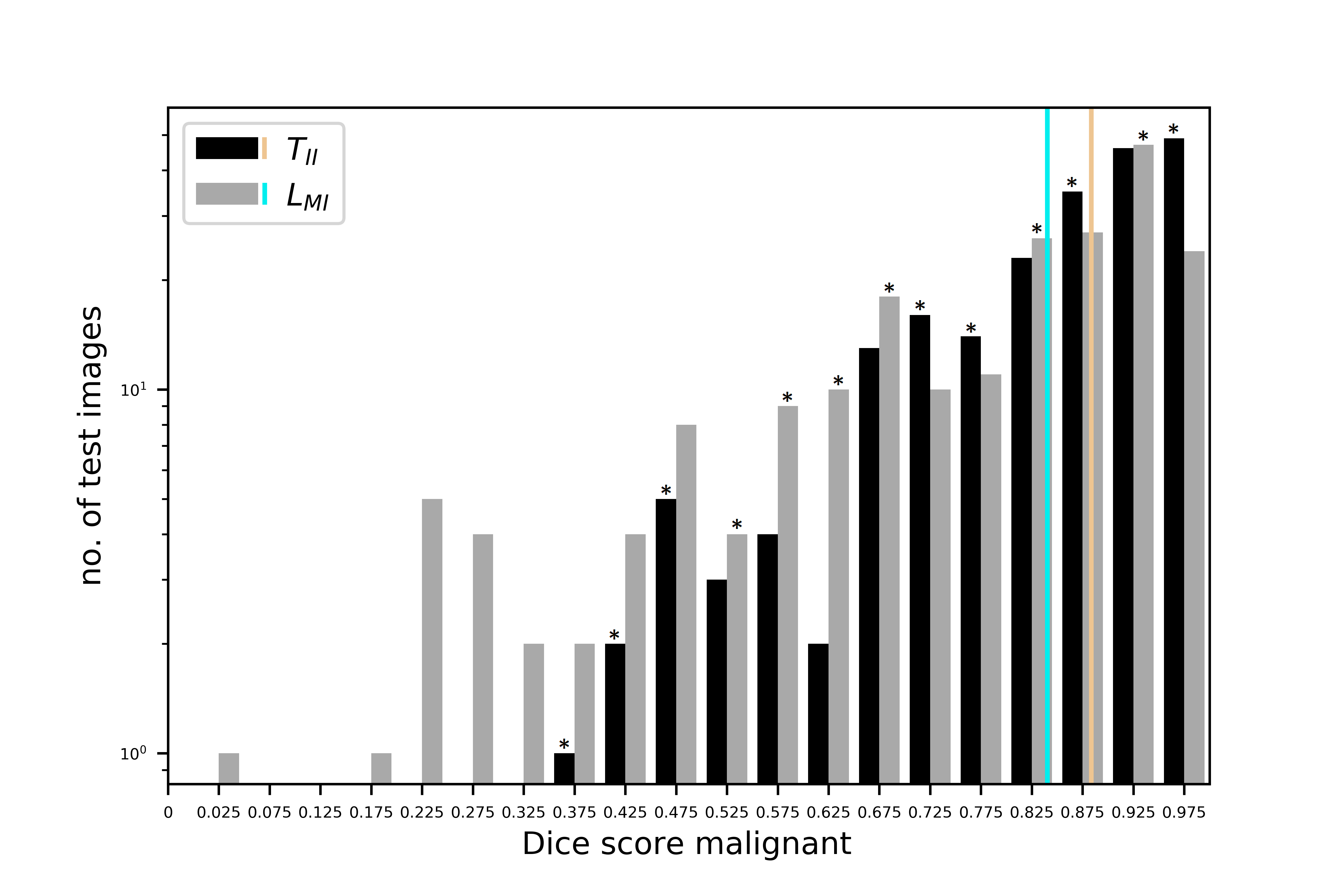}
\caption{}
\end{subfigure}
\caption{\scriptsize{Grouped bar plots presenting the distribution of area under the receiver operating characteristics curve (AUROC) and Dice scores achieved by transfer learning ($T_{II}$) and learning from medical images ($L_{MI}$) models for skin images. AUROC and Dice score distributions for benign skin images are displayed in panels (i) and (ii) and malignant skin images in panels (iii) and (iv). Brown and blue vertical lines represent the median values of the distributions for $T_{II}$ and $L_{MI}$. Bars above 0.5 and below 0.5 indicate better model performance. Better-performing models at a particular AUROC or Dice score are indicated by \textbf{*}.}}\label{sfig3}
\end{figure*}

\begin{figure*}[h!]
\begin{subfigure}{0.5\textwidth}
\centering\includegraphics[width=0.9\textwidth]{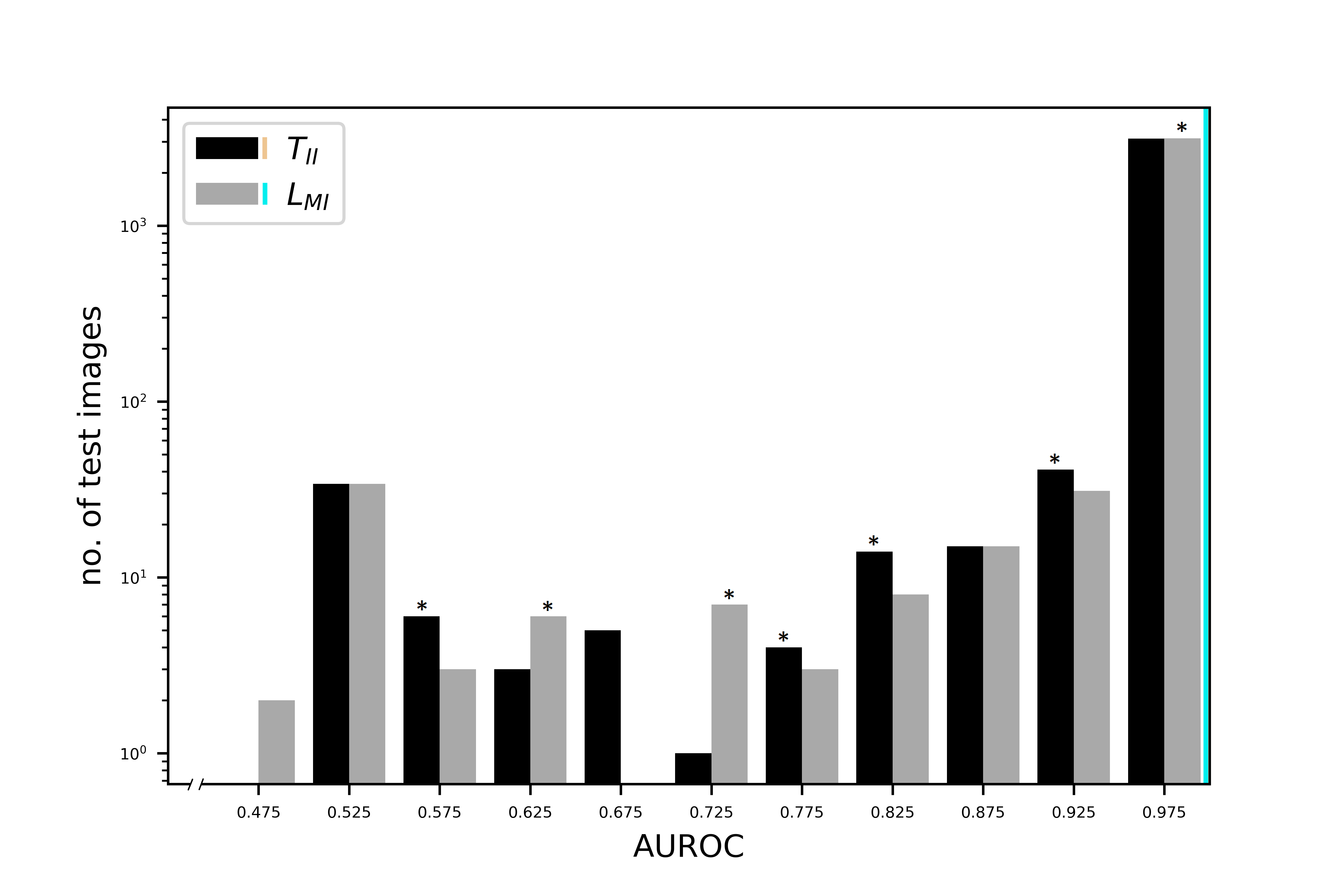}
\caption{}
\end{subfigure}
\begin{subfigure}{0.5\textwidth}
\centering\includegraphics[width=0.9\textwidth]{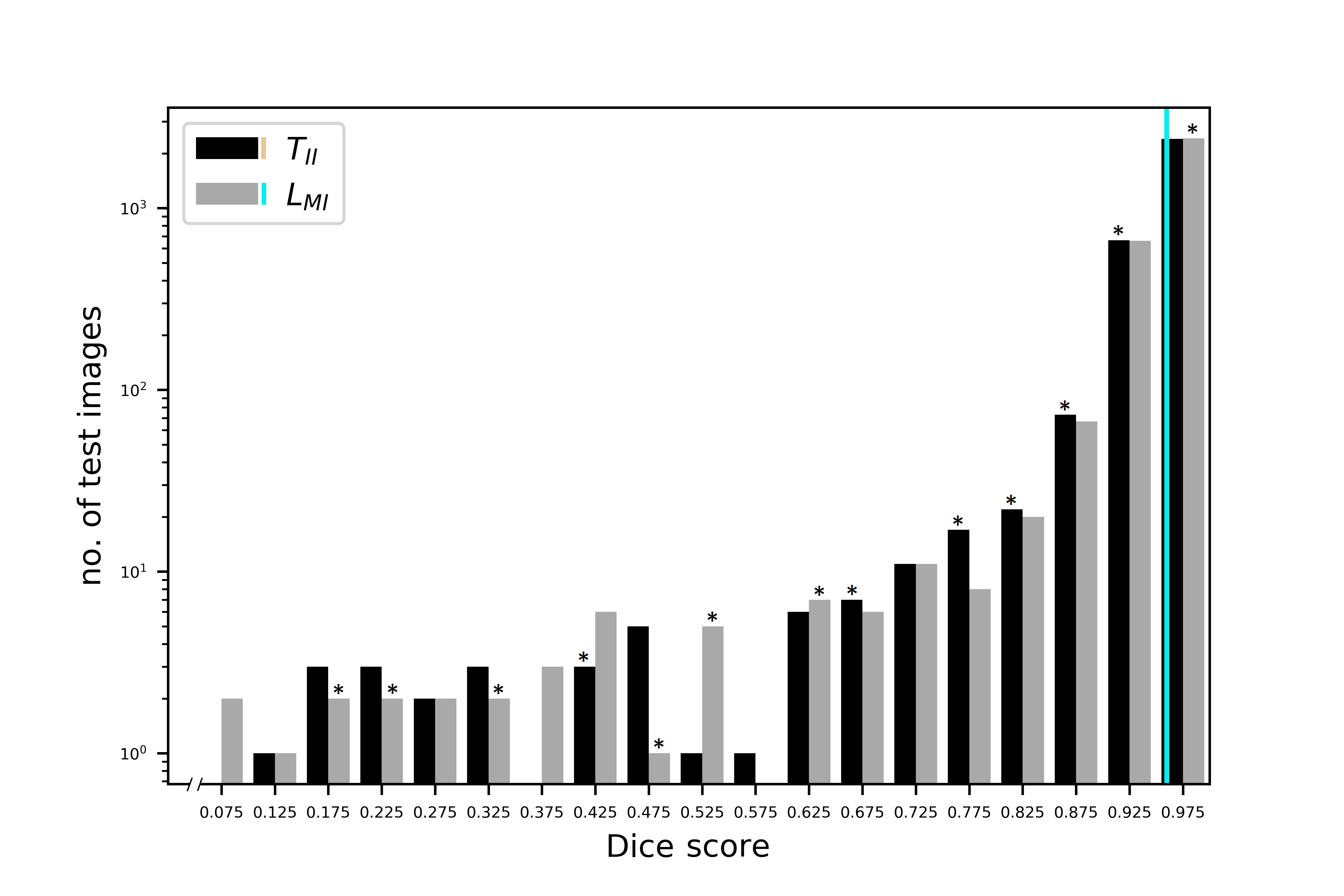}
\caption{}
\end{subfigure}\\
\caption{\scriptsize{Grouped bar plots presenting the distribution of area under the receiver operating characteristics curve (AUROC) and Dice scores achieved by transfer learning ($T_{II}$) and learning from medical images ($L_{MI}$) models for computed tomography test images with kidneys. Brown and blue vertical lines represent the median values of the distributions for $T_{II}$ and $L_{MI}$. Bars above 0.5 and below 0.5 indicate better model performance. Better-performing models at a particular AUROC or Dice score are indicated by \textbf{*}.}}\label{sfig4}
\end{figure*}


\begin{figure*}[ht!]
\begin{subfigure}{0.99\textwidth}
\centering\includegraphics[width=0.8\textwidth]{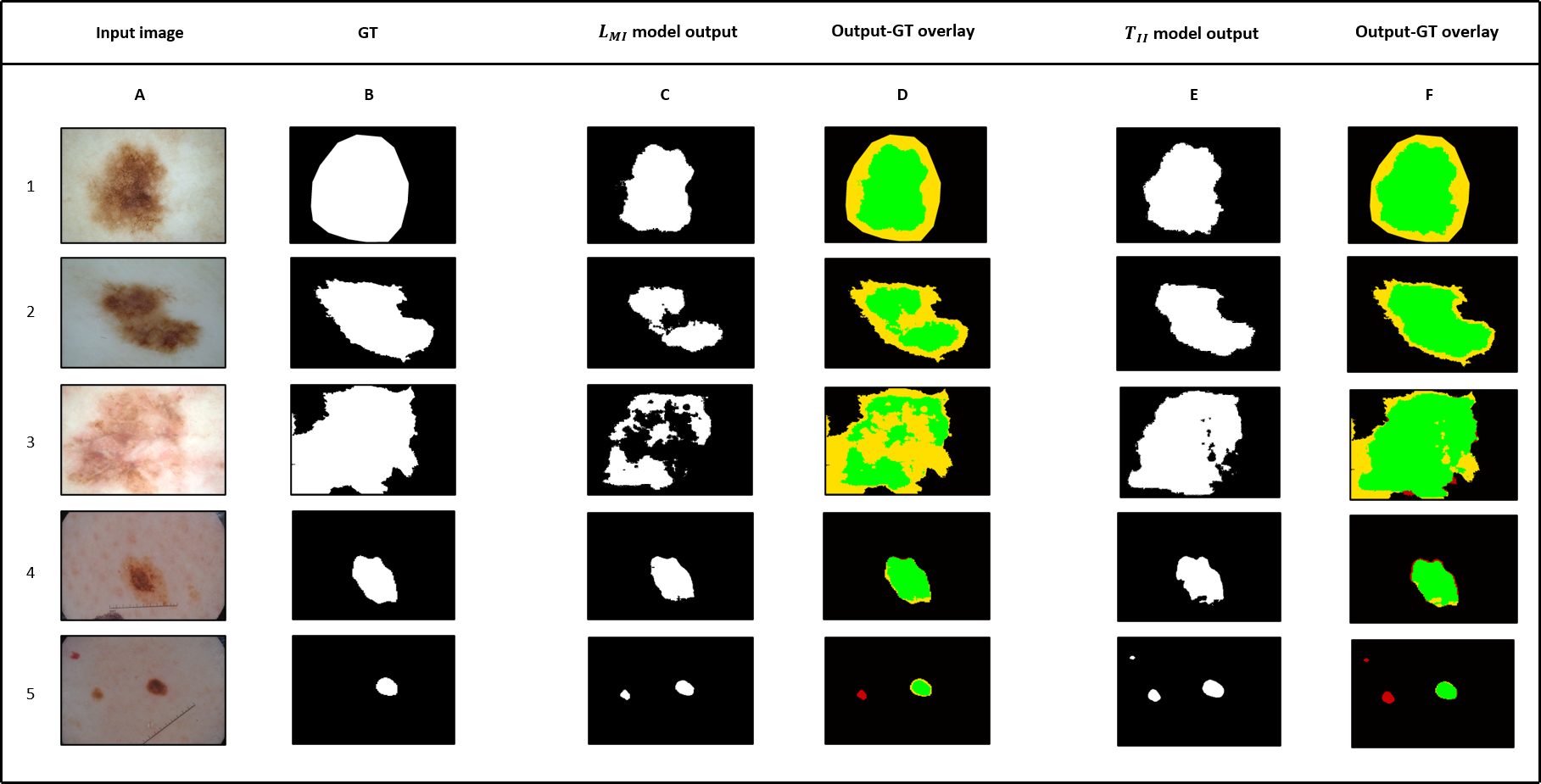}
\caption{}
\end{subfigure}\\
\\
\begin{subfigure}{0.99\textwidth}
\centering\includegraphics[width=0.6\textwidth]{figs/samp_legend.png}
\end{subfigure}\\
\\
\\
\begin{subfigure}{0.99\textwidth}
\centering\includegraphics[width=0.8\textwidth]{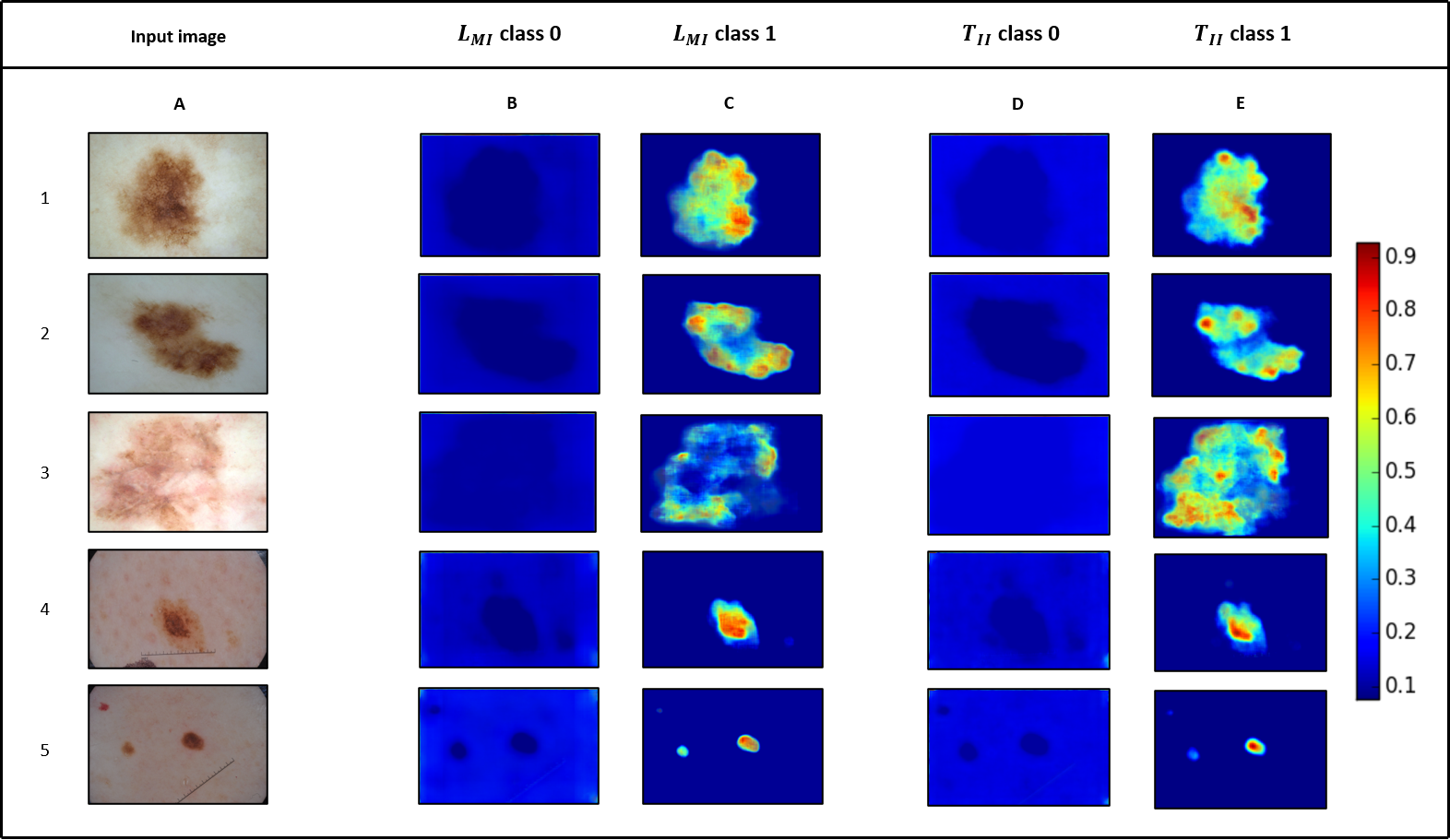}
\caption{}
\end{subfigure}
\caption{\scriptsize{Visualization and explanation of transfer learning ($T_{II}$) and learning from medical images ($L_{MI}$) models on the binary segmentation of skin images with ``indeterminate” clinical labels. \textbf{Panel (i)} left to right columns: A, input RGB image; B, binary mask of clinical ground-truth label; C, binary mask of output image after binary segmentation via $L_{MI}$ model; D, overlay of clinical and $L_{MI}$ model binary output masks; E, binary mask of output image after segmentation via $T_{II}$ model; and F, overlay of clinical and $T_{II}$ model binary output masks. Green, true positive (TP); black, true negative (TN); red, false positive (FP); and yellow, false negative (FN). \textbf{Panel (ii)} Target class-based Grad-CAM output. Class 0 represents the background or non-tumor regions of the skin. Class 1 represents tumors, moles, and lesions. ``GT” indicates clinical ground truth. Color bar represents the degree of model attention and importance, with deeper red indicating most importance and deeper blue indicating least importance.}} \label{sfig5}
\end{figure*}

\begin{figure*}[h!]
\begin{subfigure}{0.99\textwidth}
\centering\includegraphics[width=0.9\textwidth]{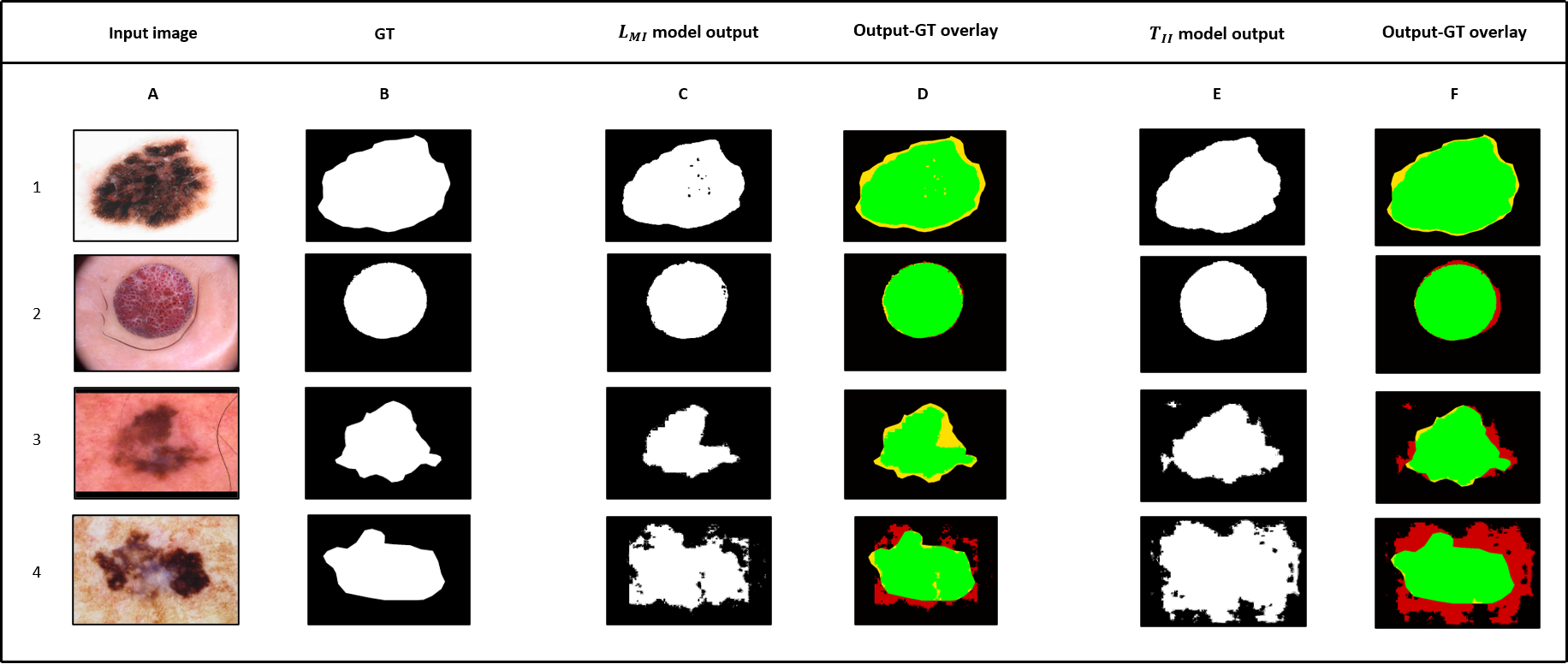}
\caption{}
\end{subfigure}\\
\\
\begin{subfigure}{0.99\textwidth}
\centering\includegraphics[width=0.6\textwidth]{figs/samp_legend.png}
\end{subfigure}\\
\\
\\
\begin{subfigure}{0.99\textwidth}
\centering\includegraphics[width=0.9\textwidth]{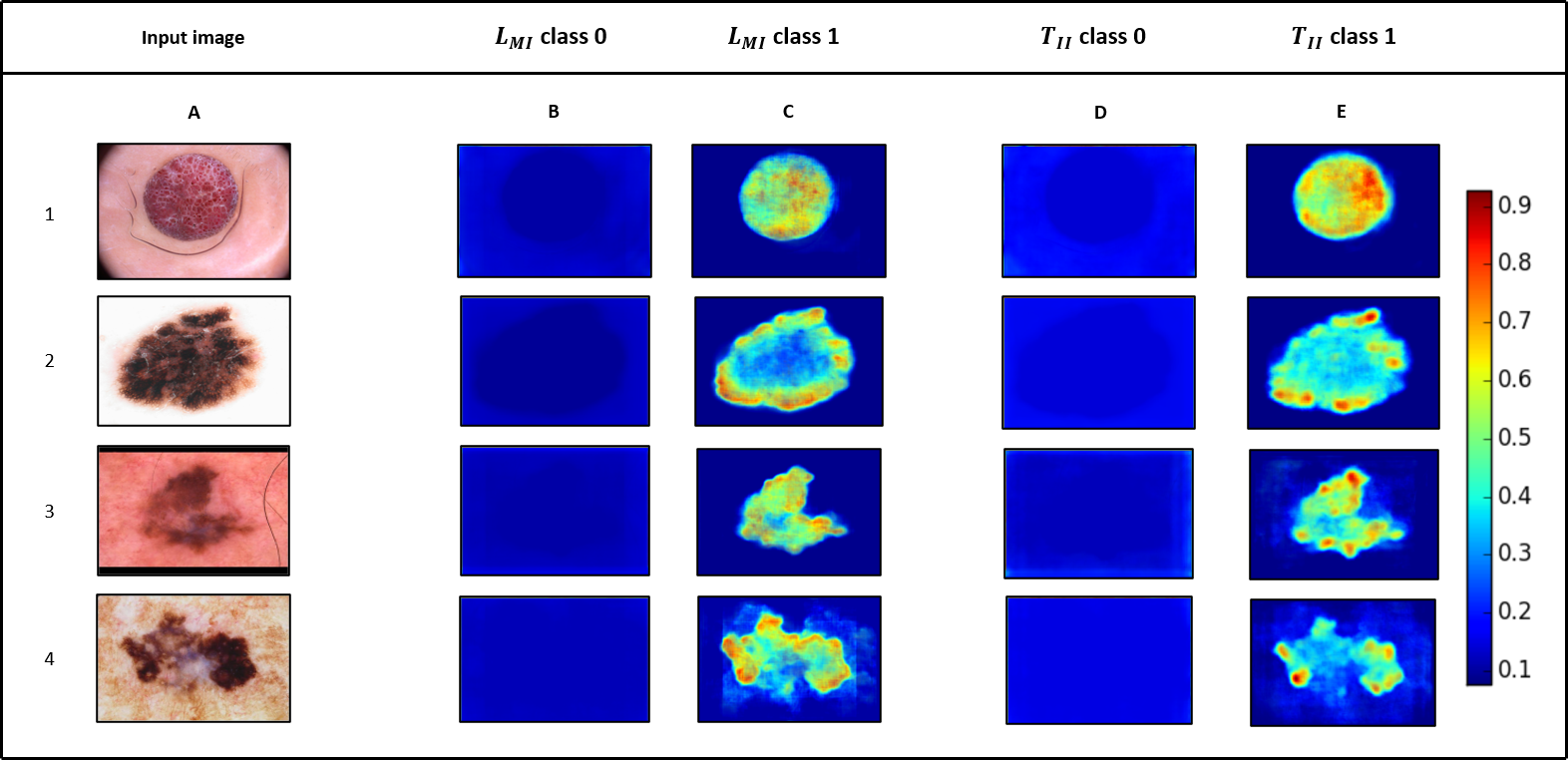}
\caption{}
\end{subfigure}
\caption{\scriptsize{Visualization and explanation of transfer learning ($T_{II}$) and learning from medical images ($L_{MI}$) models on binary segmentation of skin images. \textbf{Panel (i)} left to right columns: A, input RGB image; B, binary mask of clinical ground-truth label; C, binary mask of output image after binary segmentation via $L_{MI}$ model; D, overlay of clinical and $L_{MI}$ model binary output masks; E, binary mask of output image after segmentation via $T_{II}$ model; and F, overlay of clinical and $T_{II}$ model binary output masks. Green, true positive (TP); black, true negative (TN); red, false positive (FP); and yellow, false negative (FN). \textbf{Panel (ii)} Target class-based Grad-CAM output. Class 0 represents the background or non-tumor or non-lesion regions of the skin. Class 1 represents tumors, moles, and lesions. Color bar represents the degree of model attention and importance, with deeper red indicating most importance and deeper blue indicating least importance. ``GT” indicates clinical ground truth.}} \label{sfig6}
\end{figure*}


\begin{figure*}[ht!]
\begin{subfigure}{0.99\textwidth}
\centering\includegraphics[width=0.8\textwidth]{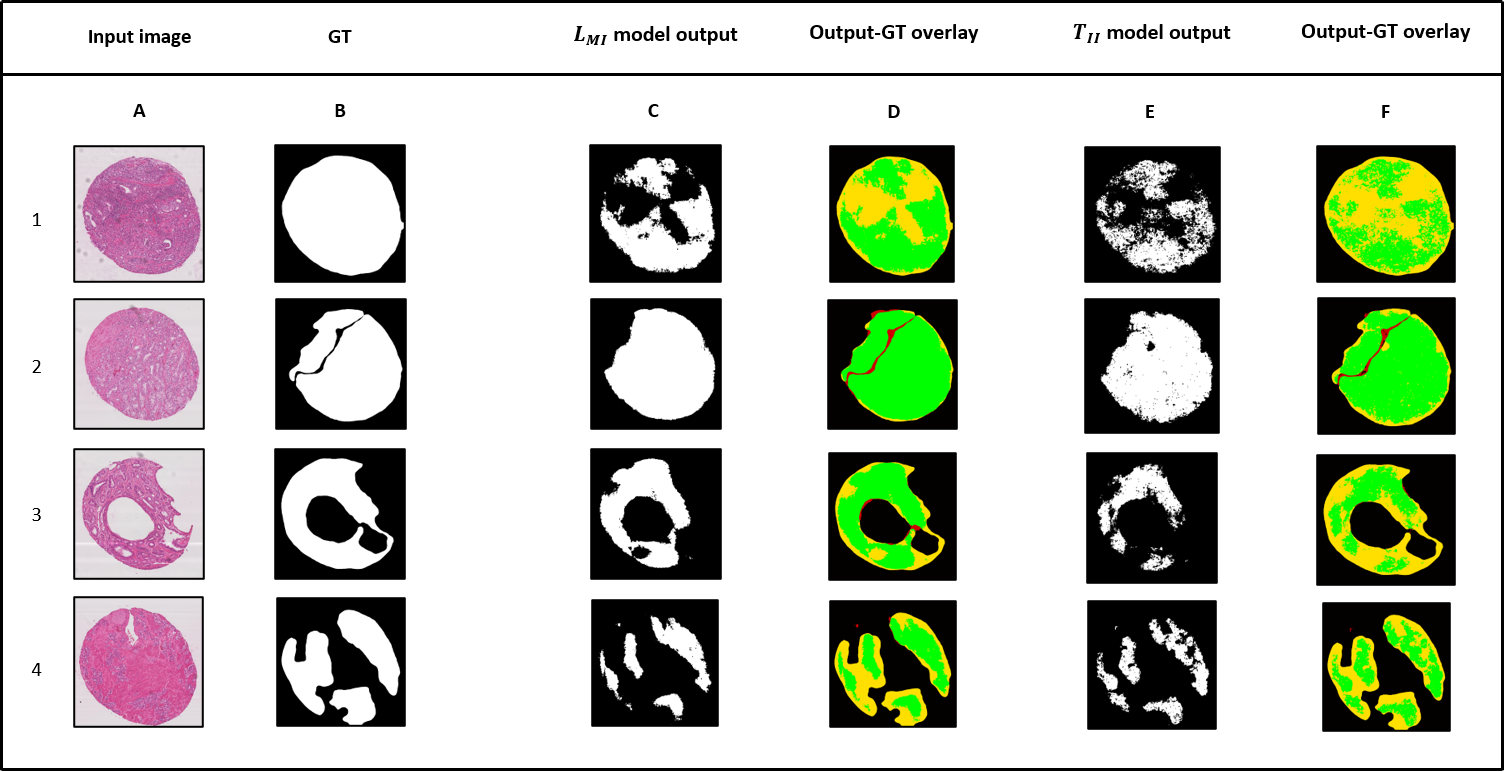}
\caption{}
\end{subfigure}\\
\\
\begin{subfigure}{0.99\textwidth}
\centering\includegraphics[width=0.6\textwidth]{figs/samp_legend.png}
\end{subfigure}\\
\\
\\
\begin{subfigure}{0.99\textwidth}
\centering\includegraphics[width=0.8\textwidth]{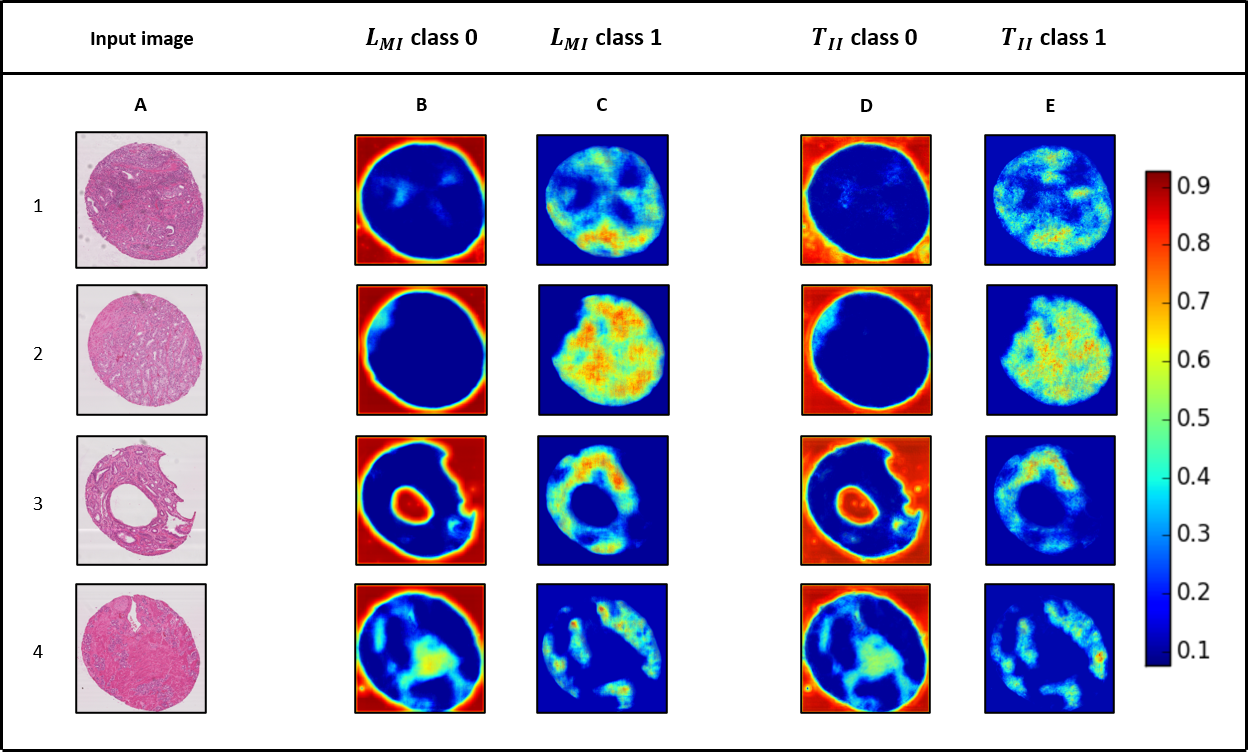}
\caption{}
\end{subfigure}\\
\caption{\scriptsize{Visualization and explanation of transfer learning ($T_{II}$) and learning from medical images ($L_{MI}$) models on the binary segmentation of prostate core biopsy images. \textbf{Panel (i)} left to right columns: A, input RGB image; B, binary mask of clinical ground-truth label; C, binary mask of output image after binary segmentation via $L_{MI}$ model; D, overlay of clinical and $L_{MI}$ model binary output masks; E, binary mask of output image after segmentation via $T_{II}$ model; and F, overlay of clinical and $T_{II}$ model binary output masks. Green, true positive (TP); black, true negative (TN); red, false positive (FP); and yellow, false negative (FN). \textbf{Panel (ii)} Target class-based Grad-CAM output. Class 0 represents the background or benign tumor regions. Class 1 represents the target or object region. For the pathology images, Gleason grade 3, 4, or 5 tumors represent the target regions. Color bar represents the degree of model attention and importance, with deeper red indicating most importance and deeper blue indicating least importance. ``GT” indicates clinical ground truth.}} \label{sfig7}
\end{figure*}

\begin{figure*}[h!]
\begin{subfigure}{0.99\textwidth}
\centering\includegraphics[width=0.6\textwidth]{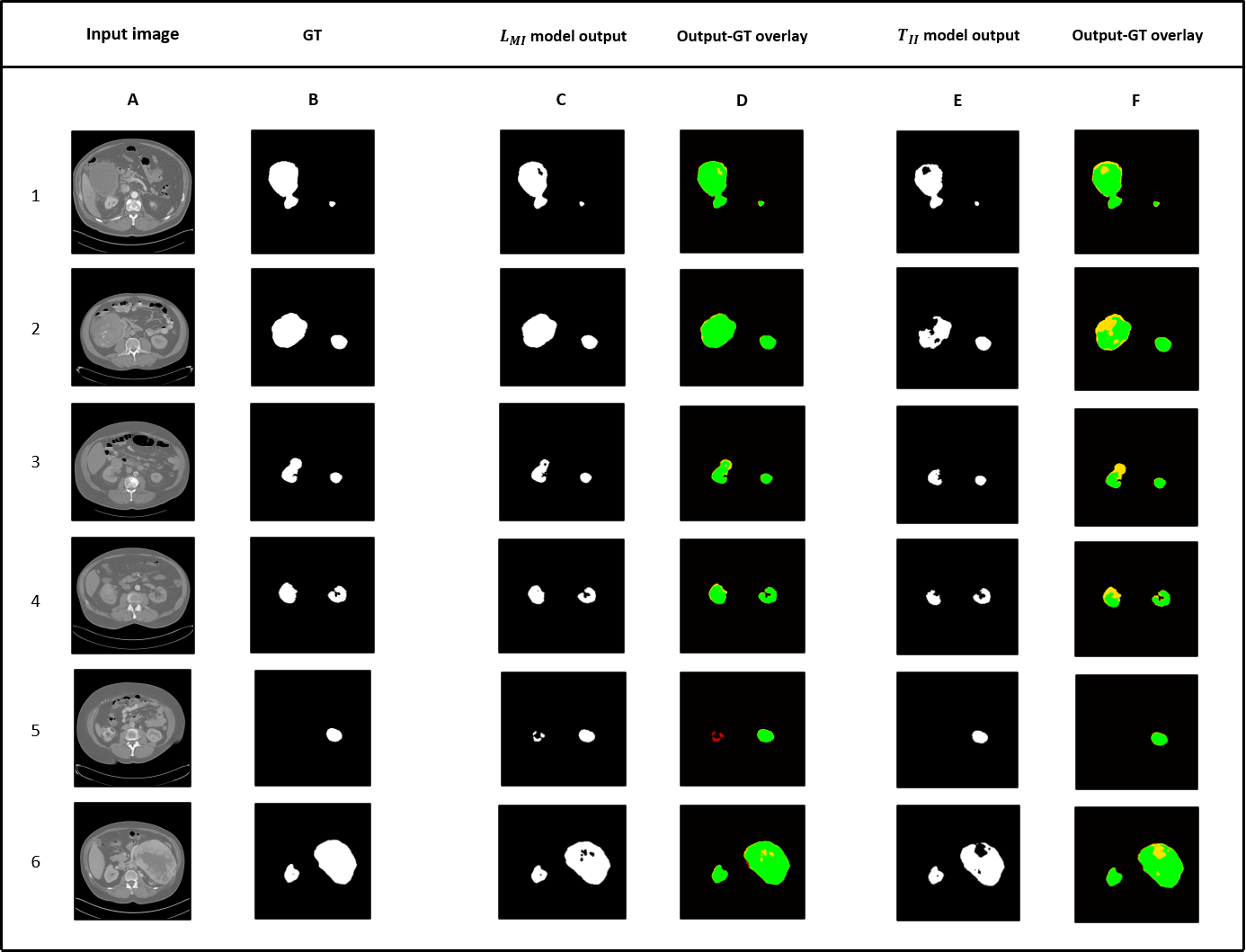}
\caption{}
\end{subfigure}\\
\\
\begin{subfigure}{0.99\textwidth}
\centering\includegraphics[width=0.4\textwidth]{figs/samp_legend.png}
\end{subfigure}\\
\\
\\
\begin{subfigure}{0.99\textwidth}
\centering\includegraphics[width=0.6\textwidth]{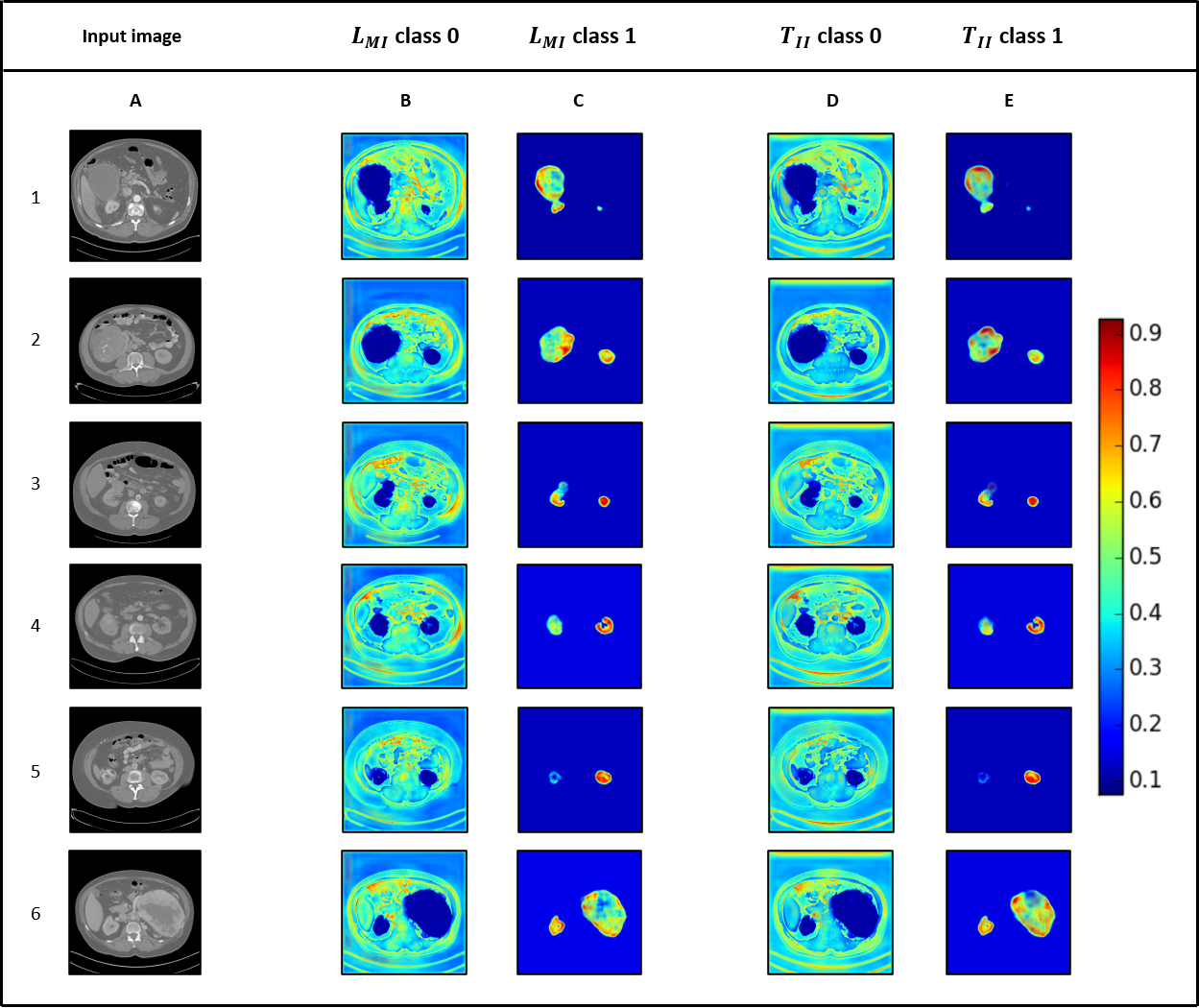}
\caption{}
\end{subfigure}
\caption{\scriptsize{Visualization and explanation of transfer learning ($T_{II}$) and learning from medical images ($L_{MI}$) models on the binary segmentation of computed tomography images with kidneys. \textbf{Panel (i)} left to right columns: A, input RGB image; B, binary mask of clinical ground-truth label; C, binary mask of output image after binary segmentation via $L_{MI}$ model; D, overlay of clinical and $L_{MI}$ model binary output masks; E, binary mask of output image after segmentation via $T_{II}$ model; and F, overlay of clinical and $T_{II}$ model binary output masks. Green, true positive (TP); black, true negative (TN); red, false positive (FP); and yellow, false negative (FN). \textbf{Panel (ii)} Target class-based Grad-CAM output. Class 0 represents the background or non-kidney class or region of the computed tomography image. Class 1 represents the kidney tissue region. Color bar represents the degree of model attention and importance, with deeper red indicating most importance and deeper blue indicating least importance. ``GT” indicates clinical ground truth.}} \label{sfig8}
\end{figure*}

\end{document}